\documentclass[Afour,sageh,times]{sagej}

\usepackage{moreverb,url}
\usepackage{float, bm, caption, mathtools}
\usepackage{amsmath}
\usepackage[colorlinks,bookmarksopen,bookmarksnumbered,citecolor=red,urlcolor=red]{hyperref}
\usepackage[dvipsnames]{xcolor}

\newcommand\BibTeX{{\rmfamily B\kern-.05em \textsc{i\kern-.025em b}\kern-.08em
T\kern-.1667em\lower.7ex\hbox{E}\kern-.125emX}}

\def\volumeyear{2025}

\newcommand{\hb}[1]{\textcolor{black}{#1}}
\newcommand{\hbg}[1]{\textcolor{black}{#1}}

\begin{document}

\runninghead{Modular Robot Control with Motor Primitives}

\title{Modular Robot Control with \\ Motor Primitives}

\author{Moses C. Nah\affilnum{1}, Johannes Lachner\affilnum{1,2}, and Neville Hogan\affilnum{1,2}}

\affiliation{\affilnum{1}Massachusetts Institute of Technology, Department of Mechanical Engineering, USA.\\
\affilnum{2}Massachusetts Institute of Technology, Department of Brain and Cognitive Sciences, USA.}

\corrauth{Moses C. Nah, Massachusetts Institute of Technology, Department of Mechanical Engineering,
Cambridge, Massachusetts, USA.}

\email{mosesnah@mit.edu}

\begin{abstract}
Despite a slow neuromuscular system, humans easily outperform modern robot technology, especially in physical contact tasks. 
How is this possible? 
Biological evidence indicates that motor control of biological systems is achieved by a modular organization of motor primitives, which are fundamental building blocks of motor behavior. 
Inspired by neuro-motor control research, the idea of using simpler building blocks has been successfully used in robotics. 
Nevertheless, a comprehensive formulation of modularity for robot control remains to be established.
In this paper, we introduce a modular framework for robot control using motor primitives. 
We present  two essential requirements to achieve modular robot control: \textit{independence of modules} and \textit{closure of stability}.
We describe key control modules and demonstrate that a wide range of complex robotic behaviors can be generated from this small set of modules and their combinations.
The presented modular control framework demonstrates several beneficial properties for robot control, including task-space control without solving Inverse Kinematics, addressing the problems of kinematic singularity and kinematic redundancy, and preserving passivity for contact and physical interactions. 
Further advantages include exploiting kinematic singularity to maintain high external load with low torque compensation, as well as controlling the robot beyond its end-effector, extending even to external objects.
Both simulation and actual robot experiments are presented to validate the effectiveness of our modular framework.
We conclude that modularity may be an effective constructive framework for achieving robotic behaviors comparable to human-level performance.

\end{abstract}



\maketitle

\section{Introduction}\label{sec:introduction}
\footnotetext{This manuscript has been submitted to the International Journal of Robotics Research for review.}
Despite the significant progress made in recent decades, modern robotic technology has yet to match human-level performance, especially in physical contact tasks. 
Humans can seamlessly manage contact and physical interaction \citep{hogan2022contact}, rapidly adapt to unknown environments \citep{billard2022learning}, efficiently learn complex and dynamic manipulation tasks \citep{billard2019trends}, and can easily generalize the learned movements to novel tasks \citep{black2024pi_0}. 

How do humans achieve such remarkable performance? 
Identifying the underlying principles and applying them to robot control may be a key to bridging the performance gap between humans and robots. 
Nevertheless, this immediately leads to a paradox: Humans achieve their remarkable performance despite their significantly slow neuromuscular system \citep{wolpert1998internal,kawato1999internal,kandel2000principles,slotine2006modularity,hogan2012dynamic,hogan2017physical}.
The fastest neural transmission speed in humans is no more than 120m/s \citep{sperelakis2012cell}, about a million times slower than its robotic counterparts \citep{myers2009exploring}.
The bandwidth of skeletal muscle is considerably less than 10Hz, whereas electro-mechanical actuators can achieve up to hundreds of Hz \citep{paine2013design}.
The transcortical feedback loop delay easily exceeds 100ms \citep{kandel2000principles,safavynia2013sensorimotor}, comparable to a typical update rate of GPS satellites \citep{nikolaidis2018validity}.
Not only are the ``wetware'' (e.g., neurons) and ``actuators'' (e.g., muscles) slower, but also the human neuromuscular system is vastly more complex, further exacerbating the complexity of control \citep{bernstein1967co,slotine2006modularity,jaquier2022riemannian}.
Humans have about 200 degrees of freedom (about 600 skeletal muscles \citep{kuo1994mechanical}), whereas modern robotic systems have fewer than 50 degrees of freedom \citep{kuindersma2016optimization}.

How is this possible? In fact, how the Central Nervous System manages to generate complex motor behavior despite its slow neuromuscular system is one of the central questions in motor control research \citep{ito1970neurophysiological,kawato1987hierarchical,miall1993cerebellum,burdet2001central,todorov2004optimality,jordan2013forward,d2016modularity}. 
One hypothesis which may resolve this paradox is that the motor control of biological systems is achieved by a modular combination of fundamental building blocks called ``motor primitives'' \citep{mussa1994linear,jordan1995modular,thoroughman2000learning,slotine2001modularity,slotine2003modular,flash2005motor, giszter2015motor} (Section \ref{subsec:motor_primitives_biological_systems}).
Using motor primitives and their modular combinations, the significant feedback loop delays of the neuromuscular system may be circumvented by initiating or launching a set of predefined motor primitives.
Once the motor primitives are initiated, robust dynamic behavior can be generated and maintained autonomously, thereby enabling dynamic motor behavior with minimal high-level intervention from the Central Nervous System \citep{hogan2012dynamic,hogan2013dynamic,hogan2017physical}. 
Since motor learning happens at the level of modules and their combinations \citep{d2016modularity}, a wide range of dynamic behaviors can be achieved with remarkable efficiency and flexibility. 
Modularity has also been recognized as a key factor for both the stability and robustness of biological systems \citep{slotine2001modularity,kitano2004biological,kozachkov2022rnns}.

Inspired by motor control research, the idea of using fundamental building blocks (or motor primitives) for robot control has proven effective in many successful applications \citep{billard2022learning,saveriano2023dynamic,nah2024robot} (Section \ref{subsec:control_w_motor_primitives}, \hbg{Appendix \ref{subsubsec:review_DS}}). 
Nevertheless, modularity---a framework to effectively combine these learned motor primitives---has neither been fully articulated nor thoroughly addressed in the extent of their ramifications for robot control (Section \ref{subsec:challenge_for_modularity}). 
Achieving modularity for robot control may be pivotal, as once achieved, the challenge of generating complex motor behavior of the robot can be dramatically simplified \citep{pastor2009learning,alvarez2010switched,mulling2013learning,neumann2014learning,daniel2016hierarchical}.
As much as modularity serves as an effective ``descriptive'' (i.e., analytic) framework for biological systems \citep{mountcastle1979organizing,fodor1983modularity,ghahramani1997modular,wolpert1998multiple,grossberg1998complementary,hartwell1999molecular,mussa1999modular,d2010modularity,davidson2004scaling,lacquaniti2013evolutionary,tagliabue2014modular,d2016modularity,stetter2020modularity}, modularity can serve as an effective ``constructive'' (i.e., synthetic) framework for robot control. 

\subsection{Contributions}\label{subsec:contributions}
In this paper, we present a modular framework for robot control using motor primitives. We articulate 
two essential requirements to achieve functional modularity for robot control: (i) \textit{independence of modules} and (ii) \textit{closure of stability}. 
\begin{itemize}
    \item \textit{Independence of modules}: The superposition principle of virtual trajectories (Section \ref{subsubsubsec:superposition_principle_of_trajectories}) and the superposition principle of mechanical impedances (Section \ref{subsubsubsec:superposition_principle_of_mech_imps}) enable 
    independent modification of the action
    and the corresponding joint torque-command to the robot (Section \ref{subsec:modular_imitation_learning}). The former achieves modular motion planning, where a combination of both discrete and rhythmic movements can be achieved. The latter achieves modularity at the level of robot (torque) command, hence a divide-and-conquer (\textit{divide-et-impera}) \citep{lachner2024divide} strategy for robot control can be applied.
    \item \textit{Closure of Stability}: Using mechanical impedances, the controller can be made robust against contact and physical interaction. Using sufficiently large symmetric and positive-definite joint damping matrices, passivity of the robot is guaranteed against passive environments (Section \ref{subsubsec:achieving_closure_of_stability}). The dynamics of physical interaction can be explicitly regulated by modulating mechanical impedances. The problem of solving Inverse Kinematics is completely avoided (Section \ref{subsec:managing_kinematic_singularity}). The robot can seamlessly go in and out of singularity while preserving passivity, hence the whole robot's workspace can be utilized (Sections \ref{subsubsec:planar_robot_kinematic_singularity} and \ref{subsubsec:iiwa14_singularity}). The two separate problems involved with Inverse Kinematics---kinematic singularity and kinematic redundancy---are both resolved using the same controller (Section \ref{subsubsec:kinematic_singularity_w_redundancy}). Kinematic singularity can be exploited rather than avoided:  we show that high external load can be compensated by low torque actuation (Section \ref{subsubsec:exploiting_kinematic_singularity}).
\end{itemize}
The key to achieving modular robot control is to combine the best of both motor primitives approaches in robotics \citep{nah2024robot}: Elementary Dynamic Actions (EDA) \citep{hogan2012dynamic,hogan2013dynamic,hogan2017physical,nah2024robot} and Dynamic Movement Primitives (DMP) \citep{schaal1999imitation,ijspeert2013dynamical,saveriano2023dynamic} (Section \ref{sec:modules}). 
By utilizing the Norton equivalent network model of EDA \citep{hogan2013general}, the advantages of these two approaches are seamlessly combined. 
As a result, the requirements for independence of modules and closure of stability are both met
(Section \ref{appx:stability_proof}).

This paper extends the work of \cite{nah2024robot} and \cite{nah2024modularity}. 
\cite{nah2024robot} highlighted the differences between EDA and DMP, and concluded the paper with a brief discussion on combining the two approaches. \cite{nah2024modularity} further emphasized the combination of the two approaches and demonstrated that the method is completely free from solving Inverse Kinematics.
\hbg{This paper introduces a rigorous definition of modularity and demonstrates its implementation via a combination of EDA and DMP.}

\subsection{Organization of the Paper}\label{subsec:organization_paper}
Section \ref{sec:prior_research} reviews prior research on motor primitives in biological systems (Section \ref{subsec:motor_primitives_biological_systems}) and their application in robot control (Section \ref{subsec:control_w_motor_primitives}). Additionally, challenges to achieving modular robot control are reviewed (Section \ref{subsec:challenge_for_modularity}).
Section \ref{sec:modules} provides a definition of a module and introduces the fundamental modules used for robot control. 
\hb{We show that using these fundamental modules and their combination satisfies the independence and closure of stability properties required for modular robot control. }
Section \ref{sec:applications} presents robotic applications of the modular robot control algorithm.
Section \ref{sec:discussion_futurework} covers discussion and future work, and Section \ref{sec:conclusion} provides a conclusion.

Appendices \ref{appx:SO3} and \ref{appx:H1} provide the necessary mathematical details which are essential for understanding this paper. 
Readers familiar with the work of \cite{shuster1993survey}, \cite{murray1994mathematical}, and \cite{lynch2017modern} may skip these appendices. 
Appendix \ref{sec:DMP_review} presents a detailed review of DMP used in this paper. 
Readers interested in the implementation details of the robotic demonstration may refer to this Appendix.
Appendix \ref{appx:alternative_formulation} presents an alternative formulation of the module for controlling the robot's spatial orientation.
\hbg{Appendix \ref{subsubsec:review_DS} provides an overview of the Dynamical Systems (DS) approach, one of the major motor primitives approaches in robotics, but not used in this paper.}

\section{Review of Prior Research}\label{sec:prior_research}
\subsection{Motor Primitives in Biological Systems}\label{subsec:motor_primitives_biological_systems}
The concept of motor primitives and their modular architecture dates back at least a century, with a number of subsequent experiments providing support for its existence in biological systems. 

Sherrington was one of the first to suggest ``reflex'' as a fundamental element of complex motor behavior \citep{sherrington1906integrative, elliott2001century}.
It was suggested that reflexes can be treated as basic units of motor behavior that, when chained together, produce more complex movements \citep{clower1998early}.

Bernstein, who first formulated the Degrees of Freedom (or  Motor Equivalence) problem \citep{bernstein1935problem,bernstein1967co,latash2021one}, suggested ``synergies'' as motor primitives to account for the simultaneous motion of multiple joints (i.e., kinematic synergies \citep{santello1998postural,santello2013neural}) or activation of multiple muscles (i.e., muscle synergies \citep{tresch1999construction,d2003combinations,d2015modularity,d2016modularity}). 
Synergies account for the complexity of controlling high-dimensional neuromuscular systems by dimensionality reduction. 
Complex high-dimensional movements of biological systems can be reduced to a small set of synergies and their modular combinations \citep{bizzi2008combining,hogan2012dynamic,d2013control, d2015modularity,aoi2016neuromusculoskeletal}.
Experiments with spinalized frogs showed that complex lower limb movements can be deconstructed into a small set of convergent ``force fields'' and their linear combinations \citep{giszter1993convergent,mussa1994linear,bizzi1995modular,kargo2000rapid,giszter2013motor}. 

Rhythmic, repetitive movements and goal-directed discrete movements have also been suggested as two distinct classes of primitives \citep{sternad2000interaction,schaal2004rhythmic,hogan2007rhythmic,sternad2013transitions}.
Rhythmic movements (e.g., locomotion) are phylogenetically old motor behaviors found in most biological species \citep{ronsse2009computational}.
Central Pattern Generators \citep{brown1911intrinsic, brown1912factors,dimitrijevic1998evidence,marder2001central} which are specialized neural circuits for generating rhythmic motor patterns, have been identified in biological systems.
For rhythmic and planar hand movements of unimpaired human subjects, the two-thirds power law \citep{morasso1982trajectory,lacquaniti1983law,viviani1985segmentation,viviani1995minimum,karklinsky2015timing} and its generalization \citep{huh2015spectrum} provide an empirical relation between (angular) speed and curvature of the hand trajectory \citep{hermus2020evidence}. 
Upper-limb cyclical aiming tasks showed lower variability (i.e., the speed-accuracy trade-off, Fitts' law \citep{fitts1954information}) than discrete movements \citep{guiard1993fitts,guiard1997fitts,sternad2003rhythmic,sternad2008towards,huys2015does,sternad2017human}. 

Discrete movements (e.g., goal-directed reaching movements) are phylogenetically younger motor behaviors, particularly observed in primates with developed upper extremities \citep{ronsse2009computational}.
For discrete arm reaching movements of unimpaired human subjects, the hand trajectory in external coordinates remains essentially invariant, with a distinctive unimodal bell-shaped velocity profile \citep{morasso1981spatial,flash1985coordination,hogan1987moving,flash1991arm, won1995stability,krebs1998robot,rohrer2002movement,hogan2006motions,hogan2012dynamic,berret2016don,park2017moving}.
Studies have shown that (a sequence of) discrete movement(s) can be decomposed into finite submovements and their linear combinations \citep{flash2005motor,park2017moving}. 
Kinematic patterns composed of submovements have also been observed in stroke patients with upper-extremity motor impairments \citep{krebs1998robot}. 
Although their movements appeared fragmented, each segment followed a highly stereotyped submovement profile \citep{rohrer2004submovements}.
While a concatenation of discrete movements may generate rhythmic movements, neural imaging studies have effectively ruled out this hypothesis \citep{schaal2004rhythmic}, further supporting that rhythmic and discrete movements constitute distinct classes of primitives.

Recently, there is growing evidence that ``stable postures'' may be considered to be a distinct class of motor primitives \citep{shadmehr2017distinct, jayasinghe2022neural}. 
Studies have shown that neural circuits responsible for maintaining postures are distinct from those that control movement.

\subsection{Robot Control based on Motor Primitives}\label{subsec:control_w_motor_primitives}
Inspired by human motor control, the idea of using motor primitives as fundamental building blocks for robot control has been used.
The approach has been successful in a wide range of applications, including robot juggling \citep{schaal1994robot,schaal1996one,ploeger2022controlling,andreu2024beyond}, dynamic object throwing and grasping \citep{kim2014catching,salehian2016dynamical,liu2022solution,bombile2023bimanual,abeyruwan2023agile}, table tennis \citep{muelling2010learning,peters2013towards,peters2014experiments,gomez2016using}, peg-in-hole assembly \citep{fasse1997spatial,lachner2024divide,haddadin2024unified}, in-hand object manipulation \citep{khadivar2023adaptive}, controlling flexible and high-dimensional objects \citep{nah2020dynamic,nah2021manipulating,nah2023learning}, robotic locomotion \citep{righetti2006programmable,ijspeert2008central,ajallooeian2013central}, and many others \citep{billard2022learning,saveriano2023dynamic}. 

Three major motor-primitives approaches exist in robotics: Dynamical Systems (DS) \citep{billard2022learning}, Dynamic Movement Primitives (DMP) \citep{ijspeert2013dynamical,saveriano2023dynamic}, and Elementary Dynamic Actions (EDA) \citep{hogan2012dynamic,nah2024robot}. 
\hbg{Since this paper primarily focuses on the combination of DMP and EDA with its modular property, overviews of DMP and EDA are included. For an overview of DS and its comparison with the presented modular approach, readers may refer to Appendix \ref{subsubsec:review_DS}.} 

\subsubsection{Dynamic Movement Primitives}\label{subsubsec:review_DMP}
As with DS-based approaches, Dynamic Movement Primitives (DMP) also encode movements as dynamical systems with specific attractor dynamics. 
First proposed by \cite{ijspeert2002learning,schaal2003control} and later extended by various formulations \citep{righetti2008pattern,pastor2009learning,hoffmann2009biologically,khansari2012dynamical,zhou2017task,zhou2019learning,koutras2020novel}, DMP has been an effective framework for trajectory planning and generation. 

Both DS-based approaches and DMP share the same fundamental principle for movement planning and generation. 
Nevertheless, technical differences between the two approaches exist.\footnote{For readers interested in further details beyond this paper, refer to Section 3.5 of \cite{saveriano2023dynamic}.}
Compared to DS-based approaches which learn a general form of autonomous dynamical system, DMP uses a specific form of dynamical system, which consists of a stable linear system and an additional nonlinear input $\mathbf{F}$, expressed by $\dot{\mathbf{x}} = \mathbf{A}\mathbf{x} + \mathbf{F}(\mathbf{x})$ \citep{ijspeert2013dynamical,saveriano2023dynamic}. 
Given a stable (or Hurwitz \citep{bullo2024contraction}) matrix $\mathbf{A}$, the desired movement (or attractor dynamics) is learned via the nonlinear input $\mathbf{F}$, which consists of a weighted sum of nonlinear activation functions (Sections \ref{subsubsec:nonlinear_forcing_term} and \ref{subsubsec:transformation_system}).
This learning process is often referred to as Imitation Learning (IL) \citep{schaal1999imitation,schaal2007dynamics} (Section \ref{subsubsec:imitation_learning}), which involves finding the best-fit weights of the activation functions using (or imitating) trajectory data provided by human demonstration. 

In principle, any regression method can be used to learn the weights of the nonlinear input from the given data \citep{stulp2013learning}. 
A common approach is Locally Weighted Regression (LWR) \citep{ijspeert2013dynamical}. 
The weights can be learned through batch learning \citep{saveriano2023dynamic} or updated incrementally as data is collected over time \citep{atkeson1997locally,schaal1998constructive}. 
Probabilistic formulation can also be used to learn the best-fit weights \citep{paraschos2013probabilistic}. 
Recently, methods which account for the geometric structure of the learned weights have been proposed \citep{lee2023equivariant,lee2024mmp++}. 
Not only from human demonstration, the weights can also be learned by Reinforcement Learning \citep{peters2008natural,argall2009survey,muelling2010learning,theodorou2010reinforcement,theodorou2010generalized, stulp2011hierarchical,buchli2011learning,kober2013reinforcement}. 
Given a reward function, the weights \hb{(and hence the control policy)} which maximize the total reward are learned. 
Methods such as Natural Actor-Critic \citep{peters2008natural}, Policy Learning by Weighting Exploration with  Return (PoWER) \citep{muelling2010learning}, Policy Improvement with Path Integrals (PI$^2$) \citep{theodorou2010reinforcement,theodorou2010generalized,buchli2011learning} have been proposed. 

One of the benefits of using DMP is the temporal and spatial invariance property for trajectory generation \citep{ijspeert2013dynamical,saveriano2023dynamic}. 
Once the best-fit weights are learned, the trajectory can be spatially scaled or rotated, or even temporally scaled (i.e., making the trajectory faster or slower), while preserving its qualitative behavior.
These temporal and spatial scalings can be achieved by modifying a small set of parameters. 
Another advantage is real-time trajectory modification, allowing rapid adaptation to unknown environments---a feature shared with DS-based approaches (Appendix \ref{subsubsec:review_DS}). 
For example, real-time obstacle avoidance can be achieved by adding a repulsive force field into the learned attractor dynamics \citep{park2008movement,hoffmann2009biologically,pastor2009learning,zhou2017task}. 
These properties make DMP preferable over spline-based methods \citep{ijspeert2013dynamical}, which explicitly depend on spline nodes. 
Note that spline-based methods require recalculation of these nodes when performing spatial scaling or real-time trajectory modification.

As with DS-based approaches, DMP formulations which account for various geometric structures of the trajectory have been proposed. 
For instance, DMP to learn robot trajectories for spatial orientation \citep{pastor2011online,ude2014orientation,koutras2020correct,abu2021periodic} and symmetric positive-definite matrices (e.g., stiffness, damping matrices) \citep{abu2020geometry} have been proposed.

Further variations of DMP have been proposed for specific applications.
For instance, generating a combination of discrete and rhythmic movements for DMP has been addressed, either by using bifurcation \citep{ernesti2012encoding} or through Contraction Theory \citep{lohmiller1998contraction,slotine2003modular,nah2025combining}. 
DMP which can adapt to arbitrary via-points have been introduced to improve extrapolation capabilities \citep{zhou2019learning}.

Compared to DS-based approaches, DMP employs a specific dynamical system to control the phase (or temporal dynamics) of the trajectory (Section \ref{subsubsec:canonical_system}). 
While this offers a temporal invariance property and enables explicit temporal modulation of the trajectory \citep{koutras2020dynamic,anand2021real}, it also introduces an implicit dependency on time for DMP, making it distinct from DS-based approaches \citep{saveriano2023dynamic}. 
Moreover, as with DS-based approaches, additional methods are necessary to map the learned movements into robot commands \citep{nah2024robot}.

\subsubsection{Elementary Dynamic Actions}\label{subsubsec:review_EDA}
Elementary Dynamic Actions (EDA)\footnote{As discussed in \cite{nah2024robot}, the original name suggested by \cite{hogan2012dynamic} was ``Dynamic \emph{Motor} Primitives.'' However, to avoid confusion due to similarity to ``Dynamic \emph{Movement} Primitives,'' here we use the term ``Elementary Dynamic Actions'' (EDA). For the differences between these two approaches, readers may refer to \cite{nah2024robot}.} is a generalization of impedance control \citep{hogan1985impedance} prominently used in robotics, but to additionally account for the observable motor behavior of biological systems. 
In addition to encoding discrete and rhythmic movements, EDA includes mechanical impedance as a distinct class of motor primitives to manage contact and physical interaction. 
Movement primitives and mechanical impedances are seamlessly integrated through a nonlinear network model \citep{hogan2013general,hogan2017physical}, inspired by nonlinear electrical circuit theory.

A fundamental concept of EDA is to explicitly manage the dynamics of physical interaction through mechanical impedances. 
For instance, by monitoring or regulating the energy of the robot, safe physical interaction with a dynamically changing but passive environment can be achieved by imposing passivity.
In addition, mechanical impedances can be shaped to handle multiple tasks with different priorities \citep{lachner2022shaping}. 

Using EDA, motion planning can be simplified by leveraging kinematic patterns observed in human motor behavior. 
For instance, dynamic manipulation of flexible objects can be achieved by optimizing the parameters of a single point-to-point discrete movement defined in joint space \citep{nah2020dynamic, nah2021manipulating, nah2023learning}. 
\hb{The kinematic pattern of this discrete movement is directly derived from observable human movement behavior.}
Furthermore, the nonlinear network model used in EDA facilitates motion planning by allowing direct combination of discrete and/or rhythmic movements \citep{nah2024modularity} (Section \ref{subsubsubsec:superposition_principle_of_trajectories}).

The asymptotic stability of EDA for achieving convergence towards a fixed target location has been demonstrated using constant mechanical impedances. Control strategies have been proposed for both joint-space and task-space position, accounting for cases without kinematic redundancy \citep{takegaki1981new} and with kinematic redundancy \citep{arimoto2005natural}. 
Studies addressing the case of time-varying mechanical impedances also exist \citep{kronander2016stability, abu2020variable}.

By shaping the virtual elastic potential field for robot control, repulsive force fields can also be integrated. A key application is real-time obstacle or collision avoidance, achieved by superimposing a repellent potential field around the obstacle \citep{andrews1983impedance, khatib1986potential, newman1987high, koren1991potential, huang2009velocity, tulbure2020closing, hjorth2020energy}. 
Compared to DS-based approaches (Appendix \ref{subsubsec:review_DS}) and DMP (Section \ref{subsubsec:review_DMP}), EDA directly incorporates the repulsive force field into the robot's (torque) command.

EDA has been studied not only for controlling joint-space or task-space positions but also for task-space orientation. For example, mechanical impedances for controlling spatial orientation have been developed using spatial rotation matrices \citep{fasse1997spatial, fasse1997spatial2} and unit quaternions \citep{caccavale1998quaternion, caccavale1999six, caccavale2000quaternion}. A comprehensive review of these approaches is provided in \citep{seo2023geometric,seo2025se}. 

The primary control objective of EDA is to determine appropriate movement and impedance parameters for a given robotic task. 
Although methods for finding the movement parameters \citep{nah2020dynamic, nah2023learning} and impedance parameters \citep{lachner2024divide} have been proposed, finding these parameters for general manipulation tasks is still an open challenge.

\subsection{Challenge to Modularity for Robot Control}\label{subsec:challenge_for_modularity}
Modularity is a fundamental concept across various fields, including biology \citep{hartwell1999molecular, schlosser2000modularity}, engineering \citep{baldwin1999design, simon2012architecture}, and control theory \citep{popov1973hyperstability, lohmiller1998contraction, slotine2006modularity}, serving as both a framework for developing and understanding complex behaviors.

In robot control, research to incorporate modularity has been explored, as once achieved, the challenge of generating complex motor behavior of the robot can be dramatically simplified \citep{bruyninckx2001open,hermann2005modular,cui2021toward,decre2013extending}.
However, several challenges remain in ensuring the essential properties required for modular robot control: \textit{independence} and \textit{closure of stability}.

\subsubsection{Independence}\label{subsubsec:modularity_independence}
One of the key properties to achieve modular robot control is independence: The properties of each individual control module must be preserved after combination, and the parameters of each module should be independently modifiable without the need to modify the others.
This independence property of modularity is crucial for flexibility and adaptability in complex motor tasks.

DMP proposed a nominally modular control framework such as Mixture of Motor Primitives \citep{kulic2009online,alvarez2010switched,niekum2012learning,mulling2013learning,paraschos2013probabilistic,daniel2016hierarchical}. 
The key concept is to separately learn each movement module, which can then be combined with others to generate complex movements.
However, these approaches conflate task description (i.e., extrinsic coordinates) and task execution (i.e., intrinsic coordinates), as the learning happens at the level of joint-space rather than in task-space \citep{mulling2013learning,paraschos2013probabilistic}.
As a result, independence is violated in task-space due to the nonlinear Forward Kinematics map of the robot.
Note that this result applies to any motor primitive approach that relies on joint-space learning for task-space control \citep{ploeger2021high,ploeger2022controlling}.

Another consequence is the coupling of task-space position and orientation.
Fundamentally, task-space control must separately account for both position and orientation of the robotic manipulator \citep{murray1994mathematical,lynch2017modern}.
Achieving independent control of task-space position and orientation would enable a flexible and adaptive robot control framework \citep{saveriano2019merging,sun2024se}.
However, for kinematic primitives learned in joint-space, the corresponding motions for task-space position and orientation are coupled and independent control of both movements cannot be achieved. 

Another challenge is to generate a combination of movements, while enabling independent modulation of each movement.
For instance, consider generating a combination of discrete and rhythmic movements, a movement which has various potential applications \hb{including} polishing \citep{khadivar2021learning} and peg-in-hole assembly \citep{sloth2020towards,lachner2024divide}.
Prior approaches achieve a combination of these movements by using a specific form of dynamical system and its property (e.g., bifurcation) \citep{ernesti2012encoding,khadivar2021learning}.
However, these approaches do not allow an independent modulation of each movement, which thereby limits the range of movements that can be generated. 
Note that this challenge of independence has been recently addressed using Contraction Theory for DMP \citep{nah2025combining}.

Another example is to generate a sequence of movements, which has been successfully used for dexterous robot manipulations \citep{burridge1999sequential} or for ``backchaining'' \citep{lozano1984automatic} motion planning for Unmanned Aerial Vehicles \citep{majumdar2017funnel}.
Central to this approach is to use a funnel which is associated with a strictly stable Lyapunov function\footnote{A strictly stable Lyapunov function is a positive-definite function whose time derivative is negative definite \citep{slotine1991applied}. } as a basic module. 
Robust motion planning is achieved by sequentially chaining the learned funnels towards the goal location.
Nevertheless, the method requires a strict relation between the adjacent funnels---the end of the previous funnel must be within the start of the subsequent funnel.
As a result, modifying a single module affects the entire subsequent sequence of funnels. 
Moreover, the method is restricted to sequencing discrete movements and excludes the incorporation of rhythmic movements, further restricting its versatility for motion planning.
The reason is the usage of strictly stable Lyapunov functions \citep{strogatz2018nonlinear,gan2021existence}. 
As stated by \cite{strogatz2018nonlinear}, closed orbits are forbidden for strictly stable Lyapunov functions. 
The problem is often circumvented by decomposing the rhythmic movement into a sequence of discrete movements \citep{medina2017learning}.
Nevertheless, the problem of violating independence still remains. 

\subsubsection{Closure of Stability}\label{subsubsec:closure_of_stability}
Another essential property for modular robot control is \textit{closure of stability}: the stability of the robot must be guaranteed, even against contact and physical interaction. 
In fact, \textit{a combination of stable elements has no reason to be stable} \citep{lohmiller1998contraction,slotine2001modularity,slotine2003modular,tsukamoto2021learning}, hence care is required to immediately conclude stability even when using independently stable dynamical systems.

A majority of approaches based on movement primitives have focused on using position-commanded robots with trajectories generated by kinematic primitives \citep{ijspeert2013dynamical,saveriano2019merging,koutras2020correct,billard2022learning,saveriano2023dynamic}. 
Although these approaches can achieve high tracking accuracy for free-space motion, a position-commanded robot is not appropriate for tasks involving contact and physical interaction \citep{de2008atlas,abu2020variable}. 
\hb{Note that this limitation is shared with other modular control algorithms, but not with those based on motor primitives, using joint-position (or velocity) commands \citep{bruyninckx2001open,decre2009extending,decre2013extending}.}
Moreover, to generate task-space trajectories using position-commanded robots, solving Inverse Kinematics is required \citep{billard2022learning}. 
Hence, the problem of kinematic singularity and kinematic redundancy must be explicitly handled. 
Solutions to these two separate problems have been identified \citep{nakamura1986inverse,baillieul1990resolution,vahrenkamp2012manipulability,haviland2023manipulator}, yet the problem is still commonly observed even in modern control approaches \citep{chi2023diffusion,seo2023geometric,cohn2024constrained,haddadin2024unified}.
The problem of Inverse Kinematics is often circumvented by learning joint-space trajectories.
But again, that violates the independence property of modularity (Section \ref{subsubsec:modularity_independence}).

The limitations of using position-commanded robots are often addressed using torque-commanded robots. 
A common approach is to use Operational Space control \citep{khatib1987unified} and its variations \citep{park2006contact,nakanishi2008operational,martin2019variable,shaw2022rmps}. 
With this controller, a task-space trajectory can be directly commanded to the robot, and compliant robot behavior is generated to manage contact and physical interaction.
Unfortunately, the controller violates passivity \citep{nakanishi2008operational,lachner2022geometric}, which is essential to achieve safe physical interaction with an unknown environment (Section 3.4 of \cite{stramigioli2015energy}). 
Dynamic decoupling via the inertia matrix \citep{khatib1987unified,martin2019variable,shaw2022rmps}, and a null-space projection matrix to manage kinematic redundancy \citep{khatib1987unified,dietrich2015overview,ott2015prioritized} violate the passivity of the robot. 
Furthermore, the problem of kinematic singularity remains, and additional methods to manage kinematic singularity (e.g., damped least-square inverses \citep{wampler1986manipulator,chiaverini1994review,buss2005selectively}) must be employed. 
This further exacerbates the complexity of the robot controller \citep{lachner2022geometric}. 
The problem of kinematic singularity can be avoided by using impedance control \citep{abu2024unified}, as the controller does not require solving Inverse Kinematics \citep{siciliano2008springer}. 
However, such controllers can still encounter the problem of getting stuck in singular configurations. 
Moreover, for kinematically redundant robots, the problem of undesirable drift in joint-space occurs \citep{mussa1991integrable,hermus2021exploiting}.

\section{Basic Modules for Robot Control}\label{sec:modules}
\hb{In this section, we define a control module and present the four basic modules used for robot control.  
We also show that composing robot control using these modules achieves both independence and closure of stability for modularity. }

\subsection{Preliminary}
For robot control, an $n$ degrees of freedom open-chain robotic manipulator with ideal torque actuators is considered. 
The governing differential equations of the robot dynamics are given by \citep{spong2008robot}:
\begin{equation}\label{eq:manipulator_equation}
    \mathbf{M}(\mathbf{q})\ddot{\mathbf{q}} + \mathbf{C}(\mathbf{q}, \dot{\mathbf{q}})\dot{\mathbf{q}} + \mathbf{g}(\mathbf{q}) = \bm{\tau}_{in}(t) + \bm{\tau}_{ext}(t)
\end{equation}
In this equation, $\mathbf{q}\equiv \mathbf{q}(t)\in \mathcal{Q}(=\mathbb{R}^{n}
)$ is the robot joint configuration;\footnote{Strictly speaking, the robot joint trajectory $\mathbf{q}$ is a curve on the Configuration Manifold $\mathcal{Q}$, i.e., $\mathbf{q}(t)\in \mathcal{Q}$. The geometric structure of the Configuration Manifold depends on the robot's topology. For instance, if the robot consists of $n$ revolute joints, $\mathcal{Q}=\mathcal{T}^{n}$, where $\mathcal{T}^{n}$ is an $n$-Torus. Nevertheless, we consider that the element of $\mathcal{Q}$ is mapped to an element of $\mathbb{R}^{n}$ by some choice of coordinates, hence $\mathbf{q}(t)$ can be expressed by $\mathbb{R}^{n}$, i.e., $\mathbf{q}(t)\in \mathbb{R}^{n}$. Note that the charted elements of $\mathcal{Q}$ are locally isomorphic to $\mathbb{R}^{n}$ \citep{spivak1999comprehensive,do1992riemannian}. } $\mathbf{M}(\mathbf{q})\in\mathbb{R}^{n\times n}$ and $\mathbf{C}(\mathbf{q}, \dot{\mathbf{q}})\in\mathbb{R}^{n\times n}$ are the mass and Coriolis/centrifugal matrices, respectively; $\mathbf{g}(\mathbf{q})\in\mathbb{R}^{n}$ is the (co)vector arising from the gravitational potential energy $U_{g}:\mathcal{Q} \rightarrow \mathbb{R}$, i.e., $\mathbf{g}(\mathbf{q})=\frac{\partial U_{g}}{\partial \mathbf{q}}(\mathbf{q})$; $\bm{\tau}_{ext}(t)\in\mathbb{R}^{n}$ is the resultant effect of external forces expressed as torque; $\bm{\tau}_{in}(t)\in\mathbb{R}^{n}$ is the torque input commanded to the robot. $\bm{\tau}_{in}(t)$ consists of a summation of control modules. \hb{Details of the control modules are presented in Section \ref{subsec:basic_control_modules}}. 

The proposed modular robot control algorithm assumes a torque-controlled robot.
For a position-commanded robot, the inverse dynamics model of the robot is required to map the calculated torque to the corresponding joint-position command. This further complicates the approach and deviates from the original purpose of modular robot control \citep{nah2024robot}. 
Therefore, to fully leverage the advantages of modular control, a torque-commanded robot should be used. 

\begin{figure}
    \centering
    \includegraphics[trim={0.0cm 0.0cm 0.0cm 0.0cm}, width=0.99\columnwidth, clip, page=1]{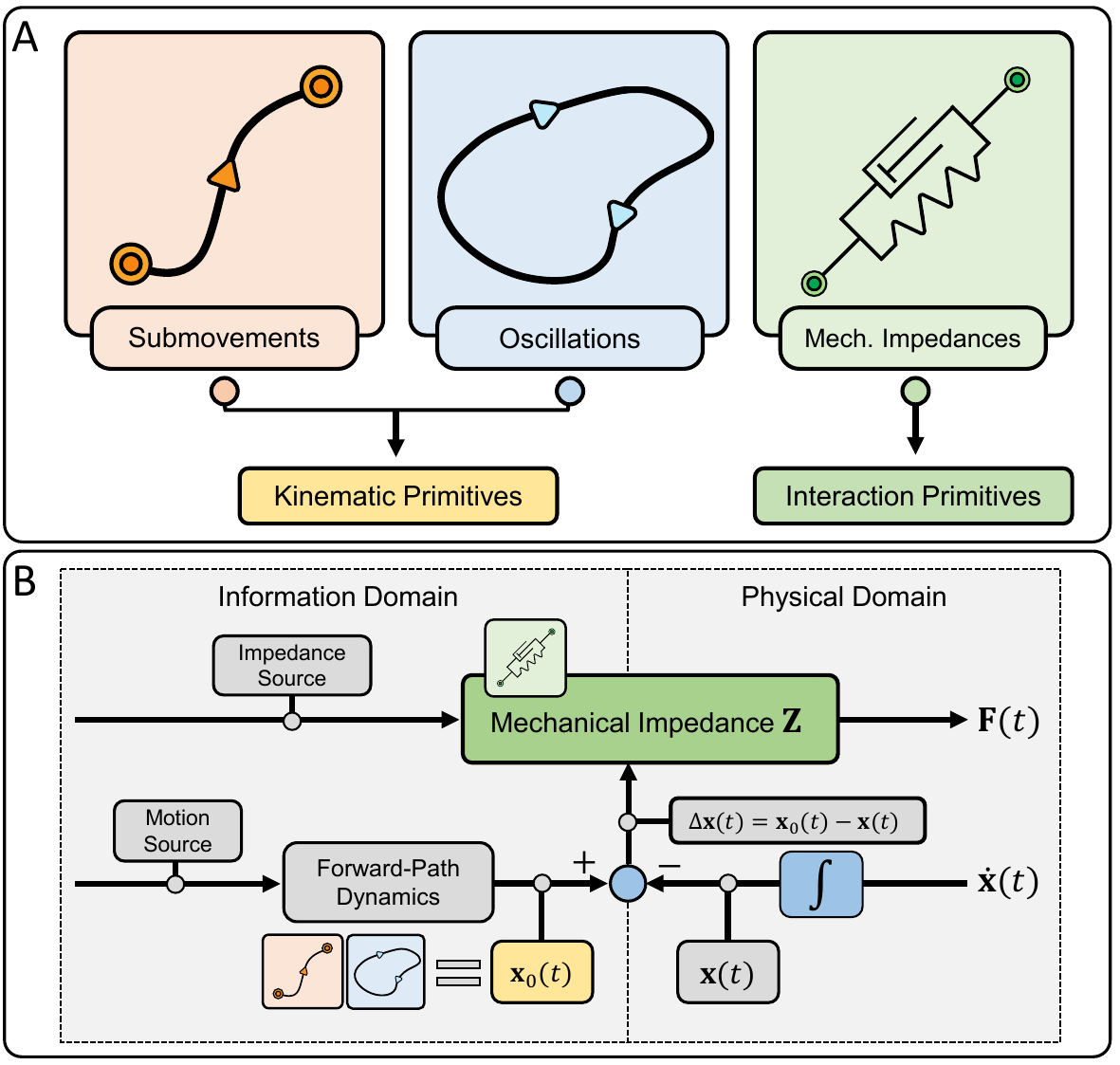}
    \caption{ (A) The three Elementary Dynamic Actions (EDA). Submovements (orange box) and oscillations (blue box) correspond to kinematic primitives and mechanical impedances (green box) manage physical interaction. (B) Elements of EDA combined using a Norton equivalent network model. The virtual trajectory $\mathbf{x}_0(t)$ (yellow box) consists of submovements (orange box) and/or oscillations (blue box), and mechanical impedances $\mathbf{Z}$ (green box) govern the dynamics of physical interaction. The Norton equivalent network model provides an effective framework to combine the two distinct domains in robotics: the information domain (left) and physical domain (right). Figure modified from \cite{hogan2013general,hogan2017physical}. }
    \label{fig:eda}
\end{figure}

\subsection{Definition of a Module}\label{subsec:def_module}
\hb{A module consists of a combination of EDA and DMP. 
Since the details of both EDA and DMP have been presented elsewhere, this Section includes only the necessary information relevant to this paper.
For more details on EDA, refer to \cite{nah2024robot}; 
for more details on DMP, refer to Appendix \ref{sec:DMP_review}.}

\hb{EDA, introduced by \cite{hogan2012dynamic, hogan2013dynamic,hogan2017physical}, consists of (at least) three distinct classes of motor primitives (Figure \ref{fig:eda}A):
\begin{itemize}
    \item Submovements for goal-directed (possibly path-constrained) discrete movements.
    \item Oscillations for rhythmic, repetitive movements.
    \item Mechanical impedances to manage physical interaction. 
\end{itemize}
Submovements and oscillations comprise the kinematic primitives, while mechanical impedances comprise the interaction primitives of EDA. }

\hb{The three distinct classes of EDA can be combined using a Norton equivalent network model \citep{hogan2013general,hogan2017physical}, which provides an effective framework to relate the three classes of EDA (Figure \ref{fig:eda}B). 
In detail, the forward-path dynamics specifies the virtual trajectory $\mathbf{x}_0(t)$, which consists of submovements and/or oscillations. 
The interactive dynamics, which consists of mechanical impedances $\mathbf{Z}$, determines the generalized force output $\mathbf{F}(t)$ with the generalized displacement input $\Delta \mathbf{x}(t)$. }

\hb{ Not only for EDA, but using the Norton equivalent network model also provides an effective framework to merge the two distinct domains that are involved in robot control \citep{hogan2013general,hogan2017physical,nah2024robot}.
Planning the virtual trajectory $\mathbf{x}_0(t)$ occurs within the ``information domain,'' (Figure \ref{fig:eda}B) which is fundamentally uni-directional (i.e., the input affects output but not vice-versa).
On the other hand, mechanical impedance governs the dynamics occurring in the ``physical domain,'' (Figure \ref{fig:eda}B) which is fundamentally bi-directional (i.e., mutual causality between input and output, exemplified by the Newton's Third Law of Action-reaction).
Integration of the three elements of EDA, with the two distinct domains for robot control, is achieved by the Norton equivalent network model \citep{hogan2017physical}. }

\hb{Based on the three elements of EDA, the Norton equivalent network model which combines the elements of EDA and DMP for motion planning of the virtual trajectory (Appendix \ref{sec:DMP_review}), a definition of a module for robot control can be established. 
A module, which is a basic, distinct functional unit for robot control, is defined by a pair of mechanical impedance $\mathbf{Z}$ and virtual trajectory $\mathbf{x}_{0}(t)$ to which the impedances are connected.
Given $\mathbf{x}(t)$, the (generalized) displacement $\Delta\mathbf{x}(t)$ is the input to the mechanical impedance, which outputs (generalized) force $\mathbf{F}(t)$. 
This output is mapped to the robot torque command input $\bm{\tau}_{in}(t)$ (Equation \eqref{eq:manipulator_equation}).}

\begin{figure*}
    \centering
    \includegraphics[trim={0.0cm 1.5cm 0.0cm 0.0cm}, width=0.90\textwidth, clip, page=1]{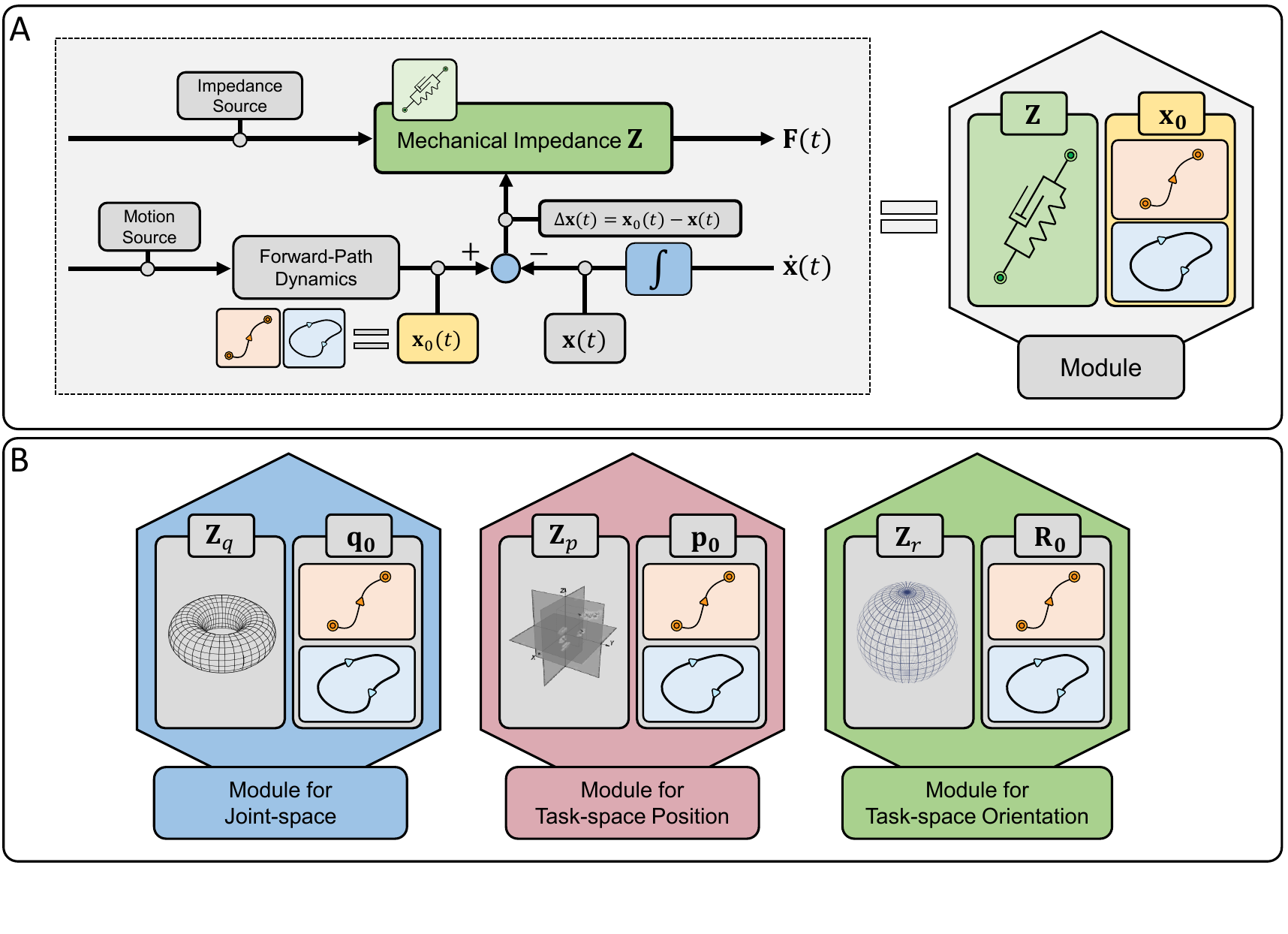}
    \caption{(A) A definition of a module, which consist of a pair of mechanical impedance $\mathbf{Z}$ and the virtual trajectory $\mathbf{x}_0$ to which the impedance is connected \citep{nah2024robot}. For the virtual trajectory, a combination of discrete and/or rhythmic movements is used. (B) The major modules used for robot control: a module for joint-space ($\mathbf{Z}_q, \mathbf{q}_0$) (Section \ref{subsec:module_for_joint_space}), a module for task-space position ($\mathbf{Z}_p, \mathbf{p}_0$) (Section \ref{subsec:module_for_task_space_position}), a module for task-space orientation, $(\mathbf{Z}_r, \mathbf{R}_0)$, which could either use spatial rotation matrices (Section \ref{subsec:module_for_task_space_orientation_SO3}) or unit quaternions (Section \ref{subsec:module_for_task_space_orientation_H1}). Note that the conversion between spatial rotation matrices and unit quaternions can be conducted (Appendix \ref{appx_sec:quat_so3_conversion}). }
    \label{fig:modules}    
\end{figure*}

\subsection{Four Basic Modules for Robot Control}\label{subsec:basic_control_modules}
\hb{Given a definition of a module, we present the} four major modules which are extensively used for a wide range of control tasks. 
The four modules include a module for joint-space control, a module for task-space position, and modules for task-space orientation, both $\text{SO}(3)$ and $\mathbb{H}_1$ (Figure \ref{fig:modules}).
To introduce these four modules, their notations and definitions are presented:
\begin{itemize}
     \item For the module associated with joint-space control, the parameters of the module are denoted by $\mathbf{Z}_{q}$ and $\mathbf{q}_0$. Impedance $\mathbf{Z}_{q}$ refers to ``joint-space impedance,'' while $\mathbf{q}_0$ denotes the ``virtual joint configuration'' to which the joint-space impedance is connected. 
    \item For the module associated with task-space control of position, the parameters of the module are denoted by $\mathbf{Z}_{p}$ and $\mathbf{p}_0$. Impedance $\mathbf{Z}_{p}$ refers to ``task-space impedance for position,'' or ``translational impedance,'' while $\mathbf{p}_0$ denotes the ``virtual task-space position'' to which the the translational impedance is connected.
    \item For the module associated with task-space control of orientation using spatial rotation matrices (Appendix \ref{appx:SO3}) (i.e., elements of $\text{SO}(3)$), the parameters of the module are denoted by $\mathbf{Z}_{r}$ and ${}^{S}\mathbf{R}_0$. Impedance $\mathbf{Z}_{r}$ refers to ``task-space impedance for orientation, $\text{SO}(3)$'' or ``rotational impedance for spatial rotation matrix,'' while ${}^{S}\mathbf{R}_0$ denotes the ``virtual task-space orientation'' (expressed with respect to frame $\{S\}$) to which the rotational impedance is connected, using spatial rotation matrices.
    \item For the module associated with task-space control of orientation using unit quaternions (Appendix \ref{appx:H1}) (i.e., elements of $\mathbb{H}_1$), the parameters of the module are denoted by $\mathbf{Z}_r$ and ${}^{S}\vec{\mathbf{q}}_0$. ${}^{S}\vec{\mathbf{q}}_0$ denotes the ``virtual task-space orientation'' (expressed with respect to frame $\{S\}$) to which the rotational impedance is connected, using unit quaternions. Note that the notation for mechanical impedance is identical to those for spatial rotation matrices. The reason for this choice will be clarified in Section \ref{subsec:module_for_task_space_orientation_H1}.
\end{itemize}
Note that one can simply combine the task-space position and orientation using the Homogeneous transformation matrix $\mathbf{H}\in\text{SE}(3)$, where $\text{SE}(3)$ is the Special Euclidean Group in three-dimensional space \citep{murray1994mathematical,lynch2017modern}. 
Mechanical impedances defined over $\text{SE}(3)$ have been extensively discussed based on the rigorous theory of Lie Groups and Lie Algebras \citep{fasse1997spatial,fasse1997spatial2,stramigioli2001variable,rashad2019port,seo2023geometric}.
However, we show that there are advantages to explicitly separating the control of position and orientation, such as enabling modular control for task-space position and orientation (Section \ref{subsec:modular_imitation_learning}).

\subsubsection{Module for Joint-space Control}\label{subsec:module_for_joint_space}
A module for joint-space control is defined by a pair comprising joint-space impedance $\mathbf{Z}_{q}$ and its virtual joint trajectory $\mathbf{q}_{0}$:
\begin{equation}\label{eq:joint_space_module}
    \mathbf{Z}_{q}(\mathbf{q},\mathbf{q}_{0}) = \mathbf{K}_q(\mathbf{q}_0-\mathbf{q}) +\mathbf{B}_q(\dot{\mathbf{q}}_0-\dot{\mathbf{q}}) 
\end{equation}
In this equation, $\mathbf{K}_q, \mathbf{B}_q\in \mathbb{R}^{n\times n}$ correspond to joint-space stiffness and damping matrices, respectively. 
This module is often referred to as a first-order joint-space impedance controller \citep{takegaki1981new,hogan1985impedance,slotine1991applied}. 
It is also commonly known as Proportional-Derivative (PD) control in joint-space, although care is required since that definition is only valid when the robot is an ideal torque-actuated system \citep{won1997comment}. 

Given $\mathbf{q}_0$, the virtual elastic potential energy associated with $\mathbf{K}_{q}$, $U_{q}:\mathcal{Q}\rightarrow \mathbb{R}$ is defined by:
\begin{equation}\label{eq:elastic_potential_joint_space}
    U_{q}(\mathbf{q},\mathbf{q}_0) = \frac{1}{2}(\mathbf{q}-\mathbf{q}_0)^{\top}\mathbf{K}_{q}(\mathbf{q}-\mathbf{q}_0)
\end{equation}
Hence, the stiffness term in Equation \eqref{eq:joint_space_module} is derived from the partial derivatives of $U_{q}$ with respect to $\mathbf{q}$:
$$
    \mathbf{K}_q(\mathbf{q}_0-\mathbf{q}) = -\frac{\partial U_{q}}{\partial \mathbf{q}} (\mathbf{q})
$$

\subsubsection{Module for Task-space Control, Position}\label{subsec:module_for_task_space_position}
A module to control task-space position is defined by a pair comprising task-space impedance for position $\mathbf{Z}_{p}$ and its virtual trajectory for task-space position $\mathbf{p}_{0}(t)$:
\begin{equation}\label{eq:task_space_module_position}
    \mathbf{Z}_{p}(\mathbf{p},\mathbf{p}_{0}) = \mathbf{J}_p^{\top}(\mathbf{q}) \{ \mathbf{K}_p(\mathbf{p}_0-\mathbf{p}) +\mathbf{B}_p(\dot{\mathbf{p}}_0-\dot{\mathbf{p}}) \}
\end{equation}
In this equation, $\mathbf{p}(t)\in\mathbb{R}^{3}$ is the position of the point on a robot of interest; usually, $\mathbf{p}(t)$ denotes the end-effector of the robot, although any point on (or even off) the robot can be used (Section \ref{subsubsec:virtual_trajectory_outside_robot}); $\mathbf{K}_p, \mathbf{B}_p\in \mathbb{R}^{3\times 3}$ are the translational stiffness and damping matrices, respectively; $\mathbf{J}_p(\mathbf{q})\in \mathbb{R}^{3\times n}$ is the Jacobian matrix for task-space position, where $\dot{\mathbf{p}}=\mathbf{J}_p(\mathbf{q})\dot{\mathbf{q}}$.
This module is often referred to as a first-order task-space impedance controller for position \citep{takegaki1981new}.

Note that with this module, the problem of Inverse Kinematics is completely avoided; one can directly command $\mathbf{p}_0(t)$ to the robot. 
Moreover, the controller is free from the problem of kinematic singularity, since the Jacobian transpose is used rather than its (generalized, or pseudo-) inverse. 
Finally, the kinematic redundancy of the robot can be effectively managed by incorporating the module for joint-space control (Equation \eqref{eq:joint_space_module}) (Section \ref{subsubsec:kinematic_singularity_w_redundancy}).

Given $\mathbf{p}_0$, the virtual elastic potential energy associated with $\mathbf{K}_{p}$, $U_{p}:\mathbb{R}^{3}\rightarrow \mathbb{R}$ is defined by:
\begin{equation}\label{eq:task_space_impedance_position_potential}
    U_{p}(\mathbf{p}, \mathbf{p}_{0}) = \frac{1}{2}(\mathbf{p}-\mathbf{p}_0)^{\top}\mathbf{K}_{p}(\mathbf{p}-\mathbf{p}_0)
\end{equation}
With the Forward Kinematics map of the robot $\mathbf{h}_{p}$, the energy $U_{p}$ defined over $\mathbb{R}^{3}$ can also be defined over the joint-space $\mathcal{Q}$, $U_{p} \circ \mathbf{h}_{p}:\mathcal{Q}\rightarrow \mathbb{R}$, \hbg{where $\circ$ is a composition operator.}
From this, for a given $\mathbf{p}_0$, the stiffness term in Equation \eqref{eq:task_space_module_position} is derived from the partial derivatives of $U_{p} \circ \mathbf{h}_{p}$ with respect to $\mathbf{q}$:
$$
 \mathbf{J}_p^{\top}(\mathbf{q}) \mathbf{K}_p\{\mathbf{p}_0-\mathbf{h}_{p}(\mathbf{q}) \}=-\frac{\partial }{\partial \mathbf{q}} (U_{p} \circ \mathbf{h}_{p}) (\mathbf{q})
$$
Again, the virtual trajectories $\mathbf{p}_0$ and $\mathbf{p}(\mathbf{q})$ need not be the robot's end-effector, and can be defined at any arbitrary point, even outside the robot's physical structure. 
This flexibility is particularly advantageous for tasks requiring the stabilization of specific external points during manipulation, such as pouring liquids or pointing with an attached tool (Section \ref{subsubsec:virtual_trajectory_outside_robot}).

\subsubsection{Module for Task-space Control, Orientation, SO(3)}\label{subsec:module_for_task_space_orientation_SO3}
Using $\text{SO}(3)$ spatial rotation matrices, a module to control task-space orientation is defined by a pair of task-space impedance for orientation $\mathbf{Z}_{r}$ and ${}^{S}\mathbf{R}_{0}(t)$, where ${}^{S}\mathbf{R}_{0}(t)\in \text{SO(3)}$ is a rotation matrix which expresses the virtual frame $\{0\}$ with respect to $\{S\}$ \citep{fasse1996control,fasse1997spatial,fasse1997spatial2}:
\begin{align}\label{eq:module_for_SO3_complex}
\begin{split}
    \mathbf{Z}_{r}({}^{S}\mathbf{R}_{B}, {}^{S}\mathbf{R}_{0}) =& \frac{\partial}{\partial \mathbf{q}}\text{tr}(\mathbf{G}_{r} {}^{S}\mathbf{R}_{B}^{\top}  {}^{S}\mathbf{R}_{0}) \\
    &- {}^{B}\mathbf{J}_r^{\top}(\mathbf{q})\mathbf{B}_{r}{}^{B}\bm{\omega}
\end{split}
\end{align}
In this equation, ${}^{S}\mathbf{R}_{B}\in\text{SO}(3)$ is the rotation matrix which can be derived by the Forward Kinematics map $\mathbf{h}_{r}:\mathcal{Q}\rightarrow \text{SO}(3)$, where $\mathbf{h}_{r}(\mathbf{q})={}^{S}\mathbf{R}_{B}(\mathbf{q})$;
${}^{B}\bm{\omega}(t)\in\mathbb{R}^{3}$ is the angular velocity of $\{B\}$ with respect to $\{S\}$, expressed in $\{B\}$; ${}^{B}\bm{\omega}(t)$ can be derived by ${}^{B}\bm{\omega}={}^{B}\mathbf{J}_r(\mathbf{q})\dot{\mathbf{q}}$, where ${}^{B}\mathbf{J}_r(\mathbf{q}(t))\in\mathbb{R}^{3\times n}$ is the Body Jacobian matrix \citep{murray1994mathematical,lynch2017modern} for angular velocity; $\mathbf{B}_{r}\in\mathbb{R}^{3\times 3}$ is a rotational damping matrix.

$\mathbf{G}_r \in \mathbb{R}^{3\times 3}$ is a co-stiffness matrix derived from a stiffness matrix $\mathbf{K}_{r}\in\mathbb{R}^{3\times 3}$, where a one-to-one mapping between $\mathbf{G}_r$ and $\mathbf{K}_{r}$ exists \citep{chillingworth1982symmetry}. The definition $\mathbf{K}_r$ and its relation with $\mathbf{G}_r$ is clarified in the next section, when we discuss the module using unit quaternions (Section \ref{subsec:module_for_task_space_orientation_H1}).

Given ${}^{S}\mathbf{R}_0$, the virtual elastic potential energy associated with $\mathbf{G}_r$, $U_{r}:\text{SO}(3)\rightarrow \mathbb{R}$ is defined by:
\begin{equation}\label{eq:potential_impedance_SO3}
    U_{r}( {}^{S}\mathbf{R}_{B}, {}^{S}\mathbf{R}_{0}) = - \text{tr}( \mathbf{G}_{r} {}^{S}\mathbf{R}_{B}^{\top}  {}^{S}\mathbf{R}_{0} )
\end{equation}
This form of potential energy over the $\text{SO}(3)$ manifold was provided by \cite{koditschek1989application}, which is originally from \cite{meyer1971design}.

With the Forward Kinematics map of the robot $\mathbf{h}_{r}$, the energy $U_{r}$ defined over the $\text{SO}(3)$ manifold can also be defined over the joint-space $\mathcal{Q}$, $U_{r} \circ \mathbf{h}_{r}:\mathcal{Q}\rightarrow \mathbb{R}$.
Hence, for a given ${}^{S}\mathbf{R}_0$, the term for stiffness in Equation \eqref{eq:module_for_SO3_complex} is derived from the partial derivatives of $U_{r} \circ \mathbf{h}_{r}$ with respect to $\mathbf{q}$:
$$
 \frac{\partial}{\partial \mathbf{q}}\text{tr}(\mathbf{G}_{r} {}^{S}\mathbf{R}_{B}^{\top}(\mathbf{q}) {}^{S}\mathbf{R}_{0}) =-\frac{\partial }{\partial \mathbf{q}} (U_{r} \circ \mathbf{h}_{r}) (\mathbf{q})
$$
Note that the presented torque command requires a partial derivative of potential energy $U_{r}\circ \mathbf{h}_r$. 
The analytical derivation of this term may be computationally expensive. 
  While the module using unit quaternion addresses this problem (Section \ref{subsec:module_for_task_space_orientation_H1}), an alternative formulation using spatial rotation matrices, which does not require partial derivatives, is also available (Appendix \ref{appx:alternative_formulation}).

\subsubsection{Module for Task-space Control, Orientation, \texorpdfstring{$\mathbb{H}_{1}$}{H1}}\label{subsec:module_for_task_space_orientation_H1}
For controlling task-space orientation using Equation \eqref{eq:module_for_SO3_complex}, the partial derivative with respect to $\mathbf{q}$ is involved.
One can avoid the partial derivatives by an equivalent controller using unit quaternion and its operation \citep{caccavale1998quaternion,caccavale1999six,caccavale1999robot,caccavale2000quaternion,natale2004interaction}.

Given ${}^{S}\mathbf{R}_{B}(t), {}^{S}\mathbf{R}_{0}(t) \in \text{SO}(3)$, the corresponding unit quaternions ${}^{S}\vec{\mathbf{q}}_{B}(t),{}^{S}\vec{\mathbf{q}}_{0}(t)\in\mathbb{H}_{1}$ can be defined (Appendix \ref{appx:subsec_SO3_to_unit_quaternion}). 
The quaternion error between these two quaternions is ${}^{S}\vec{\mathbf{q}}_{B}^{*}(t) \otimes {}^{S}\vec{\mathbf{q}}_{0}(t)\equiv({}^{B}\eta_{0}(t), {}^{B}\bm{\epsilon}_{0}(t))$, which is the unit quaternion representation of ${}^{B}\mathbf{R}_{0}(t)={}^{S}\mathbf{R}_{B}^{\top}(t) {}^{S}\mathbf{R}_{0}(t)$.
With these parameters $({}^{B}\eta_{0}(t),{}^{B}\bm{\epsilon}_{0}(t))$, a module to control task-space orientation using unit quaternions $\mathbb{H}_1$ is defined by a pair of task-space impedance for orientation $\mathbf{Z}_{r}$ and ${}^{S}\vec{\mathbf{q}}_{0}(t)$ \citep{lachner2022geometric}:
\begin{align}\label{eq:impedance_control_quaternion}
    \begin{split}
        \mathbf{Z}_r({}^{S}\vec{\mathbf{q}}_{B}, {}^{S}\vec{\mathbf{q}}_{0}) =& {}^{B}\mathbf{J}_{r}^{\top}(\mathbf{q}) \{ 2 \mathbf{E}^{\top}({}^{B}\eta_{0},{}^{B}\bm{\epsilon}_{0}) \mathbf{K}_r{}^{B}\bm{\epsilon}_{0} \\
        &- \mathbf{B}_r {}^{B}\bm{\omega} \}
    \end{split}
\end{align}
In this equation, $\mathbf{E}({}^{B}\eta_{0},{}^{B}\bm{\epsilon}_{0}) \in \mathbb{R}^{3\times 3}$ is the matrix derived from the quaternion kinematic (or propagation \citep{natale2004interaction}) (Appendix \ref{sec:quat_kin_equation}); $\mathbf{K}_r, \mathbf{B}_r \in \mathbb{R}^{3\times 3}$ are the rotational stiffness and damping matrices.

Given ${}^{S}\vec{\mathbf{q}}_0$, the associated virtual elastic potential energy $U_r:\mathbb{H}_1\rightarrow \mathbb{R}$ is defined by:
\begin{equation}\label{eq:potential_impedance_H1}
    U_r( {}^{S}\vec{\mathbf{q}}_{B}, {}^{S}\vec{\mathbf{q}}_{0} ) = 2 {}^{B}\bm{\epsilon}_{0}^{\top} \mathbf{K}_{r}{}^{B}\bm{\epsilon}_{0}
\end{equation}
The elastic potential energy $U_r( {}^{S}\vec{\mathbf{q}}_{B}, {}^{S}\vec{\mathbf{q}}_{0} )= 2 {}^{B}\bm{\epsilon}_{0}^{\top} \mathbf{K}_r{}^{B}\bm{\epsilon}_{0}$ is equivalent to $U_{r}( {}^{S}\mathbf{R}_{B}, {}^{S}\mathbf{R}_{0})$ (Equation \eqref{eq:potential_impedance_SO3}) with a constant offset, where $\mathbf{G}_{r}$ and $\mathbf{K}_r$ have a one-to-one correspondence (Appendix \ref{appx:sec_equivalence_potential_energy}):
\begin{equation}\label{eq:stiffness_costiffness_matrix}
    \mathbf{K}_r = \text{tr}(\mathbf{G}_{r} )\mathbb{I}_{3}-\mathbf{G}_{r} ~~~~~ \mathbf{G}_{r} = \frac{1}{2}\text{tr}(\mathbf{K}_{r} )\mathbb{I}_{3} - \mathbf{K}_{r}
\end{equation}
Matrices $\mathbf{K}_{r}$ and $\mathbf{G}_{r}$ are referred to as the rotational stiffness and rotational co-stiffness matrices, respectively \citep{chillingworth1982symmetry}.
Note that this relation is equivalent to the relation between inertia and co-inertia matrices (or convected inertia tensor \citep{betsch2001constrained}), that are reported by \cite{wensing2017linear,lee2019geometric,lee2023robot}.

\subsection{Modular Properties}\label{appx:stability_proof}
\hb{By constructing the robot torque controller using the four major modules and their combinations, we demonstrate that the resulting controller satisfies both independence and closure of stability, thereby achieving modularity.} 

\subsubsection{Independence}
\hb{Using EDA and the Norton equivalent network model (Figure \ref{fig:eda}) provides favorable modular properties that simplify the planning and generation of complex robot behaviors.
The two principles that provide such modular properties are the superposition principle of virtual trajectories and the superposition principle of mechanical impedances. }

\paragraph{Superposition Principle of Virtual Trajectories}\label{subsubsubsec:superposition_principle_of_trajectories}
\hb{For a given impedance operator $\mathbf{Z}$, motion planning can be conducted independently with respect to the robotic manipulator and the environment with which it interacts (Section \ref{subsubsec:modularity_independence}). In detail, a combination of movements (both submovements and oscillations) can be achieved by a linear summation of virtual trajectories:}
\begin{equation}
    \mathbf{x}_0(t) = \sum \mathbf{x}_{0,i}(t)
\end{equation}
\hb{This simple yet effective framework provides notable simplification for motion planning or analysis, as a wide repertoire of movements can be generated by (or decomposed into) a linear summation of distinct kinematic primitives. 
For instance, a sequence of discrete movements can be achieved by a linear summation of submovements; a combination of both discrete and rhythmic movements can be achieved by a linear summation of submovements and/or oscillations \citep{nah2024modularity}. }

Aside from an account of observable motor behavior of biological systems, for practical applications, not only submovements and oscillations but also additional trajectory generation methods such as DMP or splines can be used to plan the virtual trajectory. 
This further extends the range of movements that can be achieved for robot control. 
Exploiting this principle also merges the advantages of DMP (Appendix \ref{sec:DMP_review}) with EDA. 
A key question of EDA is planning the virtual trajectory, whereas DMP typically requires an additional method to map the learned movements into robot commands. By encoding the virtual trajectory using DMP, the strengths of both approaches can be seamlessly combined (Section \ref{subsec:modular_imitation_learning}).
 
\paragraph{Superposition Principle of Mechanical Impedances}\label{subsubsubsec:superposition_principle_of_mech_imps}
\hb{Under the assumption that the environment is a (mechanical) admittance (i.e., the dual operator\footnote{Dual operator implies the input/output relation is opposite with the original operator \citep{hogan2018impedance}.} of mechanical impedance, which inputs generalized force $\mathbf{F}(t)$ and outputs generalized displacement $\Delta\mathbf{x}(t)$), mechanical impedances can be linearly superimposed even though each mechanical impedance is a nonlinear operator \citep{hogan2017physical}: }
\begin{equation}\label{eq:superposition_of_mechanical_impedances}
    \mathbf{Z} = \sum \mathbf{Z}_{i}
\end{equation}
\hb{In this equation, generalized displacement $\Delta\mathbf{x}(t)$ which is the argument of each impedance operator is omitted to avoid clutter. Note that the impedance operators of Equation \eqref{eq:superposition_of_mechanical_impedances} can include transformation maps.}

The superposition principle of mechanical impedances provides the independence property at the level of robot command (Section \ref{subsubsec:modularity_independence}). 
One of the key consequences of this modular property is it enables a divide-and-conquer (\textit{divide-et-impera}) strategy for robot control.
A complex control task (with possibly multiple objectives) can be broken down into a set of simpler sub-problems, each sub-problem solved by associating an impedance operator with its virtual trajectory, and then the EDA for each sub-problem can be \textit{linearly} combined to solve the original control task. 
This modular strategy can drastically reduce the complexity of the original control tasks. 
As a result, one can work around the ``curse of dimensionality,'' \citep{bellman1966dynamic} since it reduces the dimensionality (or complexity) of the original control task to multiple sub-problems that are much more computationally manageable.

Furthermore, with this modular property, trajectories planned in different spaces can be linearly combined at the level of joint torque command. 
For instance, trajectories of both task-space position ($\mathbb{R}^3$) and task-space orientation (SO(3)) can be separately planned and linearly combined to generate a combination of both movements (Section \ref{subsec:modular_imitation_learning}).

\subsubsection{Closure of Stability}\label{subsubsec:achieving_closure_of_stability}
We demonstrate that combining the four modules (Section \ref{subsec:basic_control_modules}) using the two superposition principles (Section \ref{subsubsubsec:superposition_principle_of_trajectories}) also ensures closure of stability.
For the stability proof, we show that the robot preserves passivity when interacting with passive environments.
Proofs for both constant and time-varying module parameters are provided. 

\paragraph{For Constant Module Parameters}
Consider a robot controller which consists of three module pairs, $(\mathbf{Z}_q, \mathbf{q}_0)$, $(\mathbf{Z}_p, \mathbf{p}_0)$, $(\mathbf{Z}_r, {}^{S}\mathbf{R}_0)$ (Section \ref{sec:modules}). 
Assume that the parameters of the modules are constant:
\begin{align}
\begin{split}
    &\mathbf{Z}_{q}(\mathbf{q}, \mathbf{q}_0) = \mathbf{K}_q\{\mathbf{q}_0-\mathbf{q}\} - \mathbf{B}_q \dot{\mathbf{q}} \\
    &\mathbf{Z}_{p}(\mathbf{p}, \mathbf{p}_0) = \mathbf{J}_p^{\top}(\mathbf{q}) \{\mathbf{K}_p(\mathbf{p}_0-\mathbf{p})\} \\
    &\mathbf{Z}_{r}({}^{S}\mathbf{R}_{B}, {}^{S}\mathbf{R}_{0}) = \frac{\partial}{\partial \mathbf{q}}\text{tr}(\mathbf{G}_{r} {}^{S}\mathbf{R}_{B}^{\top}  {}^{S}\mathbf{R}_{0}) \\
    &\bm{\tau}_{in}(t) = \mathbf{Z}_{q}(\mathbf{q}, \mathbf{q}_0) + \mathbf{Z}_{p}(\mathbf{p}, \mathbf{p}_0) + \mathbf{Z}_{r} ({}^{S}\mathbf{R}_{B}, {}^{S}\mathbf{R}_{0})
\end{split}    
\end{align}
Assume no external forces are applied to the robot, and gravitational forces are compensated for the robotic controller, i.e., $\mathbf{g}(\mathbf{q}(t))$ can be neglected (Equation \eqref{eq:manipulator_equation}). 
Since the parameters of the modules are constant, the resulting dynamics of the robotic manipulator is an autonomous dynamical system:
\begin{equation}
    \mathbf{M}(\mathbf{q})\ddot{\mathbf{q}} + \mathbf{C}(\mathbf{q}, \dot{\mathbf{q}})\dot{\mathbf{q}}= -\frac{\partial \mathcal{U}}{\partial \mathbf{q}}(\mathbf{q}) - \mathbf{B}_q \dot{\mathbf{q}}
\end{equation}
where $\mathcal{U}(\mathbf{q}):\mathcal{Q}\rightarrow \mathbb{R}$ is defined by:
\begin{equation}
    \mathcal{U}(\mathbf{q}) = (U_q + U_p \circ \mathbf{h}_p + U_r \circ \mathbf{h}_r)(\mathbf{q})
\end{equation}

Define a following Lyapunov function $\mathcal{V}(\mathbf{q}, \dot{\mathbf{q}})$, which is the total kinetic and potential energy of the robotic manipulator:
\begin{equation}\label{eq:lyapunov_function_robot}
    \mathcal{V}(\mathbf{q}, \dot{\mathbf{q}}) = \frac{1}{2}\dot{\mathbf{q}}^{\top}\mathbf{M}(\mathbf{q})\dot{\mathbf{q}} + \mathcal{U}(\mathbf{q})
\end{equation}
The time derivative of the Lyapunov function provides us:
\begin{align*}
    \frac{d}{dt}\mathcal{V}(\mathbf{q}, \dot{\mathbf{q}}) &= \frac{1}{2}\dot{\mathbf{q}}^{\top}\dot{\mathbf{M}}(\mathbf{q})\dot{\mathbf{q}} + \dot{\mathbf{q}}^{\top}\mathbf{M}(\mathbf{q})\ddot{\mathbf{q}} + \dot{\mathbf{q}}^{\top} \frac{\partial \mathcal{U}}{\partial \mathbf{q}}(\mathbf{q}) \\
    &= - \dot{\mathbf{q}}^{\top}\mathbf{B}_q \dot{\mathbf{q}} \le 0 
\end{align*}
For the derivation, we used $\dot{\mathbf{M}}(\mathbf{q}) - 2\mathbf{C}(\mathbf{q}, \dot{\mathbf{q}})$ is a skew-symmetric matrix \citep{slotine1991applied}.
Given an autonomous dynamical system with a Lyapunov function whose derivative is negative semi-definite, according to LaSalle's invariance principle \citep{lasalle1960some,slotine1991applied,sastry2013nonlinear}, the robot asymptotically converges to one of the local minima of $U_{total}(\mathbf{q})$. 

\paragraph*{Remarks}
\begin{itemize}
    \item For modules with translational damping matrix $\mathbf{B}_p$ (Section \ref{subsec:module_for_task_space_position}) and rotational damping matrices $\mathbf{B}_r$ (Sections \ref{subsec:module_for_task_space_orientation_SO3} and \ref{subsec:module_for_task_space_orientation_H1}), the time derivative of the Lyapunov function is given by:
    \begin{align*}
        \frac{d}{dt}\mathcal{V}(\mathbf{q}, \dot{\mathbf{q}}) =& - \dot{\mathbf{q}}^{\top}\{ \mathbf{B}_q +\mathbf{J}^{\top}_p(\mathbf{q})\mathbf{B}_p \mathbf{J}_p(\mathbf{q}) \\
        &+ \mathbf{J}^{\top}_r(\mathbf{q})\mathbf{B}_r \mathbf{J}_r(\mathbf{q})  \} \dot{\mathbf{q}}
    \end{align*}
    Hence, a faster asymptotic convergence to one of the local minima of $\mathcal{U}(\mathbf{q})$ is achieved. 
    \item Function $\mathcal{U}(\mathbf{q})$ consists of the Forward Kinematics map of the robotic manipulator. Hence, a numerical method can be used to identify the local minima of $\mathcal{U}(\mathbf{q})$.
\end{itemize}

\paragraph{For time-varying module parameters}
Assume that the module parameters are now time-varying, i.e., $\mathcal{U}(\mathbf{q},t):\mathcal{Q}\times \mathbb{R}_{\ge 0}\rightarrow \mathbb{R}$. 
The total virtual elastic potential energy of the modules is now an explicit function of time. 
To prove the stability of the robotic manipulator, we use passivity analysis \citep{albu2007unified,ortega2008control,ortega2013passivity,keppler2016passivity,haddadin2024unified}.

Assume the robot is interacting with a passive environment. 
Let $\mathcal{V}(\mathbf{q}, \dot{\mathbf{q}}, t)$ (Equation \eqref{eq:lyapunov_function_robot}) be the storage function of the robot. 
The time derivative of this storage function is given by:
\begin{align*}
    \frac{d}{dt}\mathcal{V}(\mathbf{q}, \dot{\mathbf{q}}) = - \dot{\mathbf{q}}^{\top}\mathbf{B}_q \dot{\mathbf{q}} + \frac{\partial \mathcal{U}}{\partial t}(\mathbf{q}, t)
\end{align*}
In this equation, \( \frac{\partial \mathcal{U}}{\partial t}(\mathbf{q}, t) \) represents the change in total virtual elastic potential energy due to time-varying module parameters, e.g., time-varying mechanical impedances, time-varying virtual trajectories. 
If the joint-space damping matrix \( \mathbf{B}_{q} \) is sufficiently large relative to \( \frac{\partial \mathcal{U}}{\partial t}(\mathbf{q}, t) \), the time derivative of the storage function can be made negative semi-definite, resulting in a passive control system for the robot.

\begin{figure*}
    \centering
    \includegraphics[trim={0.0cm 3.0cm 0.0cm 0.0cm}, width=0.99\textwidth, clip, page=1]{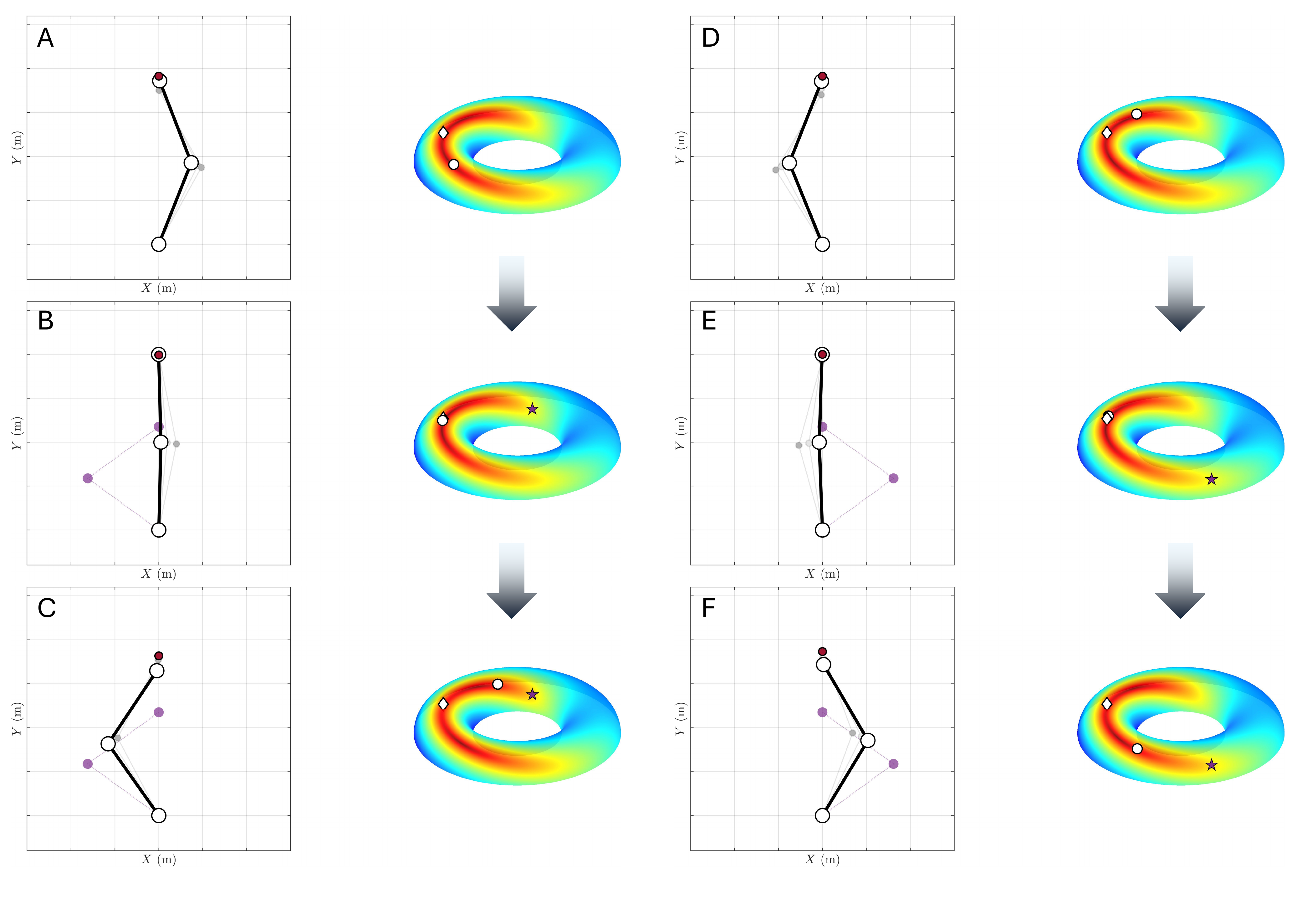}
    \caption{ A two degrees-of-freedom planar robotic manipulator controlled using Equation \eqref{eq:two_modules_superimposed}, and its configuration manifold which is a $T^2$ torus. (A–C) (respectively (D-F)) The robot passing through kinematic singularity (i.e., the straight-arm posture) to change from right-hand (respectively left-hand) to left-hand (respectively right-hand) configuration. In the planar robot diagram, red markers depict $\mathbf{p}_0$ for task-space position impedance $\mathbf{Z}_p$ (Section \ref{subsec:module_for_task_space_position}). In (B, C) and (E, F), the purple robot configurations depict the virtual left-hand $\mathbf{q}_{0,L} = [0.2\pi, 0.6\pi]$ and right-hand $\mathbf{q}_{0,R} = [0.8\pi, -0.6\pi]$ configurations, respectively. On the torus, the potential energy $U_q + U_p \circ \mathbf{h}_p$ is plotted, where red indicates lower values. Circle markers depict the robot’s current configuration; diamond markers depict the singular configuration (i.e., the straight-arm posture). In (B, C) and (E, F), purple star markers depict the virtual joint configurations $\mathbf{q}_{0,L}$ and $\mathbf{q}_{0,R}$, respectively. Parameters of the impedance modules: $\mathbf{K}_p=60\mathbb{I}_2$, $\mathbf{B}_p=20\mathbb{I}_2$, $\mathbf{K}_q=2\mathbb{I}_2$. Code script used for MuJoCo simulation: \texttt{2DOF\_singularity.py}. MATLAB script used for visualization: \texttt{main\_2DOF\_singularity.m}. }
    \label{fig:kinematic_singularity_2DOF}
\end{figure*}

\section{Modular Robot Control: Applications}\label{sec:applications}
In this Section, applications of the modular robot control approach are provided.
Examples in both simulation and real robot implementations are presented.
For the simulation, MuJoCo Python robotic simulator was used \citep{todorov2012mujoco}.
For the real robot implementation, a seven degrees-of-freedom KUKA LBR iiwa14 was used. 
For the control of iiwa14, KUKA's Fast Robot Interface (FRI) was employed. 
The Forward Kinematic map of iiwa14 to derive $\mathbf{p}$, $\mathbf{R}$, and the Jacobian matrices $\mathbf{J}_p(\mathbf{q})$, $\mathbf{J}_r(\mathbf{q})$ were calculated using the Exp[licit]\textsuperscript{TM}-FRI Library \citep{lachner2024exp}. 
For visualization of the robot, MATLAB was used. 

All codes are available in the following Github repository: \url{https://github.com/mosesnah-shared/ModularRobotControl}.
In detail:
\begin{itemize}
    \item MuJoCo simulation: \texttt{MuJoCoApplications}
    \item Control of iiwa14: \texttt{KUKARobotApplications}
    \item MATLAB visualization: \texttt{MATLABApplications}
\end{itemize}
For all of the applications, an open-chain $n$ degrees of freedom robotic manipulator with ideal torque actuators was assumed. 
Moreover, gravitational force $\mathbf{g}(\mathbf{q}(t))$ (Equation \eqref{eq:manipulator_equation}) was assumed to be compensated by the controller and was neglected. 
For MuJoCo simulation, the environment's gravitational acceleration was set to be zero. 
For the control of iiwa14, the built-in gravity compensation was activated. 

\subsection{Managing Kinematic Singularity}\label{subsec:managing_kinematic_singularity}
In this application, we demonstrate that the problem of kinematic singularities can be effectively resolved using our modular approach.
Examples are provided through both MuJoCo simulations with a planar robot (Section \ref{subsubsec:planar_robot_kinematic_singularity}) and real-world control experiments using the KUKA iiwa14 (Section \ref{subsubsec:iiwa14_singularity}). 
We show that not only the problem of kinematic singularity, but also the problem of kinematic redundancy can be resolved using the same controller (Section \ref{subsubsec:kinematic_singularity_w_redundancy}).  
Finally, we present a robot demonstration illustrating that kinematic singularities can be exploited rather than avoided (Section \ref{subsubsec:exploiting_kinematic_singularity}).

\begin{figure*}
    \centering
    \includegraphics[trim={0.0cm 0.0cm 0.0cm 0.0cm}, width=0.99\textwidth, clip, page=1]{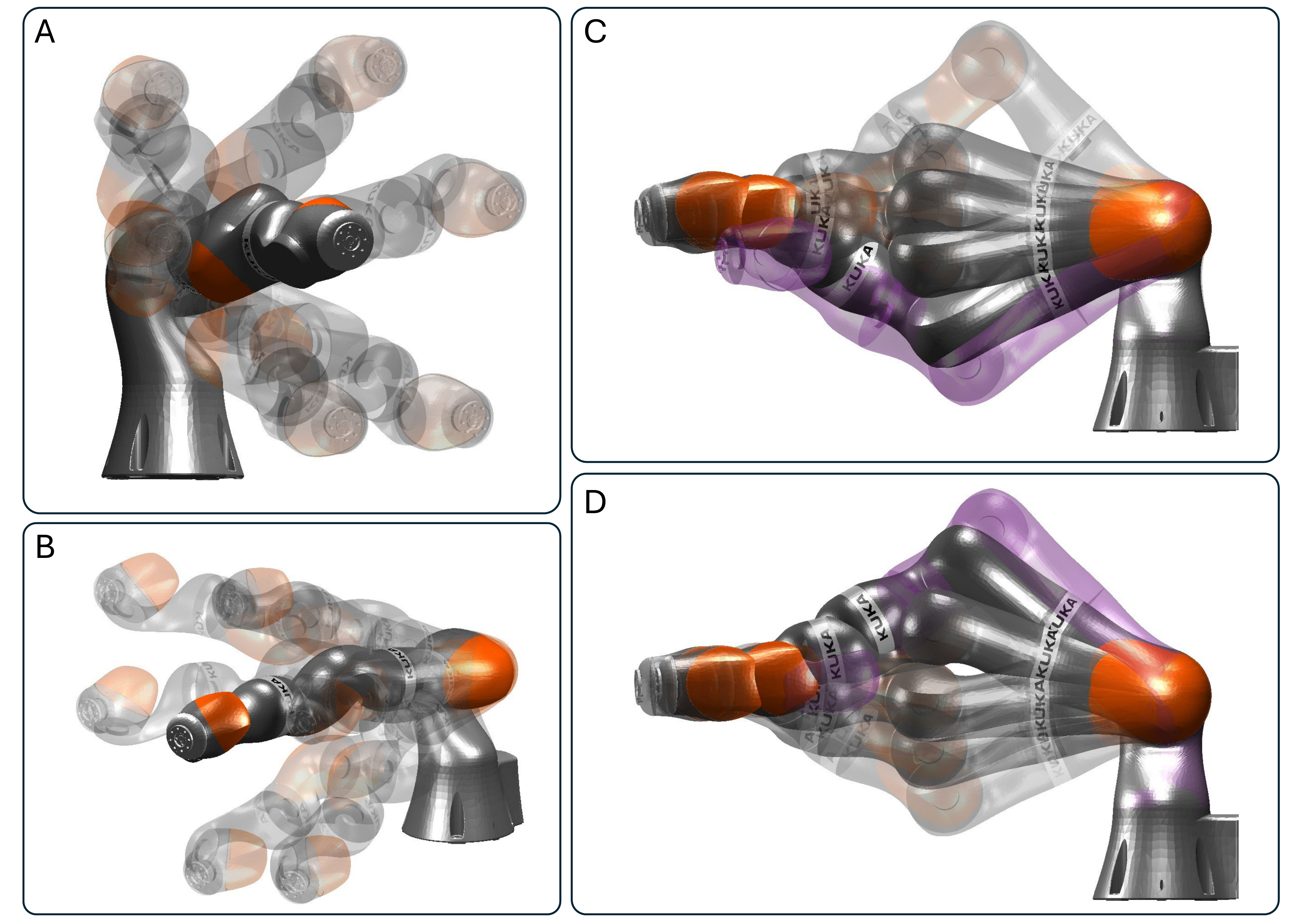}
    \caption{ KUKA iiwa14 robotic manipulator controlled using Equation \eqref{eq:three_modules_superimposed}. (A,B) Using the three control modules, the whole robot's workspace can be utilized. For the experiment, \texttt{iiwa14\_singularity1} KUKA application was used. (C,D) Using the three control modules, the robot can seamlessly pass through singular configuration to change between ``up-hand'' and ``down-hand'' configurations. The purple robots depict the virtual (C) down-hand $\mathbf{q}_{0,D} =[-0.06,   0.81,0.31,-1.52,-0.09,-0.66, 0.00]$rad and (D) up-hand $\mathbf{q}_{0,U} = [0.06,   2.12, -0.28,   1.15,  0.20,  0.53,  0.24]$rad configurations, respectively. For the experiment, \texttt{iiwa14\_singularity2} was used. The experimental data was visualized in MATLAB using \texttt{main\_iiwa14\_singularity\_visualize.m}. }
    \label{fig:kinematic_singularity_iiwa14_through_singularity}
\end{figure*}

\subsubsection{A Planar Robot}\label{subsubsec:planar_robot_kinematic_singularity}
Consider a planar two degrees-of-freedom robotic manipulator with ideal torque actuators. 
To control the robot, two modules were used: joint-space impedance (Equation \eqref{eq:joint_space_module}) and task-space impedance for position (Equation \eqref{eq:task_space_module_position}):
\begin{equation}\label{eq:two_modules_superimposed}
    \bm{\tau}_{in}(t) = \mathbf{Z}_q(\mathbf{q},\mathbf{q}_0) + \mathbf{Z}_p(\mathbf{p},\mathbf{p}_0)
\end{equation}
As shown in \cite{takegaki1981new} and \cite{nah2024robot}, for a robot without kinematic redundancy, using task-space module $\mathbf{Z}_p, \mathbf{p}_0$ with a symmetric positive-definite joint damping matrix $\mathbf{B}_q$ for joint-space module $\mathbf{Z}_q$ is sufficient to achieve goal-directed discrete movement in task-space.
Nevertheless, by additionally using a symmetric positive-definite joint stiffness matrix $\mathbf{K}_q$ for $\mathbf{Z}_q$, the robot can seamlessly go in and out of kinematic singularity without getting stuck in a singular configuration. 
Moreover, smooth transitions between different robot configurations (e.g. ``left-hand'' vs. ``right-hand'') can be achieved.

A result using the two control modules is presented in Figure \ref{fig:kinematic_singularity_2DOF}. 
The code script used for the MuJoCo simulation was \texttt{2DOF\_singularity.py}. 
For $\mathbf{p}_0$, a minimum-jerk trajectory was used to go into and out of singular configuration \citep{flash1987control,nah2024robot}, although any discrete trajectories can be used. 
Since the controller which consists of two modules does not require solving Inverse Kinematics, the robot can seamlessly go in and out of singularity configuration while achieving task-space control for position. 
The robot avoids getting stuck in singular configurations, due to the joint-space module $\mathbf{Z}_q$ with a symmetric and positive-definite joint stiffness matrix $\mathbf{K}_q$

\begin{figure*}
    \centering
    \includegraphics[trim={0.0cm 1.0cm 0.0cm 0.0cm}, width=0.99\textwidth, clip, page=1]{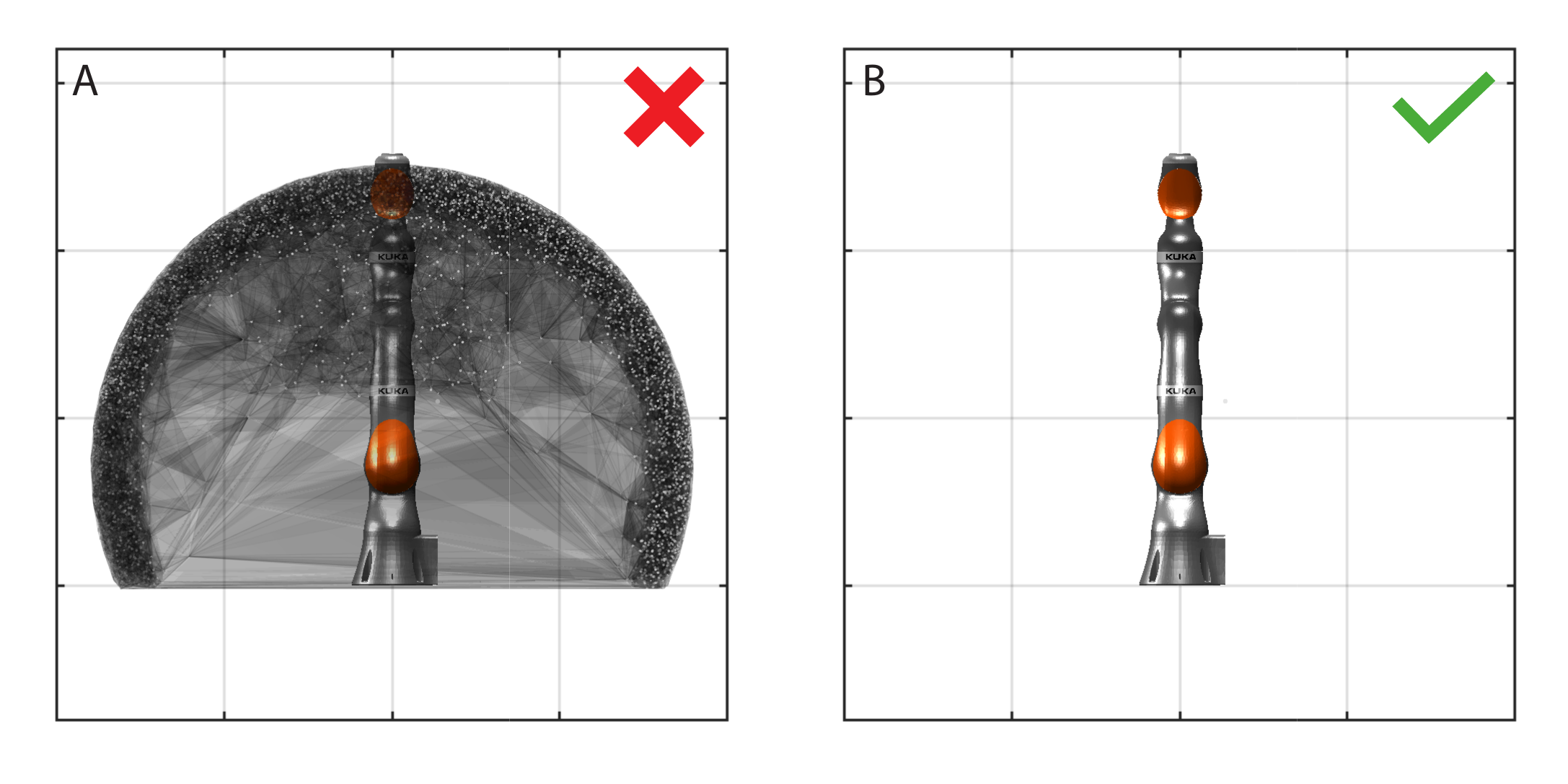}
    \caption{ Analysis and quantification of kinematic singularity of the KUKA LBR iiwa14 robotic manipulator.
    (A) The robot’s workspace regions that were inaccessible using conventional methods \citep{khatib1987unified,chiaverini1997singularity}. Dots depict workspace locations where the singular value of matrix $\bm{\Lambda}^{-1}(\mathbf{q})$ is less than or equal to 0.03 \citep{lachner2020influence}, where $\bm{\Lambda}^{-1}(\mathbf{q}) = \mathbf{J}(\mathbf{q})\mathbf{M}^{-1}(\mathbf{q})\mathbf{J}^{\top}(\mathbf{q}) \in \mathbb{R}^{6\times6}$. Based on this threshold value, 30\% of the robot's workspace is unavailable. For visualization, points that meet the threshold but were less than 0.1m apart were excluded. A Delaunay triangulation algorithm was applied to tessellate the remaining points. (B) Using the proposed modular approach, which allows the robot to seamlessly go in and out of singular configuration, the entire workspace of the robot becomes accessible. MATLAB script used for computation and visualization: \texttt{main\_iiwa14\_singularity\_quantify.m}. For the analysis and quantification, the first and last joints of the iiwa14 were fixed at zero. For each of the remaining five joints, 30 equally spaced sample points were generated between their respective minimum and maximum joint limits. The percentage of singular configurations was calculated by the ratio of sample points meeting the threshold to the total number of sampled points.}
    \label{fig:kinematic_singularity_iiwa14_workspace}
\end{figure*}

Intuitively, the two modules shape the (virtual) elastic potential field $U_q + U_p \circ \mathbf{h}_p$ through their parameters—--namely, the joint $\mathbf{K}_q$ and translational stiffness matrices $\mathbf{K}_p$, and the virtual trajectories to which they are connected--—to control the robot's configuration (Figure \ref{fig:kinematic_singularity_2DOF}).
By controlling the robot’s (virtual) elastic potential energy rather than its motion (Section \ref{subsec:energy_rather_than_motion}), the robot is completely free from the problem of kinematic singularities, as the potential energy $U_q + U_p \circ \mathbf{h}_p$ is well-defined over the whole joint configuration space, including singular configurations.
By \textit{independently} adjusting the stiffness matrices (or weights) between the two potential energies $U_q$ and $U_p \circ \mathbf{h}_p$, one can control which module has greater influence over the robot's behavior.
Finally, since the robot's total energy is the summation of kinetic and potential energy $U_q + U_p \circ \mathbf{h}_p$, passivity can be ensured with a sufficiently high joint-space damping (Section \ref{appx:stability_proof}).  
Hence, passivity of the robot is guaranteed while maintaining stability near (or even at) singular configurations.

Unless $\mathbf{p}_0=\mathbf{h}_{p}(\mathbf{q}_0)$, task conflict exists and the eventual robot configuration will be neither at $\mathbf{p}_0$ nor $\mathbf{q}_0$. 
Nevertheless, if one uses a symmetric positive-definite joint-space damping matrix $\mathbf{B}_{q}$, the robot asymptotically converges to a (local) minimum of potential field $U_q+U_p\circ \mathbf{h}_p$ (Section \ref{appx:stability_proof}). 
To avoid task conflicts, one can use null-space projection methods \citep{ott2015prioritized}; however, such approaches violate passivity \citep{lachner2022geometric} and require additional methods to handle kinematic singularities \citep{buss2005selectively}.

\subsubsection{KUKA LBR iiwa14}\label{subsubsec:iiwa14_singularity}
To control iiwa14, three modules were used: joint-space impedance (Equation \eqref{eq:joint_space_module}), task-space impedance for position (Equation \eqref{eq:task_space_module_position}), and task-space impedance for orientation, either using spatial rotation matrices (Equation \eqref{eq:module_for_SO3_complex}) or unit quaternions (Equation \eqref{eq:impedance_control_quaternion}):
\begin{equation}\label{eq:three_modules_superimposed}
    \bm{\tau}_{in}(t) = \mathbf{Z}_q(\mathbf{q},\mathbf{q}_0) + \mathbf{Z}_p(\mathbf{p},\mathbf{p}_0) + \mathbf{Z}_r(\mathbf{R},\mathbf{R}_0) 
\end{equation}
As shown in \cite{fasse1997spatial} and \cite{nah2024robot}, using task-space modules for position ($\mathbf{Z}_p$, $\mathbf{p}_0$) and orientation ($\mathbf{Z}_r$, $\mathbf{R}_0$), together with a symmetric positive-definite joint damping matrix $\mathbf{B}_q$ for the joint-space module $\mathbf{Z}_q$, is sufficient to achieve goal-directed discrete movement in task space, even with a kinematically redundant robot.
Nevertheless, the robot can potentially get stuck at singular configuration \citep{abu2024unified}.
To address this problem, as shown in Section \ref{subsubsec:planar_robot_kinematic_singularity} using a planar robot, a joint-space control module $(\mathbf{Z}_q, \mathbf{q}_0)$ with a symmetric positive-definite stiffness matrix $\mathbf{K}_q$ can be superimposed, allowing the robot to smoothly go in and out of singular configurations.
As a result, the robot can utilize its entire workspace.
Since the total energy of the robot is a summation of kinetic and potential energies, $U_q + U_p \circ \mathbf{h}_p + U_r \circ \mathbf{h}_r$, passivity can be ensured with sufficiently high joint-space damping (Section \ref{appx:stability_proof}).

A result using the three control modules is presented in Figure \ref{fig:kinematic_singularity_iiwa14_through_singularity}.
The codes used to control iiwa14 were \texttt{iiwa14\_singularity1} (Figure \ref{fig:kinematic_singularity_iiwa14_through_singularity}A, \ref{fig:kinematic_singularity_iiwa14_through_singularity}B) and \texttt{iiwa14\_singularity2} (Figure \ref{fig:kinematic_singularity_iiwa14_through_singularity}C, \ref{fig:kinematic_singularity_iiwa14_through_singularity}D).
The virtual spatial orientation $\mathbf{R}_0$ (or $\vec{\mathbf{q}}_0$) was kept constant, and a minimum-jerk trajectory was used for the virtual task-space position $\mathbf{p}_0$, although any discrete movement could be used.
Because the robot can seamlessly go in and out of singular configurations, the robot's entire workspace can be used (Figure \ref{fig:kinematic_singularity_iiwa14_through_singularity}A, \ref{fig:kinematic_singularity_iiwa14_through_singularity}B). 
By superimposing a joint-space module $(\mathbf{Z}_q, \mathbf{q}_0)$ with a symmetric positive-definite joint stiffness matrix $\mathbf{K}_q$, the robot's configuration can smoothly change between the ``up-hand'' and ``down-hand'' configurations (Figure \ref{fig:kinematic_singularity_iiwa14_through_singularity}C, \ref{fig:kinematic_singularity_iiwa14_through_singularity}D).

As shown in Figure \ref{fig:kinematic_singularity_iiwa14_workspace}, using our modular robot controller, 30\% of the robot's workspace which was previously inaccessible using conventional methods \citep{chiaverini1997singularity} becomes available.

\begin{figure*}
    \centering
    \includegraphics[trim={0.0cm 0.0cm 0.0cm 0.0cm}, width=1.00\textwidth, clip, page=1]{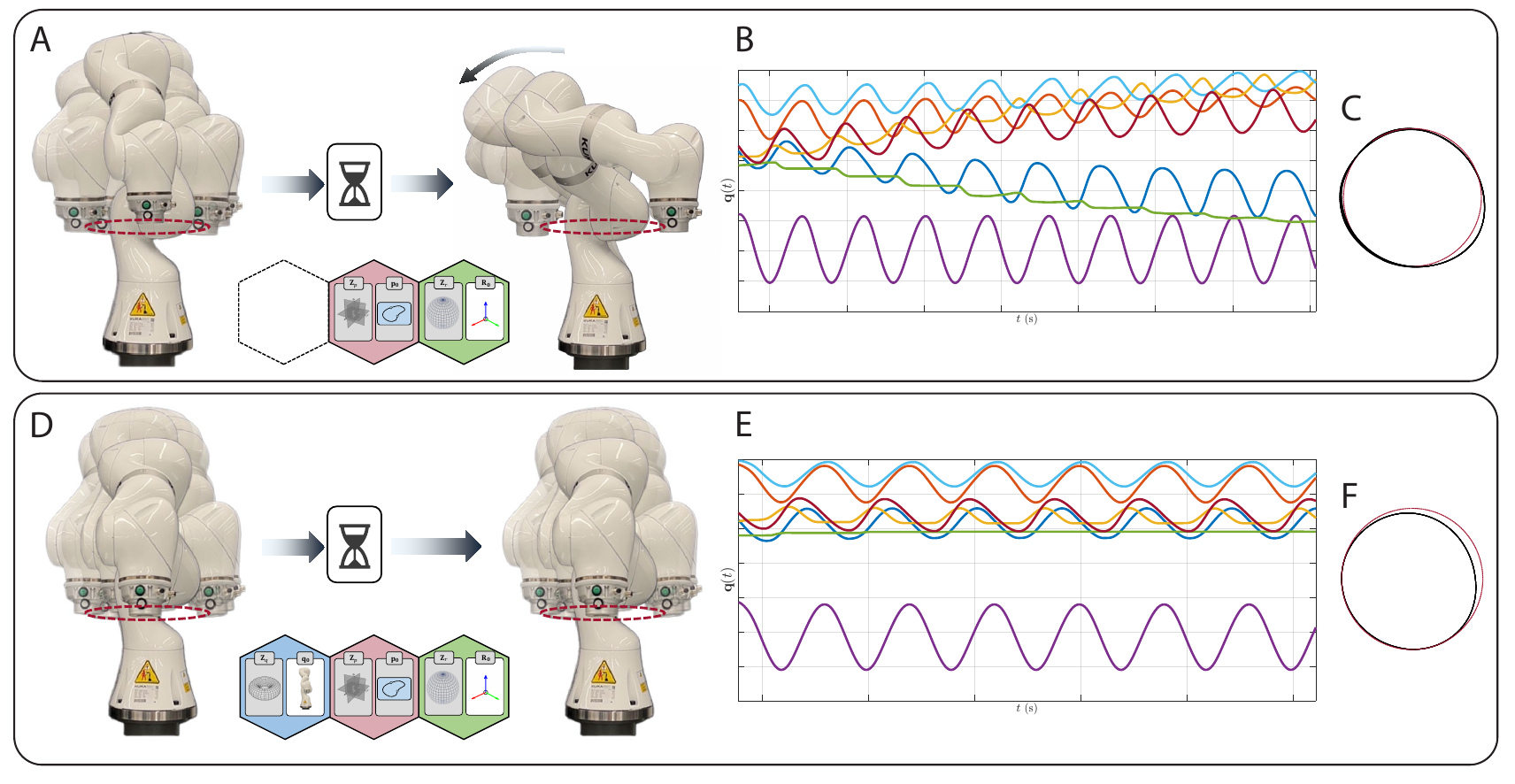}
    \caption{ KUKA iiwa14 robotic manipulator controlled using Equation \eqref{eq:three_modules_superimposed}, with the task of maintaining the end-effector's orientation while following a circular trajectory in task-space. (A,B,C) Result using a zero joint-stiffness matrix $\mathbf{K}_q=\mathbf{0}$. (D,E,F) Result using a symmetric and positive-definite joint-stiffness matrix $\mathbf{K}_q\succ \mathbf{0}$. (B) and (E) show time $t$ vs. joint trajectories $\mathbf{q}(t)$ of iiwa14. (C) and (F) show virtual task-space trajectory for position $\mathbf{p}_0(t)$ (dotted red line) and the actual end-effector position $\mathbf{p}(t)$. \hbg{Module parameters: $\mathbf{K}_p=1600\mathbb{I}_3$, $\mathbf{B}_p=120\mathbb{I}_3$, $\mathbf{K}_r=70\mathbb{I}_3$, $\mathbf{B}_r=5\mathbb{I}_3$, $\mathbf{B}_q=4.5\mathbb{I}_7$. For (D,E,F), $\mathbf{K}_q=6.0\mathbb{I}_7$. For the circular trajectory, radius and period were 0.15m and 4s, respectively.} The code used to control iiwa14 was \texttt{iiwa14\_singularity\_w\_redundancy}. MATLAB code for visualization: \texttt{main\_iiwa14\_singularity\_w\_redundancy.m}. }
    \label{fig:kinematic_singularity_w_redundancy}
\end{figure*}

\subsubsection{Managing Kinematic Redundancy}\label{subsubsec:kinematic_singularity_w_redundancy}
By using the controller composed of three modules (Equation \eqref{eq:three_modules_superimposed}), both kinematic singularity and kinematic redundancy are addressed simultaneously using the same controller. 
Note that these two are mathematically distinct problems: the problem of kinematic singularity arises from the rank drop of the Jacobian matrix, whereas the problem of kinematic redundancy arises from inverting a wide (i.e., more columns than rows) Jacobian matrix. 

As discussed in \cite{mussa1991integrable}, the problem of kinematic redundancy includes the problem of joint drift during the execution of repeatable tasks in task space. 
Joint drift is undesirable since it may cause the robot to reach joint limits or other unwanted states. 
While joint limit avoidance can be achieved using additional approaches \citep{munoz2018operational}, our modular robot controller offers an alternative approach that not only addresses the problem of joint drift but also resolves kinematic singularities and preserves passivity of the robot.

\begin{figure*}
    \centering
    \includegraphics[trim={0.0cm 0.0cm 0.0cm 0.0cm}, width=1.00\textwidth, clip, page=1]{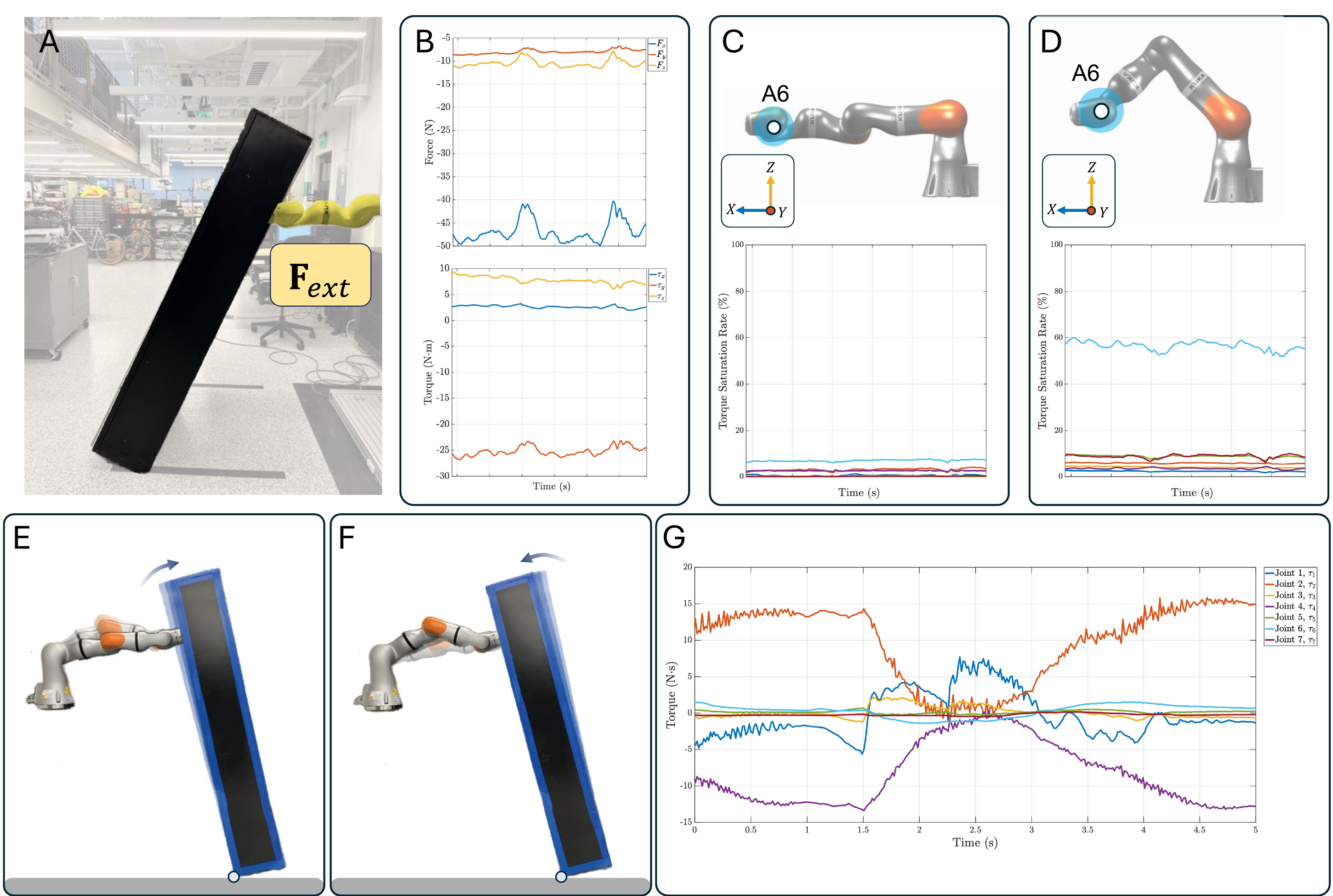}
    \caption{ \textbf{(A)} The KUKA LBR iiwa14 maintains an external wrench exerted by a 31 kg heavy-load bookshelf. \textbf{(B)} Measured force (top) and torque (bottom) from the force/torque sensor. An ATI Industrial Automation Gamma force/torque sensor was used. \textbf{(C)} Torque 
    as a percentage of saturation near a singular configuration. \textbf{(D)} Torque as a percentage of saturation away from a singular configuration. While in (C) torque remains below 10\% of saturation, it reaches nearly 60\% for the A6 joint in (D). The maximum torque limits for the iiwa14’s seven joints are: $\tau_{1,\text{max}} = \tau_{2,\text{max}} = 320\,\mathrm{N{\cdot}m}$, $\tau_{3,\text{max}} = \tau_{4,\text{max}} = 176\,\mathrm{N{\cdot}m}$, $\tau_{5,\text{max}} = 110\,\mathrm{N{\cdot}m}$, $\tau_{6,\text{max}} = \tau_{7,\text{max}} = 40\,\mathrm{N{\cdot}m}$. \hb{\textbf{(E)} The KUKA LBR iiwa14 lifting the heavy-load bookshelf up, and \textbf{(F)} lowering it down near singular configuration. \textbf{(G)} Time $t$ vs. Joint torque $\bm{\tau}_{ext}=\mathbf{J}^{\top}(\mathbf{q})\mathbf{F}_{ext}$ due to the external force from the bookshelf.
    The robot entered and emerged from a singularity at time = 2.5s. Video for the robot demonstration: \url{https://youtu.be/o58iXV63DCU}.
    }}
    \label{fig:exploit_singularity}
\end{figure*}

A result using iiwa14 is shown in Figure \ref{fig:kinematic_singularity_w_redundancy}. 
The code used to control iiwa14 was \texttt{iiwa14\_singularity\_w\_redundancy}. 
The controller without a symmetric and positive-definite joint matrix $\mathbf{K}_q=\mathbf{0}$ resulted in non-negligible joint drift over time (Figure \ref{fig:kinematic_singularity_w_redundancy}A, \ref{fig:kinematic_singularity_w_redundancy}B). 
In contrast, using a symmetric and positive-definite joint matrix $\mathbf{K}_q\succ \mathbf{0}$ effectively eliminated joint drift (Figure \ref{fig:kinematic_singularity_w_redundancy}D, \ref{fig:kinematic_singularity_w_redundancy}E). 
Although the joint drift was eliminated, this improvement came at the expense of reduced tracking performance in task space due to task conflicts (Figure \ref{fig:kinematic_singularity_w_redundancy}C, \ref{fig:kinematic_singularity_w_redundancy}F). 
Task conflict can be avoided through the use of null-space projection. 
But again, this approach comes at the expense of violating the robot's passivity \citep{lachner2022geometric}. 
Moreover, as discussed in \citep{hermus2021exploiting}, the modular controller can exploit kinematic redundancy to achieve improved tracking performance.

\subsubsection{Exploiting Kinematic Singularity}\label{subsubsec:exploiting_kinematic_singularity}
Kinematic singularities can be exploited rather than avoided. 
As the modular approach allows seamless transitions into and out of singular configurations, we demonstrate that entering singular configurations can offer practical benefits. 
In detail, given an external wrench applied to the robot $\mathbf{F}_{ext}\in\mathbb{R}^{6}$, the resulting torque due to external wrench is $\bm{\tau}_{ext}=\mathbf{J}^{\top}(\mathbf{q})\mathbf{F}_{ext}$. 
Hence, high external load can be maintained with low joint-torque actuation $\bm{\tau}_{in}(t)$.

\begin{figure*}
    \centering
    \includegraphics[trim={0.0cm 0.0cm 0.0cm 0.0cm}, width=1.00\textwidth, clip, page=1]{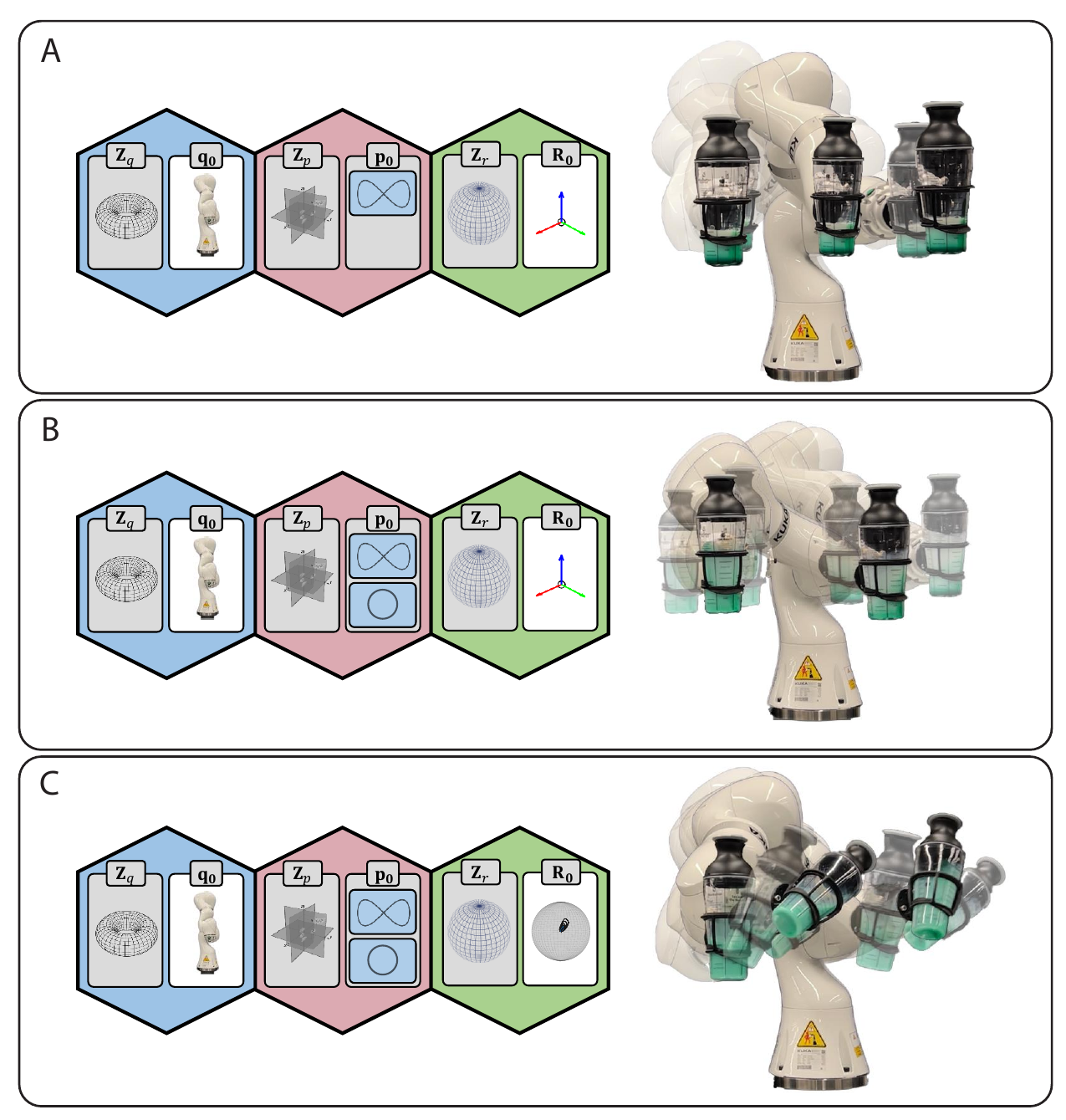}
    \caption{KUKA iiwa14 robotic manipulator shaking a cocktail. The robot was controlled using Equation \eqref{eq:three_modules_superimposed}, and Imitation Learning was used to learn the virtual task-space position $\mathbf{p}_0(t)$ (Appendix \ref{subsubsubsec:imit_learning_task_position}) and orientation $\mathbf{R}_0(t)$ (or $\vec{\mathbf{q}}_0(t)$) (Appendix \ref{subsubsubsec:imit_learning_task_orientation}). (A) Robot movement generated by defining a virtual trajectory $\mathbf{p}_0(t)$ in the shape of a figure-eight, while maintaining a fixed end-effector orientation. (B) Robot movement generated by defining a virtual trajectory $\mathbf{p}_0(t)$ as a summation of figure-eight and a circular trajectory, while maintaining a fixed end-effector orientation. The summation of virtual trajectory was conducted via the superposition principle of virtual trajectories (Section \ref{subsubsubsec:superposition_principle_of_trajectories}). (C) Robot movement generated by defining a virtual trajectory $\mathbf{p}_0(t)$ used in (B) with a shaking motion for $\mathbf{R}_0(t)$ (or $\vec{\mathbf{q}}_0(t)$). For the shaking motion, Imitation Learning with data collected from human demonstration was used. Code used to control iiwa14 was \texttt{iiwa14\_cocktail\_shaking}. MATLAB code for visualization and Imitation Learning: \texttt{main\_cocktail\_shaking.m}. Module parameters: $\mathbf{K}_p=600\mathbb{I}_3$, $\mathbf{B}_p=40\mathbb{I}_3$, $\mathbf{K}_r=70\mathbb{I}_3$, $\mathbf{B}_r=5\mathbb{I}_3$, $\mathbf{K}_q=6\mathbb{I}_7$, $\mathbf{B}_q=4.5\mathbb{I}_7$. }
    \label{fig:cocktail_shaking}
\end{figure*}

A simple robotic experiment was conducted to demonstrate the effectiveness of the proposed approach. The task involved stabilizing a heavy bookshelf weighing 31 kg (Figure \ref{fig:exploit_singularity}A). The joint torques near singular configurations (Figure \ref{fig:exploit_singularity}C) were lower compared to operation away from singularities (Figure \ref{fig:exploit_singularity}D). This result shows that operating near singular configurations allows the robot to handle high external loads while simultaneously reducing motor current consumption, thereby enhancing the overall energy efficiency of the controller. 
\hb{Moreover, since the projected joint torque due to external force is small near singular configuration (Figure \ref{fig:exploit_singularity}E, \ref{fig:exploit_singularity}F, \ref{fig:exploit_singularity}G), lifting or lowering a heavy external load can be facilitated. }

Similar findings were reported by \cite{faraji2017singularity}, who exploited kinematic singularities to maintain an upright posture in a humanoid robot with reduced torque requirements. However, unlike the optimization-based inverse kinematics approach used by \cite{faraji2017singularity}, our modular robot controller achieves this capability without requiring \hb{Inverse Kinematics} computations, thereby preserving the robot's passivity.

\begin{figure*}
    \centering
    \includegraphics[trim={0.0cm 0.0cm 0.0cm 0.0cm}, width=0.90\textwidth, clip, page=1]{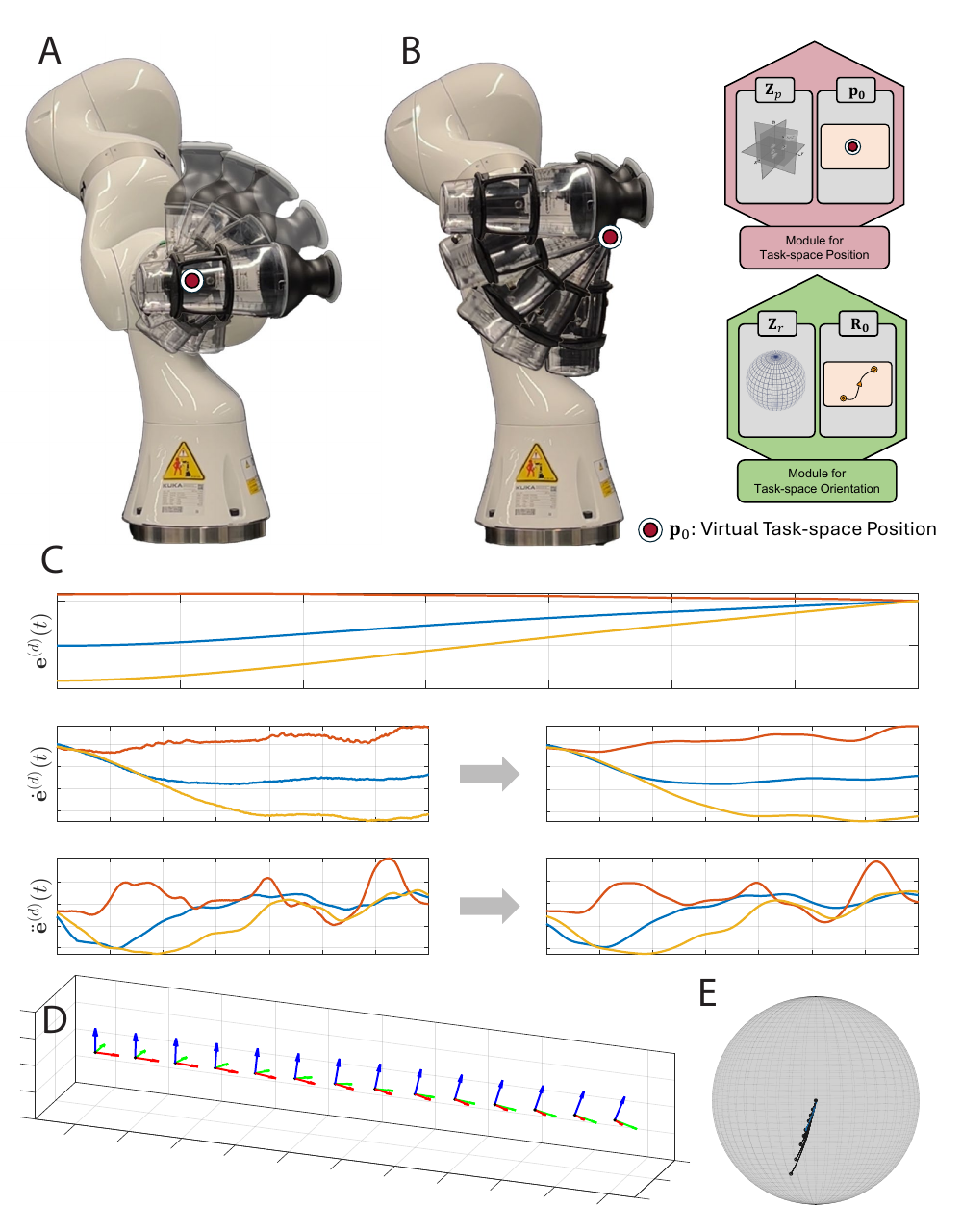}
    \caption{ The KUKA iiwa14 robotic manipulator performing a pouring motion. Robot movements resulting from defining a fixed virtual trajectory $\mathbf{p}_0$ and the task-space position $\mathbf{p}$ are illustrated for two cases: (A) defined at the robot's end-effector and (B) defined at the tip of the bottle. To generate the pouring motion, Imitation Learning was used (Section \ref{subsubsubsec:imit_learning_task_orientation}). (C) The exponential coordinates $\mathbf{e}^{(d)}(t)$ of the pouring motion collected by human demonstration. Gaussian smoothing via MATLAB's \texttt{smoothdata} function was used to the first-order $\dot{\mathbf{e}}^{(d)}(t)$ and second-order derivatives $\ddot{\mathbf{e}}^{(d)}(t)$ for denoising the data. (D) The learned pouring motion represented using an orthonormal reference frame $\mathbf{R}_0 \in \text{SO}(3)$. The trajectory's temporal evolution is depicted through an offset between successive frames. (E) The learned trajectory depicted on the SO(3) manifold, which is a three-dimensional sphere with radius $\pi$ \citep{park1995distance}. Utilizing exponential coordinates $\mathbf{e}^{(d)}(t)$ for Imitation Learning provides the spatial invariance property for trajectories on SO(3). The code used to control iiwa14 was \texttt{iiwa14\_cocktail\_pour}. MATLAB code for visualization and Imitation Learning: \texttt{main\_cocktail\_pour.m}. Module parameters: $\mathbf{K}_p=600\mathbb{I}_3$, $\mathbf{B}_p=40\mathbb{I}_3$, $\mathbf{K}_r=70\mathbb{I}_3$, $\mathbf{B}_r=5\mathbb{I}_3$, $\mathbf{K}_q=6\mathbb{I}_7$, $\mathbf{B}_q=4.5\mathbb{I}_7$. }
    \label{fig:cocktail_pouring}
\end{figure*}

\subsection{Modular Imitation Learning}\label{subsec:modular_imitation_learning}
In this section, we build upon the work presented in Section 4.2 of \cite{nah2024robot} and demonstrate that combining the strengths of EDA and DMP can significantly simplify task-space control, offering a distinct advantage for robot programming. 
This approach enables the separate learning of position and orientation trajectories, which can then be linearly combined through the superposition of mechanical impedances. 
As a result, motions learned in different spaces can be independently acquired and seamlessly combined, facilitating modular Imitation Learning.

To demonstrate modular imitation learning, we present a robotic experiment involving a cocktail-shaking task using the iiwa14 robot. 
This task requires coordinated movements in both task-space position (Appendix \ref{subsubsubsec:imit_learning_task_position}) and orientation (Appendix \ref{subsubsubsec:imit_learning_task_orientation}). 
Additionally, since the movements are repetitive in task-space, the issue of joint drift caused by kinematic redundancy must be addressed.

The resulting robot demonstration is shown in Figure \ref{fig:cocktail_shaking}. 
The control code used for iiwa14 was \texttt{iiwa14\_cocktail\_shaking}. 
Imitation Learning was used to learn both the task-space position and orientation. 
For the Imitation Learning, MATLAB script \texttt{main\_cocktail\_shake.m} was used. 

As illustrated in Figure \ref{fig:cocktail_shaking}, the movements for task-space position and orientation can be learned separately and combined at the level of the virtual trajectory (Section \ref{subsubsubsec:superposition_principle_of_trajectories}). 
By leveraging mechanical impedances, the final command is linearly combined at the joint-torque level for both position and orientation control. This enables independent movement planning (Section \ref{subsubsec:modularity_independence}) while also guaranteeing passivity by regulating the robot's total elastic potential energy, \( U_q + U_p \circ \mathbf{h}_p + U_r \circ \mathbf{h}_r \) (Section \ref{subsubsec:closure_of_stability}) with sufficiently large joint damping (Section \ref{appx:stability_proof}).

\subsubsection{Virtual Trajectory Defined outside the Robot's Body}\label{subsubsec:virtual_trajectory_outside_robot}
As discussed in Section \ref{subsec:module_for_task_space_position}, the virtual trajectory $\mathbf{p}_0$ and the actual task-space position $\mathbf{p}$ used by the module for task-space position are not required to coincide with the robot's end-effector.
This feature is particularly beneficial for tasks that require stabilizing an external point outside the robot, such as pouring liquid from a bottle.

Experimental results from a liquid pouring task are presented in Figure \ref{fig:cocktail_pouring}. 
The code used to control iiwa14 was \texttt{iiwa14\_cocktail\_pour}. 
Imitation Learning for task-space orientation (Appendix \ref{subsubsubsec:imit_learning_task_orientation}) was used to learn the pouring motion (Figure \ref{fig:cocktail_pouring}C, \ref{fig:cocktail_pouring}D, \ref{fig:cocktail_pouring}E). 
For the Imitation Learning, MATLAB script \texttt{main\_cocktail\_pour.m} was used. 

As shown in Figure \ref{fig:cocktail_pouring}A, defining a fixed virtual trajectory $\mathbf{p}_0$ and the task-space position $\mathbf{p}$ at the robot's end-effector resulted in excessive movement at the bottle's tip, thereby making it inappropriate to achieve the pouring task. 
To address this problem, one can simply redefine both $\mathbf{p}_0$ and $\mathbf{p}$ to be at the tip of the bottle (Figure \ref{fig:cocktail_pouring}B). 
This example highlights the capability of the proposed modular robot control approach to effectively handle object-centric manipulation tasks through the virtual trajectory. 

To calculate the Forward Kinematics map for points outside the robot's physical body, one can utilize Denavit–Hartenberg parameters \citep{featherstone2014rigid} or use methods based on the Product-of-Exponentials formula \citep{brockett1983robotic}, which simplifies the calculation through simple matrix algebra \citep{murray1994mathematical,lynch2017modern}. 
\hb{Further details of this appraoch are presented in \citep{lachner2024exp}.}

\section{Discussion and Future Work}\label{sec:discussion_futurework}
\subsection{Motor Primitives as an Account for Biological Motor Behavior}\label{sec:motor_primitives_for_orientation}
In this paper, we demonstrated that modular control using motor primitives simplified a wide range of control tasks, highlighting the versatility and flexibility of this approach in different robotic applications.

It is worth emphasizing that the provided definitions of motor primitives using EDA are not solely for practical robot control but also to account for observable motor behavior of biological systems.
This perspective is consistent with the strict categorization of kinematic primitives to submovements (i.e., discrete movement) and oscillations (i.e., rhythmic movements), even though mathematically rhythmic movement can be described by a combination of discrete movements \citep{hogan2007rhythmic}. 
This is also why we intentionally excluded movements involving spatial orientation from the definition of submovements, although the topic has been extensively studied from a mathematical perspective \citep{murray1994mathematical,bullo1995proportional,park1995zier,park1997smooth,lynch2017modern}. 
To date, there is insufficient data on how humans manage movements requiring control of spatial orientation.

As the definition of EDA also accounts for observable behavior of biological systems, the elements of EDA can later be modified or expanded as sufficient datasets become available.
For instance, accounting for the growing evidence that stable posture may be a distinct class of motor primitives \citep{shadmehr2017distinct,jayasinghe2022neural} (Section \ref{subsec:motor_primitives_biological_systems}), an additional fourth element of EDA can be defined. 
Again, care is required to claim that stable posture is simply a special case of submovement with zero amplitude; an account of biological motor behavior may not always provide mathematical brevity. 
Moreover, once the dataset is sufficiently accumulated, the definition of submovements can be extended to account for the movement of spatial orientation.
Since neuromotor control research has been dominated by the point-to-point reaching movements of unimpaired human subjects in $\mathbb{R}^{3}$ space \citep{morasso1981spatial,flash1985coordination,hogan1987moving,krebs1999quantization,sabes2000planning}, studies on movements for spatial orientation (i.e., on the $\text{SO}(3)$ manifold) remain an area for future research.

The level of detail at which the modular framework aims to express is at the level of observational and combinatorial level of biological motor behavior \citep{hogan2012dynamic}. 
Reminiscent of David Marr's categorization of the level of analysis \citep{marr1982vision}, the three levels of analysis proposed by \cite{hogan2012dynamic} are observational, combinatorial, and physiological levels. 
Definitions of motor primitives address the observational level of analysis, focusing on overt and measurable behavior; modularity further emphasizes the combinatorial level of analysis, explaining how motor primitives may be combined to produce complex actions. 
Although we intentionally remain silent on the possible physiological mechanism underlying such motor behavior, accounting for the physiological level of analysis is crucial to delineate the distinct elements of motor primitives. 
While rhythmic movements at the observational and combinatorial level might be composed of a combination of submovements, the neural substrates for discrete and rhythmic movements are strikingly different \citep{schaal2004rhythmic,sternad2019control}.

\subsection{Feasibility is Preferred over Optimality}\label{sec:optimality}
Both in robotics and motor control research, optimal control theory has offered valuable insights into how complex movements can be generated, planned, and executed efficiently.
\citep{bellman1966dynamic,todorov2002optimal,kirk2004optimal,todorov2007optimal,hogan1987moving,schaal2007dynamics,friston2011optimal,karaman2011sampling,berret2011evidence,posa2014direct,polyakov2017affine}. 
The central concept is that the quality of a control policy can be evaluated and refined based on a cost (or reward) function, and the goal is to identify the  controller that optimizes the total cost (or reward). 
In robotics, computational algorithms to identify the optimal control policy have been articulated; Dynamic Programming and Reinforcement Learning algorithms derive an optimal policy by optimizing the accumulated reward over a long-term horizon \citep{bertsekas1996neuro,kaelbling1996reinforcement,sutton1999reinforcement,schaal2007dynamics,theodorou2011iterative,bertsekas2012dynamic}. 
In motor control research, optimal control based on forward-inverse models \citep{miall1993cerebellum,wolpert1995internal,kawato1999internal,shadmehr2008computational,diedrichsen2010coordination}
or stochastic optimal feedback control \citep{todorov2002optimal,todorov2005stochastic,berret2021stochastic} have been proposed as an account of observable motor behavior of biological systems.

In contrast to the perspective of optimality, the presented modular approach was achieved by choosing the parameters that are ``good enough'' to achieve the task. 
Hence, as stated by \cite{billard2022learning}, we follow the spirit of \textit{feasiblity is preferred over optimality}---a principle that reflects the fact that humans do not learn a singular, optimal method for performing control tasks. 
In fact, humans tend to learn multiple strategies to achieve a task, rather than relying on a single optimal approach \citep{tassa2011theory,feix2015grasp,yao2021hand,billard2022learning}.

The emphasis of feasibility over optimality allows for a redundancy of solutions for manipulation, highlighting the adaptability and flexibility in how humans approach and execute various tasks. 
In this perspective, the curse of dimensionality associated with optimal control approaches \citep{schaal2007dynamics} can be effectively managed. 
Prior studies have shown that structures with kinematic redundancy actually provide favorable properties for constrained motion execution \citep{hermus2021exploiting} and for grasping \citep{yao2023exploiting}, suggesting that the extremely high dimensional structure of biological systems may be a ``blessing'' rather than a curse.

\subsection{Energy, not Motion, for Modularity}\label{subsec:energy_rather_than_motion}
Robotics has been dominated by motion control \citep{hogan2022contact}. 
However, motion control approaches introduce several challenges, such as managing kinematic redundancy and singularity, as well as safety against contact and physical interaction. 
Using the presented modular approach, we demonstrated that the challenges associated with motion-based control strategies are effectively addressed. 
The problem of solving Inverse Kinematics is \textit{completely} avoided: the only requirements are the Forward Kinematics map of the robot and the Jacobian transpose matrices. 
Numerical stability near and even at kinematic singularity can be achieved. 
The robot can seamlessly go into and out of kinematic singularity. 
Hence, smooth transitions between different robot configurations (e.g. ``left-hand'' vs. ``right-hand'') are available (Sections \ref{subsubsec:planar_robot_kinematic_singularity} and \ref{subsubsec:iiwa14_singularity}). 
Kinematic redundancy is addressed along with kinematic singularity using the same controller, even though these two are fundamentally separate problems (Section \ref{subsubsec:kinematic_singularity_w_redundancy}). 
Moreover, with an appropriate choice of mechanical impedances, the approach remains robust against contact and physical interaction \citep{lachner2021energy}.
Against passive environments (with constant mechanical impedances and virtual trajectories), passivity of the robot is preserved (Section \ref{appx:stability_proof}). 

The theoretical foundation that enables such simplicity lies in the perspective of ``energy'' for robot control \citep{stramigioli2001modeling,stramigioli2015energy}. 
\hb{In fact, by changing the perspective from motion to energy, modular robot control is achieved. 
Recall the manipulator equation with gravity compensation (Equation \eqref{eq:manipulator_equation}):} 
$$
   \mathbf{M}(\mathbf{q})\ddot{\mathbf{q}} + \mathbf{C}(\mathbf{q}, \dot{\mathbf{q}})\dot{\mathbf{q}} = \bm{\tau}_{in}(t)
$$
The presented modular approach uses the four major modules and their combination to derive the torque command $\bm{\tau}_{in}$ (Section \ref{subsec:basic_control_modules}). 
For the task-space modules, the corresponding virtual elastic potential fields $U_{p}:\mathbb{R}^{3}\rightarrow \mathbb{R}$ and $U_r:\text{SO}(3)\rightarrow \mathbb{R}$ can be defined over joint-space via the Forward Kinematics map $\mathbf{h}_{p}:\mathcal{Q}\rightarrow \mathbb{R}^3$ and $\mathbf{h}_r:\mathcal{Q}\rightarrow \text{SO}(3)$.
These two functions, $U_p\circ \mathbf{h}_p:\mathcal{Q}\rightarrow \mathbb{R}$ and $U_r\circ \mathbf{h}_r:\mathcal{Q}\rightarrow \mathbb{R}$ are the ``pullback'' of $U_p$ and $U_r$ by $\mathbf{h}_p$ and $\mathbf{h}_r$, respectively. 
Additionally accounting for the joint-space module and its virtual potential field $U_{q}:\mathcal{Q}\rightarrow \mathbb{R}$, the superposition principle of mechanical impedances (Section \ref{subsubsubsec:superposition_principle_of_mech_imps}) with gravity compensation is simply a summation of virtual elastic potential energies in joint-space $\mathcal{Q}$:
\[
  \bm{\tau}_{in}(t) \equiv -\frac{\partial( U_{q}+U_{p}\circ \mathbf{h}_{p} + U_{r}\circ \mathbf{h}_{r} )}{\partial \mathbf{q}}(\mathbf{q})
\]
\hb{Each virtual potential energy associated with a module can be independently modified, thereby satisfying the independence property of modularity (Section \ref{subsubsec:modularity_independence}).
Furthermore, the virtual elastic potential fields can be linearly superimposed, even though each potential function and the Forward Kinematics map is nonlinear, consistent with the superposition principle of mechanical impedances (Section \ref{subsubsubsec:superposition_principle_of_mech_imps}).
Moreover, the ``pullback'' operation via the Forward Kinematics map ensures that the resulting controller does not require solving Inverse Kinematics. 
The closure of stability (Section \ref{subsubsec:closure_of_stability}) property of modularity can also be achieved by analyzing the total energy of the robotic manipulator and its dissipation over time.
The total energy of the robot is a summation of virtual elastic potential field $\mathcal{U}(\mathbf{q})\equiv (U_{q}+U_{p}\circ \mathbf{h}_{p} + U_{r}\circ \mathbf{h}_{r})(\mathbf{q})$ and kinetic energy. 
Modulating the mechanical impedances or virtual trajectories controls the total elastic energy $\mathcal{U}(\mathbf{q})$ of the robot. 
The dissipation of the total energy of the robot can be regulated by a symmetric and positive-definite joint-space damping. 
Therefore, against passive environments, passivity of the robot is preserved with appropriate values of the modules. 
}  

Note that such modular property in the perspective of energy may be analogous to the control approach discussed in \cite{ratliff2018riemannian,cheng2020rmp,xie2020geometric} and its variations \citep{ratliff2020optimization,van2022geometric}. 
However, the presented modular approach provides simplicity for task-space control. 
For instance, the Jacobian pseudo-inverse is not required for the presented modular control framework. 
In addition, the difference between Operational Space control and the presented modular approach can also be clarified. 
For Operational Space control \citep{khatib1987unified}, dynamic decoupling via the inertia matrix and the use of null-space projection for kinematically redundant robots violate passivity, which is a crucial requirement for tasks involving contact and physical interaction \citep{stramigioli2015energy}.  
In contrast, the presented modular approach can achieve passivity by modulating $\mathcal{U}$ and the damping matrices $\mathbf{B}_q$, $\mathbf{B}_p$ and $\mathbf{B}_r$ (Section \ref{appx:stability_proof}).

\subsection{Learning the Module Parameters}\label{sec:learning_module_parameters}
A key objective of the proposed modular approach is to select the module parameters, i.e., impedance operator $\mathbf{Z}$ and the virtual trajectory $\mathbf{x}_0$ to which the impedance is connected. 
For the presented examples, constant mechanical impedances with virtual trajectories planned using either DMP or minimum-jerk trajectories were sufficient to achieve the tasks. However, these parameters were typically selected through trial-and-error or provided via human demonstrations for the Imitation Learning of DMP. 
A method to autonomously learn the appropriate module parameters for a given manipulation task remains to be established. 

For learning virtual trajectories, Imitation Learning enables learning various types of trajectories from few human demonstrations. 
However, the method focuses on learning a single trajectory that is provided by human demonstration. 
A method for autonomously combining these learned movements has not yet been fully achieved. 
The complexity of the problem is exacerbated in the presence of unknown environments, dynamic object behaviors, or additional task constraints such as real-time obstacle avoidance.
To address this problem, merging high-level task planning approaches
\citep{hauser2010multi,kaelbling2011hierarchical,holladay2024robust} such as Planning Domain Definition Language (PDDL) \citep{aeronautiques1998pddl,fox2003pddl2,garrett2021integrated} with the presented modular approach which resolves the problem of low-level robot control may be a promising direction for future research.

For learning mechanical impedances, prior approaches often used learning-based approaches that optimized a predefined cost function.
The cost function was often chosen to be the tracking error between the virtual and actual trajectories \citep{buchli2011learning,abu2020variable}. 
The resulting control policy increases the mechanical impedance when a large tracking error occurs. 
While such approaches improve tracking performance, an important required aspect is the ability to autonomously regulate the dynamics of physical interaction.
For example, simply increasing the mechanical impedance may be inappropriate for handling delicate objects, such as glass cups or biological samples. 

From the authors' perspective, integrating sensory data from the environment to autonomously determine the mechanical impedances could be one potential direction for future research. 
In fact, humans also integrate sensory-motor data to execute motor actions effectively \citep{ernst2002humans}.
For instance, using vision data, the robot could evoke low mechanical impedances to adapt to environments with high uncertainties (e.g., visually cluttered or dynamic surroundings).
Furthermore, by combining tactile sensors with vision data, one could allow the robot to learn appropriate mechanical impedances for interacting effectively with the manipulated objects (e.g., grasping a glass cup).

\subsection{Coupling the Task-space Impedance Modules}\label{sec:coupling_impedance_modules}
The examples presented in Section \ref{subsec:modular_imitation_learning} demonstrated that task-space position and orientation can be independently planned and linearly combined at the joint-torque command level. 
While the proposed modular approach simplifies motion planning by decoupling task-space position and orientation, certain manipulation tasks may necessitate coupling between the two. 
For instance, tasks that involve a specific relation between translational and rotational movements (e.g., screwing or tightening a nut, opening a door knob) impose a constraint (or coupling) between task-space position and orientation. 
Accounting for the coupling between task-space position and orientation using the presented modular approach presents opportunities for further research.

\subsection{Using Motor Primitives as Effective Inductive Bias}
Learning-based methods have provided successive breakthroughs in multiple domains \citep{lecun2015deep}, including image processing \citep{krizhevsky2012imagenet}, natural language processing \citep{vaswani2017attention}.
The learning-based method and particularly Reinforcement Learning have also been applied to address a wide range of tasks for robot control \citep{kaelbling1996reinforcement,kober2013reinforcement,silver2014deterministic,lillicrap2015continuous,schulman2015trust,haarnoja2018soft,chi2023diffusion}.

Despite such achievements, addressing learning-based methods for physical robotic systems, particularly in tasks requiring contact and physical interaction, remains a challenge unlike the successes observed in other fields \citep{lutter2019deep}.
One reason stems from the fact that learning-based methods often fail to account for the natural laws (e.g., symmetries and conservation laws) which all physical objects are subject to \citep{duruisseaux2023lie}. 
This also leads to the problem of data-efficiency for learning-based methods. 
Since Reinforcement Learning methods allow the robots to learn and adapt by trial-and-error (or the ``exploration'' process \citep{mehlhorn2015unpacking}), most of the approaches require substantial training time even with modern computational resources, making them prohibitively expensive for applications such as robotics \citep{chatzilygeroudis2019survey}. 
Supervised Learning which trains a Neural Network based on pre-collected datasets may alleviate this problem \citep{chi2023diffusion}. 
Nevertheless, the approach often requires an extremely large dataset to train the network. 
Even if a sufficient number of datasets are collected for training, the problem of generalization for robotic manipulation still remains.

To address these problems, methods that exploit prior knowledge or pre-defined structures have been explored. 
In detail, embedding prior knowledge of physics laws or structural properties of dynamical systems into the design of a robot controller has proven to be a powerful technique for improving their computational efficiency and generalization capacity. 
These structures, commonly referred to as ``inductive bias'' \citep{helmbold2015inductive}, have been employed for faster learning speed, higher accuracy and better generalization \citep{schmidt2009distilling, nguyen2010using,greydanus2019hamiltonian,ploeger2021high, nah2020dynamic,nah2021manipulating,nah2023learning}.

Despite the clear advantages, selecting the appropriate inductive biases for robot control remains an open challenge. Research has demonstrated that the choice of action space can greatly impact the efficiency of motor learning and the quality of the resulting behavior \citep{peng2017learning, martin2019variable}. 
Furthermore, it has been shown that a judicious selection of pre-defined structures significantly accelerates motor learning, improving both speed and efficacy \citep{ploeger2021high}. 
The right inductive bias not only simplifies the learning process but also enables the robot to generalize across tasks more effectively.

We propose that the modules presented in this Thesis may serve as key inductive biases to facilitate robot learning. These modules may enable the robot to adapt to a wide variety of tasks with greater efficiency, while preserving stability against contact and physical interaction. By incorporating these modules as inductive biases, future robot control systems may achieve faster learning, higher accuracy, and enhanced generalization across diverse tasks.

\section{Conclusions}\label{sec:conclusion}
In this paper, a comprehensive formulation of modular robot control has been articulated. 
Inspired by neuromotor control research, we claim that modular control based on motor primitives may be a key towards bridging the performance gap between humans and robots.
Not only may this improve the capabilities of robots, but we also expect that the presented approach may serve as a theoretical framework to account for observable motor behaviors of biological systems. 
The presented modular framework will offer both an effective constructive framework for practical applications in robotics, and a valuable descriptive framework for advancing our understanding of human motor performance.

\section*{Acknowledgements}

\subsection*{Funding}
MCN was supported in part by a Mathworks Fellowship.
\hb{JL was supported in part by the MIT-Novo Nordisk Artificial Intelligence Postdoctoral Fellows Program.}
\hb{This work was supported in part by the Eric P. and Evelyn E. Newman Fund.}

\hb{The authors would like to thank Stephan Stansfield for designing the robot’s end-effector tool used in the cocktail-shaking task (Section \ref{subsec:modular_imitation_learning}). 
The authors appreciate Dr. Federico Tessari for contributing the idea for robotic experiment exploiting kinematic singularities (Section \ref{subsubsec:exploiting_kinematic_singularity}).} 

\hb{We gratefully acknowledge the support of KUKA, an international leader in automation solutions, and specifically thank them for providing the KUKA robots used in our experiments.}

\subsection*{Declaration of Conflicting Interests}
The Authors declare that there is no conflict of interest.

\appendix

\renewcommand{\theequation}{A.\arabic{equation}}
\setcounter{equation}{0}  

\section{Special Orthogonal Group \text{SO}(3)}\label{appx:SO3}
The Special Orthogonal Group in three dimensions, $\text{SO}(3)$ is defined by:
\begin{equation}\label{appx_eq:definition_of_SO3}
\text{SO}(3) = \{  \mathbf{R} \in \mathbb{R}^{3\times 3} ~|~ \mathbf{R}\mathbf{R}^{\top} =\mathbf{R}^{\top}\mathbf{R}=\mathbb{I}_3, ~~\det(\mathbf{R})=1 \}
\end{equation} 
$\text{SO}(3)$ is a Lie Group: a Group under matrix multiplication, and a three-dimensional smooth manifold \citep{murray1994mathematical,lynch2017modern}. The element of $\text{SO}(3)$, $\mathbf{R}$ is referred to as a spatial rotation matrix.

The element of SO(3) expresses a spatial orientation between two orthonormal right-handed frames.
In detail, given two orthonormal frames $\{S\}$ and $\{B\}$, a spatial rotation matrix of frame $\{B\}$ with respect to frame $\{S\}$ is denoted by ${}^{S}\mathbf{R}_{B}\in\text{SO}(3)$ \citep{lachner2022geometric}. 

Associated with every Lie Group is its Lie Algebra, which is the tangent space at the identity element \citep{murray1994mathematical,lynch2017modern}. 
The Lie Algebra of $\text{SO}(3)$, denoted by $\text{so}(3)\equiv \text{T}_{\mathbb{I}_3}\text{SO}(3)$, is a real Vector Space of 3-by-3 skew-symmetric real matrices:
\begin{equation}\label{appx_eq:definition_of_so3}
\text{so}(3) = \{  [\bm{\omega}] \in \mathbb{R}^{3\times 3} ~|~ [\bm{\omega}]^{\top} = -[\bm{\omega}] \}
\end{equation} 
where $[ \, \cdot \, ]:\mathbb{R}^3\rightarrow \text{so}(3)$ is the skew-symmetric matrix (or a cross-product matrix \citep{fasse1998spatial}) representation of a three-dimensional vector \citep{lynch2017modern}:\footnote{There is another notation called the hat operator \citep{murray1994mathematical}, $\wedge:\mathbb{R}^{3}\rightarrow \text{so}(3)$. 
However, in this paper, hat symbol is reserved to denote unit vectors.}
\begin{equation}\label{appx_eq:skew_symmetric_matrix_representation}
    \bm{\omega}=
    \begin{bmatrix}
        \omega_1 \\
        \omega_2 \\
        \omega_3
    \end{bmatrix} ~~~~
    [\bm{\omega}] \equiv
    \begin{bmatrix}
        0 & -\omega_3 & \omega_2 \\
        \omega_3 & 0 & -\omega_1 \\
       -\omega_2 & \omega_1 & 0
    \end{bmatrix} 
\end{equation}
The inverse of $[ \,\cdot\,]$, $\vee:\text{so}(3)\rightarrow \mathbb{R}^{3}$ is defined by:
\begin{equation}\label{appx_eq:vee_operator}
    [\bm{\omega}]^{\vee} = \bm{\omega}
\end{equation}

\subsection{Exponential and Logarithmic Maps}\label{appx:sec_exp_log_maps}
One can map an element of so(3) to SO(3) using an Exponential map \citep{leonard1996matrix}. By reformulating $[\bm{\omega}]=[\hat{\bm{\omega}}]\theta$, where $\theta=\|\bm{\omega}\|$:
\begin{equation}\label{appx_eq:rodriguez_formula}
    \text{exp}_{\text{SO}(3)}( [\hat{\bm{\omega}}]\theta ) = \mathbb{I}_3 + \sin\theta[\hat{\bm{\omega}}] + (1-\cos\theta)[\hat{\bm{\omega}}]^2
\end{equation}
This is the Rodrigues' rotation formula \citep{murray1994mathematical,lynch2017modern}.

One can also define a Logarithmic Map, $\text{log}_{\text{SO}(3)}:\text{SO}(3)\rightarrow \text{so}(3)$ for $\theta \neq k\pi$ for $k\in\mathbb{Z}$:
\begin{align}\label{appx_eq:logarithm_so3}
    \begin{split}
        \text{log}_{\text{SO}(3)}( \mathbf{R} ) &= \frac{\theta}{2\sin \theta} (\mathbf{R}-\mathbf{R}^{\top}) \\
        \theta &= \arccos \Big( \frac{\text{tr}(\mathbf{R})-1}{2} \Big)
    \end{split}
\end{align}
If $\theta = k\pi$ for $k\in\mathbb{Z}$ \citep{murray1994mathematical,lynch2017modern}:
\begin{itemize}
    \item If $k$ is an even number, i.e., $\text{tr}(\mathbf{R})=3$, $\text{log}_{\text{SO}(3)}(\mathbf{R})=\mathbf{0}$.
    \item If $k$ is an odd number, i.e., $\text{tr}(\mathbf{R})=-1$, three possible choices exist for $\text{log}_{\text{SO}(3)}(\mathbf{R})$, and a feasible solution can be selected from them (\cite{lynch2017modern}, Equation (3.59)).
\end{itemize}

To simplify the notations, a capitalized versions of Exponential and Logarithmic maps, $\text{Exp}_{\text{SO}(3)}:\mathbb{R}^{3}\rightarrow \text{SO}(3)$ and $\text{Log}_{\text{SO}(3)}:\text{SO}(3)\rightarrow \mathbb{R}^{3}$, are defined:
\begin{align}\label{appx_eq:exp_log_capitalized}
    \begin{split}
        \text{Exp}_{\text{SO}(3)}(\bm{\omega}) &= \text{exp}_{\text{SO}(3)}([\bm{\omega}]) \\
        \text{Log}_{\text{SO}(3)}( \mathbf{R} ) &= \text{log}_{\text{SO}(3)}( \mathbf{R} )^{\vee}
    \end{split}
\end{align}
The components of $\text{Log}_{\text{SO}(3)}( \mathbf{R} )$ are also called exponential coordinates of $\mathbf{R}\in\text{SO}(3)$.

\subsection{Time Derivative of the Logarithmic Map}\label{appx_sec:time_derivative_of_logarithm_SO3}
Let the orientation of frame $\{B\}$ with respect to frame $\{S\}$ be time-varying, i.e., ${}^{S}\mathbf{R}_{B}(t)\in \text{SO}(3)$. 
The angular velocity of frame $\{B\}$ with respect to frame $\{S\}$, either expressed in frame $\{S\}$ or $\{B\}$, ${}^{S}\bm{\omega}(t)$ or ${}^{B}\bm{\omega}(t)$ are given by \citep{murray1994mathematical,lynch2017modern,lachner2022geometric}:
\begin{align}\label{appx_eq:angular_vel_rot_mat}
    \begin{split}
        [{}^{S}\bm{\omega}(t)] &= \bigg( \frac{d}{dt}{}^{S}\mathbf{R}_{B}(t) \bigg) {}^{S}\mathbf{R}_{B}^{\top}(t) \\
        [{}^{B}\bm{\omega}(t)] &= {}^{S}\mathbf{R}_{B}^{\top}(t)\bigg( \frac{d}{dt}{}^{S}\mathbf{R}_{B}(t) \bigg)     
    \end{split}
\end{align}

Consider three orthonormal frames $\{0\}$, $\{S\}$, $\{G\}$. The spatial orientation between these three frames is related by ${}^{0}\mathbf{R}_{G}(t)\equiv {}^{S}\mathbf{R}_{0}^{\top}(t){}^{S}\mathbf{R}_{G}\in \text{SO}(3)$ (Equation \eqref{eq:DMP_transformation_orient_SO3}). Let $\theta(t)=\|\text{Log}_{\text{SO}(3)}({}^{0}\mathbf{R}_{G}(t))\|$, and assume $\theta(t)\neq k\pi$ for $k\in\mathbb{Z}$ (Equation \eqref{appx_eq:logarithm_so3}). 
To avoid clutter, the superscript and subscript notations are often omitted, i.e., ${}^{0}\mathbf{R}_{G}(t)\equiv \mathbf{R}$.
The time derivative of the Logarithmic Map of $\mathbf{R}$ is defined by:
\begin{align}
    \begin{split}
        \frac{d}{dt}\log_{\text{SO}(3)}(\mathbf{R}) =& \bigg( \frac{\theta\cos\theta-\sin\theta}{4\sin^{3}\theta}\bigg)\text{tr}(\dot{\mathbf{R}}) (\mathbf{R}-\mathbf{R}^{\top} ) \\
        &- \frac{\theta}{2\sin\theta} \bigg\{
        \dot{\mathbf{R}} - \dot{\mathbf{R}}^{\top} \bigg\} \\
        \dot{\mathbf{R}} =& -{}^{S}\mathbf{R}_{0}^{\top}[{}^{S}\bm{\omega}] {}^{S}\mathbf{R}_{G}
    \end{split} 
\end{align}

\renewcommand{\theequation}{B.\arabic{equation}}
\setcounter{equation}{0}  

\section{Quaternions}\label{appx:H1}
A quaternion $\vec{\mathbf{q}}\in \mathbb{H}$ is defined by:
\begin{equation}
    \vec{\mathbf{q}} = q_w + q_x \mathbf{i} + q_y \mathbf{j} + q_z \mathbf{k}
\end{equation}    
where $q_w, q_x, q_y, q_z \in\mathbb{R}$ are real coefficients of a quaternion; $\mathbf{i},\mathbf{j}, \mathbf{k} \in\mathbb{H}$ are the basis vectors which satisfy:
$$
    \mathbf{i}^{2} = \mathbf{j}^{2} = \mathbf{k}^{2} = \mathbf{ijk}=-1
$$
For notational simplicity, a quaternion $\vec{\mathbf{q}}\in\mathbb{H}$ is denoted by $\vec{\mathbf{q}}=(q_w, \mathbf{q}_{v})$, where $\mathbf{q}_{v}=( q_x, q_y, q_z)\in\mathbb{R}^{3}$. 
Analogous to complex numbers, $q_w$ (respectively $\mathbf{q}_{v}$) is referred to as the real (respectively imaginary) part of quaternion $\vec{\mathbf{q}}$.
Operators $\text{Re}: \mathbb{H}\rightarrow \mathbb{R}$, $\text{Im}: \mathbb{H}\rightarrow \mathbb{R}^{3}$ are defined by:
\begin{equation}\label{appx_eq:quaternion_basic_operations}
    \text{Re}( \vec{\mathbf{q}} ) = q_w ~~~  \text{Im}( \vec{\mathbf{q}} ) = \mathbf{q}_{v}
\end{equation}
Given a quaternion $\vec{\mathbf{q}}=(q_w, \mathbf{q}_v)\in\mathbb{H}$, its conjugation, $\vec{\mathbf{q}}^{*}\in \mathbb{H}$ is defined by:
\begin{equation}\label{appx_eq:quaternion_conjugate}
    \vec{\mathbf{q}}^{*} = q_w - q_x \mathbf{i} - q_y \mathbf{j} - q_z \mathbf{k} = (q_w, -\mathbf{q}_{v})
\end{equation}
In other words, a conjugate of a quaternion flips the sign of its imaginary part.

Given a three-dimensional real vector $\bm{\omega}=(\omega_x, \omega_y,\omega_z)\in\mathbb{R}^{3}$, an operation to extend this vector to a quaternion $\mathbb{H}$ is defined by:
\begin{equation}\label{appx_eq:quaternion_extension}
    \vec{\bm{\omega}}= \omega_x \mathbf{i} + \omega_y \mathbf{j} + \omega_z \mathbf{k} \equiv (0, ~\bm{\omega})
\end{equation}

Given two quaternions $\vec{\mathbf{q}}, \vec{\mathbf{p}} \in \mathbb{H}$, their multiplication $\otimes:\mathbb{H}\times \mathbb{H}\rightarrow \mathbb{H}$ is defined by:
\begin{equation}\label{appx_eq:quaternion_multiplication}
    \vec{\mathbf{q}} \otimes \vec{\mathbf{p}} = ( q_w p_w - \mathbf{q}_{v}\cdot \mathbf{p}_v, ~ q_w \mathbf{p}_{v} + p_w \mathbf{q}_{v} + [\mathbf{q}_{v}] \mathbf{p}_v)
\end{equation}
where $\cdot:\mathbb{R}^3\times \mathbb{R}^3\rightarrow \mathbb{R}$ is a dot product between two real vectors.

\subsection{Exponential and Logarithmic Maps}\label{subsec:appx_exp_Logarithmic_quaternions}
Analogous to Appendix \ref{appx:sec_exp_log_maps}, an Exponential map of a quaternion $\vec{\mathbf{q}}=(q_w, \mathbf{q}_{v})$ is given by:
\begin{equation}\label{appx_eq:quaternion_exp_map_general_quat}
    \text{exp}_{\mathbb{H}}(\vec{\mathbf{q}}) = \exp(q_w) ( \cos||\mathbf{q}_{v}||, ~\sin||\mathbf{q}_{v}|| \hat{\mathbf{q}}_{v})
\end{equation}
where $\hat{\mathbf{q}}_{v}=\mathbf{q}_v\| \mathbf{q}_v \|$. 

A Logarithmic map of a quaternion, $\text{log}_{\mathbb{H}}:\mathbb{H}\rightarrow \mathbb{H}$ is defined by:
\begin{equation}\label{appx_eq:quaternion_log_map_unit_quat}
    \log_{\mathbb{H}}(\vec{\mathbf{q}}) = ( \log\|\vec{\mathbf{q}}\|, ~\arccos\Big( \frac{q_w}{ \| \vec{\mathbf{q}} \|}\Big) \hat{\mathbf{q}}_v )
\end{equation}

To simplify the notations, the Exponential and Logarithmic maps between $\mathbb{R}^3$ and $\mathbb{H}_1$, $\text{Exp}_{\mathbb{H}}:\mathbb{R}^{3}\rightarrow \mathbb{H}_{1}$ and $\text{Log}_{\mathbb{H}}: \mathbb{H}_{1}\rightarrow \mathbb{R}^{3}$ are defined by:
\begin{equation}\label{appx_eq:capitalized_Exp_Log_quat}
    \text{Exp}_{\mathbb{H}}(\bm{\omega}) = \text{exp}_{\mathbb{H}}( \vec{\bm{\omega}} )~~~~~ \text{Log}_{\mathbb{H}}( \vec{\mathbf{q}}) = \text{Im}(\log_{\mathbb{H}} \vec{\mathbf{q}} )
\end{equation}
As with the case for SO(3) (Appendix \ref{appx:sec_exp_log_maps}), the Logarithmic map for unit quaternions can be regarded as the exponential coordinates of the unit quaternion.

\subsection{Conversion between \texorpdfstring{$\mathbb{H}_1$}{H1} and SO(3)}\label{appx_sec:quat_so3_conversion}
As with spatial rotation matrices (Appendix \ref{appx:SO3}), unit quaternion describes a spatial orientation between two orthonormal frames. 
Hence, one can map between $\mathbb{H}_1$ and SO(3).

\subsubsection{From \texorpdfstring{$\mathbb{H}_1$}{H1} to SO(3)}
Consider an element of $\mathbb{H}_1$ parameterized by $\vec{\mathbf{q}}\equiv (\eta, \bm{\epsilon} )$, where $\eta\in\mathbb{R}$ and $\bm{\epsilon}\in\mathbb{R}^3$. 
This unit quaternion can be mapped to an element of $\text{SO}(3)$ by Equation \eqref{appx_eq:rodriguez_formula}:
\begin{equation}\label{appx_eq:relation_between_unit_quat_SO3}
    (\eta, \bm{\epsilon}) \in\mathbb{H}_1 ~~ \Longrightarrow ~~ \mathbb{I}_3 + 2\eta [\bm{\epsilon}] + 2[\bm{\epsilon}]^{2} \in\text{SO}(3)
\end{equation}
 
\subsubsection{From SO(3) to \texorpdfstring{$\mathbb{H}_1$}{H1}}\label{appx:subsec_SO3_to_unit_quaternion}
To map an element of $\text{SO}(3)$ (i.e., a spatial rotation matrix) to a unit quaternion, one can refer to Algorithm 2, Appendix B.2 of \cite{allmendinger2015computational}, which is originally derived by \cite{shepperd1978quaternion}.

\subsection{Quaternion Kinematic Equation}\label{sec:quat_kin_equation}
Let ${}^{S}\mathbf{q}_{B}(t)\in\mathbb{H}_1$ be a time-varying spatial orientation of frame $\{B\}$ with respect to frame $\{S\}$. The angular velocity of frame $\{B\}$ with respect to frame $\{S\}$, either expressed in frame $\{S\}$ or $\{B\}$, ${}^{S}\bm{\omega}(t)$ or ${}^{B}\bm{\omega}(t)$ are given by:
\begin{align}
    \begin{split}
        \frac{d}{dt}{}^{S}\vec{\mathbf{q}}_{B}(t) &= \frac{1}{2} {}^{S}\vec{\bm{\omega}}(t) \otimes {}^{S}\vec{\mathbf{q}}_{B}(t) \\
        \frac{d}{dt}{}^{S}\vec{\mathbf{q}}_{B}(t) &= \frac{1}{2} {}^{S}\vec{\mathbf{q}}_{B}(t) \otimes {}^{B}\vec{\bm{\omega}}(t)
    \end{split}
\end{align}
For notational simplicity, we omit the superscript, subscript and its time argument, i.e., ${}^{S}\mathbf{q}_{B}(t)\equiv \mathbf{q}(t)$.

By defining the parameters of a unit quaternion as $\vec{\mathbf{q}}(t)\equiv (\eta(t), \bm{\epsilon}(t))$, the time derivatives of the parameters are given by \citep{robinson1958use,wie1989quarternion}:
\begin{align}\label{appx_eq:quaternion_propagation}
    \begin{split}
        \frac{d}{dt}
        \begin{bmatrix}
            \eta(t) \\
            \bm{\epsilon}(t)
        \end{bmatrix} 
        &= 
        \frac{1}{2}
        \begin{bmatrix}
            -\bm{\epsilon}^{\top}(t) \\
            \eta(t)\mathbb{I}_{3} - [\bm{\epsilon}(t)]
        \end{bmatrix}
        {}^{S}\bm{\omega}(t) \\
        &\equiv \frac{1}{2}\mathbf{J}_{\mathbb{H}}(\eta(t), \bm{\epsilon}(t)) {}^{S}\bm{\omega}(t)    
    \end{split}
\end{align}
In this equation, $\mathbf{J}_{\mathbb{H}}(\eta(t), \bm{\epsilon}(t))\in\mathbb{R}^{4\times 3}$ is a Jacobian matrix for unit quaternions, which satisfies the following identities \citep{lizarralde1996attitude,koutras2020correct}: 
\begin{align}\label{appx_eq:quat_jacobian}
    \begin{split}
        \mathbf{J}_{\mathbb{H}}^{\top}(\eta(t), \bm{\epsilon}(t))\mathbf{J}_{\mathbb{H}}(\eta(t), \bm{\epsilon}(t)) &=\mathbb{I}_{3} \\
        \mathbf{J}_{\mathbb{H}}^{\top}(\eta(t), \bm{\epsilon}(t))
        \begin{bmatrix}
            \eta(t) \\
            \bm{\epsilon}(t)
        \end{bmatrix} &= \mathbf{0}
    \end{split}
\end{align}
The submatrix of $\mathbf{J}_{\mathbb{H}}(\eta(t), \bm{\epsilon}(t))$, $\mathbf{E}(\eta(t), \bm{\epsilon}(t))\equiv \eta(t)\mathbb{I}_{3} - [\bm{\epsilon}(t)]$ is used for the task-space impedance control for orientation (Section \ref{subsec:module_for_task_space_orientation_H1}).

\subsection{Time Derivative of the Logarithmic Map}\label{appx_sec:time_derivative_of_logarithm_H1}
Given a time-varying unit quaternion $\vec{\mathbf{q}}(t)\equiv \vec{\mathbf{q}} =(\eta, \bm{\epsilon})$, the time derivative of its Logarithmic map is given by \citep{koutras2020correct}:
\begin{equation}\label{appx_eq:time_derivative_Logarithmic_map}
    \frac{d}{dt} \text{Log}_{\mathbb{H}}(\vec{\mathbf{q}})  = 
    \begin{bmatrix}
            \Big( \frac{-\|\bm{\epsilon}\|+\arccos(\eta)\eta}{\|\bm{\epsilon}\|^3} \Big)\bm{\epsilon} & \frac{\arccos(\eta) }{\|\bm{\epsilon}\|} \mathbb{I}_{3}
    \end{bmatrix}
    \frac{d}{dt}
    \begin{bmatrix}
        \eta \\
        \bm{\epsilon}
    \end{bmatrix} 
\end{equation}

\subsection{Equivalence to Rotational Potential Energy}\label{appx:sec_equivalence_potential_energy}
In this section, the derivations from \cite{zhang2000spatial} are reformulated using the notations adopted in this paper.
Consider a unit quaternion ${}^{S}\vec{\mathbf{q}}_{B}=({}^{S}\eta_{B}, {}^{S}\bm{\epsilon}_{B})\in\mathbb{H}_{1}$ and its corresponding spatial rotation matrix (Equation \eqref{appx_eq:relation_between_unit_quat_SO3}) (Appendix \ref{appx_sec:quat_so3_conversion}):
\begin{equation}\label{eq:substituting_SO3_to_unit_quat}
    {}^{S}\mathbf{R}_{B} = \mathbb{I}_{3} + 2 {}^{S}\eta_{B}[{}^{S}\bm{\epsilon}_{B}] + 2 [{}^{S}\bm{\epsilon}_{B}]^{2}
\end{equation}
A potential energy function on the $\text{SO}(3)$ manifold, $U_{r}:\text{SO}(3)\rightarrow \mathbb{R}$ is defined by:
$$
    U_{r}({}^{S}\mathbf{R}_{B}) = -\text{tr}(\mathbf{G}{}^{S}\mathbf{R}_{B})
$$
In this equation, $\mathbf{G}\in\mathbb{R}^{3\times 3}$ is the co-stiffness matrix of stiffness matrix $\mathbf{K}\in\mathbb{R}^{3\times 3}$ where its definition will be introduced shortly.

Substituting ${}^{S}\mathbf{R}_{B}$ with the parameters of its corresponding unit quaternion results in:
\begin{align*}
   -\text{tr}(\mathbf{G}{}^{S}\mathbf{R}_{B}) &=  -\text{tr}(\mathbf{G}) +  2 \| {}^{S}\bm{\epsilon}_{B} \|^2\text{tr} ( \mathbf{G}  ) - 2\text{tr}(\mathbf{G} {}^{S}\bm{\epsilon}_{B} {}^{S}\bm{\epsilon}_{B}^{\top}) \\
    &=-\text{tr}(\mathbf{G}) + 2 {}^{S}\bm{\epsilon}_{B}^{\top} \underbrace{( \text{tr}(\mathbf{G})\mathbb{I}_3 - \mathbf{G})}_{\equiv \mathbf{K}}{}^{S}\bm{\epsilon}_{B} \\
    &= -\text{tr}(\mathbf{G}) + 2{}^{S}\bm{\epsilon}_{B}^{\top} \mathbf{K}{}^{S}\bm{\epsilon}_{B}
\end{align*}
In this equation, $\mathbf{K}\equiv  \text{tr}(\mathbf{G})\mathbb{I}_3 - \mathbf{G}$ and $\mathbf{G}\equiv \frac{1}{2}\text{tr}(\mathbf{K})\mathbb{I}_3 - \mathbf{K}$. Given a constant matrix $\mathbf{G}$ (and constant $\mathbf{K}$) matrix, potential energy $-\text{tr}(\mathbf{G}{}^{S}\mathbf{R}_{B})$ and $ 2{}^{S}\bm{\epsilon}_{B}^{\top} \mathbf{K}{}^{S}\bm{\epsilon}_{B}$ differs only by a constant $\text{tr}(\mathbf{G})$.

\renewcommand{\theequation}{C.\arabic{equation}}
\setcounter{equation}{0}

\section{Dynamic Movement Primitives}\label{sec:DMP_review}
In this part of the Appendix, an overview of Dynamic Movement Primitives (DMP) is provided.  
While multiple variations of DMP exist, in general, DMP parameterizes the trajectory with three distinct dynamical systems---``canonical systems,'' ``nonlinear forcing terms,'' and ``transformation systems'' \citep{ijspeert2013dynamical,saveriano2023dynamic}. 

The types of canonical systems and nonlinear forcing terms are divided to generate discrete and rhythmic movements.
Different types of transformation systems are employed to generate trajectories for joint-space, task-space position and task-space orientation. For orientation, either spatial rotation matrices $\text{SO}(3)$ (Appendix \ref{appx:SO3}) or unit quaternions $\mathbb{H}_{1}$ (Appendix \ref{appx:H1}) can be used with minor technical differences.

For the notations, we use subscript $0$ to indicate that DMP is later used to generate the virtual trajectory of EDA (Section \ref{subsec:def_module}).

\subsection{Canonical System}\label{subsubsec:canonical_system}
A canonical system of DMP acts as an ``internal clock'' for the kinematic primitives. 
It provides a notion of phase without an explicit time parameterization \citep{ijspeert2013dynamical,saveriano2023dynamic}.
The types of canonical systems are different for discrete and rhythmic movements, as the phase of rhythmic movement must be periodic rather than monotonically increasing as in discrete movement.

\subsubsection{For Discrete Movements}\label{subsubsubsec:canonical_system_discrete}
A canonical system for discrete movement, $s_{d}:\mathbb{R}_{\ge 0}\rightarrow \mathbb{R}_{\ge 0}$ is a scalar variable governed by a first-order linear differential equation \citep{ijspeert2013dynamical,saveriano2023dynamic}:
\begin{equation}\label{eq:DMP_canonical_system_discrete}
        \tau_{d} \dot{s}_{d}(t) = -\alpha_s s_{d}(t) 
\end{equation}
In these equations, $\alpha_s\in\mathbb{R}_{>0}$; $\tau_{d} \in\mathbb{R}_{>0}$ is the duration of the discrete movement.
The canonical system is exponentially convergent to 0 with a closed-form solution $s_{d}(t)=\exp(-\alpha_st/\tau_{d})s_{d}(0)$.
Usually, the initial condition $s_{d}(0)$ is set to be 1 \citep{ijspeert2013dynamical,saveriano2023dynamic}.

\subsubsection{For Rhythmic Movements}\label{subsubsubsec:canonical_system_rhythmic}
A canonical system for rhythmic movement, $s_{r}:\mathbb{R}_{\ge 0}\rightarrow [0,2\pi)$ is a scalar variable ranged between $0$ and $2\pi$. 
The basic form is defined by \citep{ijspeert2013dynamical,saveriano2023dynamic}:
\begin{equation}\label{eq:DMP_canonical_system_rhythmic}
    s_{r}(t) = \frac{t}{\tau_{r}} \mod 2\pi
\end{equation}
In other words, the canonical system for rhythmic movement is governed by a first-order differential equation $\dot{s}_{r}(t)=1/\tau_{r}$, but the modulo-$2\pi$ operation is applied for $s_{r}(t)$ to ensure $s_{r}(t)\in[0,2\pi)$.

Compared to a discrete canonical system (Equation \eqref{eq:DMP_canonical_system_discrete}), a rhythmic canonical system can be considered as a phase variable with constant angular velocity of $1/\tau_{r}$. 
Hence, to generate a rhythmic movement with a period of $T_r$, $\tau_{r}$ is chosen to be $T_r/2\pi$ \citep{ijspeert2013dynamical}. 
This implies that the period of the rhythmic movement should be extracted beforehand using an additional method \citep{righetti2006dynamic,gams2009line,ijspeert2013dynamical}.

Note that different variations of rhythmic canonical systems exist, where the canonical system is defined by a phase variable of a dynamical system with a stable limit cycle \citep{chung2010neurobiologically,wensing2017linear}.  
Nevertheless, the presented canonical system for rhythmic movement $s_r$ is sufficient to achieve a wide range of control tasks. 

\subsection{Nonlinear Forcing Terms}\label{subsubsec:nonlinear_forcing_term}
The nonlinear forcing term consists of a linear summation of finite basis functions with weights that can be learned through learning-based methods. 
As with the canonical system, the type of nonlinear forcing terms is different for discrete and rhythmic movements.
Details for determining the weights are deferred to Appendix \ref{subsubsec:imitation_learning}.

\subsubsection{For Discrete Movements}\label{subsubsec:nonlinear_forcing_term_discrete}
A nonlinear forcing term for discrete movements, $\mathbf{F}_{d}:\mathbb{R}_{\ge 0}\rightarrow \mathbb{R}^{n}$, which takes the discrete canonical system $s_{d}(t)$ as an argument (Equation \eqref{eq:DMP_canonical_system_discrete}), is defined by:
\begin{align}\label{eq:DMP_nonlinear_force_discrete}
    \begin{split}
        \mathbf{F}_{d}(s_{d}(t)) &= \frac{\sum_{i=1}^{N}\mathbf{w}_i\phi_i(s_{d}(t))}{\sum_{i=1}^{N}\phi_i(s_{d}(t))} s_{d}(t) \\ 
        \phi_i(s_{d}(t)) &= \exp\big\{ -h_i(s_{d}(t)-c_i)^2 \big\} 
    \end{split} 
\end{align}
In these equations, $N$ is the number of basis functions; $\phi_i:\mathbb{R}_{\ge 0}\rightarrow \mathbb{R}$ is the $i$-th basis function of the nonlinear forcing term for discrete movement, which is a Gaussian function; $\mathbf{w}_i\in\mathbb{R}^{n}$ is the weight array of the $i$-th basis function that can be learned by learning-based methods; $c_i\in\mathbb{R}$ and $h_i\in\mathbb{R}$ are the center location and inverse width of the $i$-th basis function.

Parameters $N$, $c_i$, $h_i$ are manually chosen \citep{ijspeert2013dynamical,saveriano2023dynamic}. A common choice for these parameters is $c_i=\exp(-\alpha_s(i-1)/(N-1))$ for $i\in\{1,2,\cdots, N\}$ and $h_{i}=1/(c_{i+1}-c_i)^2$ for $i\in\{1,2,\cdots, N-1\}$ and $h_{N-1}=h_{N}$ \citep{saveriano2023dynamic}.

\subsubsection{For Rhythmic Movements}\label{subsubsec:nonlinear_forcing_term_rhythmic}
A nonlinear forcing term for rhythmic movements, $\mathbf{F}_{r}:[0,2\pi)\rightarrow \mathbb{R}^{n}$, which takes the rhythmic canonical system $s_{r}(t)$ as the function argument (Equation  \eqref{eq:DMP_canonical_system_rhythmic}), is defined by:
\begin{align}\label{eq:DMP_nonlinear_force_rhythmic}
    \begin{split}
        \mathbf{F}_{r}(s_{r}(t)) &= \frac{\sum_{i=1}^{N}\mathbf{w}_i\psi_i(s_{r}(t))}{\sum_{i=1}^{N}\psi_i(s_{r}(t))} \\ 
        \psi_i(s_{r}(t)) &= \exp\big\{  h_i (\cos(s_{r}(t)-c_i)-1) \big\}
    \end{split} 
\end{align}
In these equations, $\psi_i:[0, 2\pi)\rightarrow \mathbb{R}$ is the $i$-th basis function of the nonlinear forcing term for rhythmic movements, which is a von Mises function; $\mathbf{w}_i\in\mathbb{R}^{n}$ is the weight array of the $i$-th basis function that can be learned by learning-based methods (Appendix \ref{subsubsec:imitation_learning}).

As with the discrete nonlinear forcing term (Equation \eqref{eq:DMP_nonlinear_force_discrete}), parameters $N$, $c_i$, $h_i$ are manually chosen \citep{ijspeert2013dynamical}. A common choice for $c_i$ is $c_i=2\pi(i-1)/(N-1)$, for $i\in\{1,2,\cdots, N\}$ and $h_i=2.5N$ \citep{ijspeert2002learningrhythmic,peternel2016adaptive,saveriano2023dynamic}.

\subsection{Transformation System}\label{subsubsec:transformation_system}
The nonlinear forcing term $\mathbf{F}_{d}$ (respectively $\mathbf{F}_{r})$ (Appendix \ref{subsubsec:nonlinear_forcing_term}) with canonical system $s_{d}$ (respectively $s_{r}$) (Appendix \ref{subsubsec:canonical_system}) as its function argument, is used as an input to the transformation system to generate trajectories with arbitrary complexity.

The type of canonical system and nonlinear forcing term depends on whether the movement is discrete or rhythmic. 
On the other hand, the type of transformation systems depends on the type of movements.
Four types of transformation systems exist:
\begin{itemize}
    \item Transformation system for joint-space trajectory, $\mathcal{Q} \;(=\mathbb{R}^{n})$, where $\mathcal{Q}$ is the Configuration Manifold of a robot \citep{bullo2019geometric}.
    \item Transformation system for task-space trajectory, position, $\mathbb{R}^{3}$.
    \item Transformation system for task-space trajectory, orientation, represented by spatial rotation matrices $\text{SO}(3)$ (Appendix \ref{appx:SO3}).
    \item Transformation system for task-space trajectory, orientation, represented by unit quaternions $\mathbb{H}_{1}$ (Appendix \ref{appx:H1}).
\end{itemize}

Since both nonlinear forcing terms and canonical systems for discrete and rhythmic movements can be used, subscripts $d$ and $r$ for $\mathbf{F}$, $s$ and $\tau$ are omitted for transformation systems.

\subsubsection{For Joint-space Position}\label{subsubsubsec:transformation_system_joint_space}
A transformation system to generate joint-space trajectory is defined by:
\begin{equation}\label{eq:DMP_transformation_joint_position}
    \begin{aligned}
        \tau \dot{\mathbf{q}}_0(t) &= \mathbf{z}_{q}(t) \\
        \tau \dot{\mathbf{z}}_{q}(t) &= \alpha_z \{ \beta_z ( \mathbf{q}_{g}-\mathbf{q}_{0}(t)) - \mathbf{z}_{q}(t)\} + \mathbf{S}_{q}\mathbf{F}(s(t))
    \end{aligned}    
\end{equation}
In these equations, $\mathbf{q}_{0}(t)\in\mathbb{R}^{n}$ and $\mathbf{z}_{q}(t) \in\mathbb{R}^{n}$ are position and (time-scaled) velocity of the joint-space trajectory, respectively; $\alpha_z, \beta_z\in\mathbb{R}_{>0}$ are constant positive coefficients that determine the zero-forced linear response of $\mathbf{q}_{0}(t)$.

For discrete movement $\mathbf{q}_{g}\in\mathbb{R}^{n}$ is the goal joint-configuration; for rhythmic movement, $\mathbf{q}_{g}$ is the center position for rhythmic movement to oscillate about \citep{ijspeert2013dynamical}. 
$\mathbf{S}_{q}\in\mathbb{R}^{n\times n}$ is the scaling matrix for the nonlinear forcing term $\mathbf{F}$. 
Further details on choosing these parameters are deferred to Appendix \ref{subsubsubsec:imit_learning_joint_space}.

While any positive values of $\alpha_z$ and $\beta_z$ can be used, usually, the values of $\alpha_z$ and $\beta_z$ are chosen such that if $\mathbf{F}(s(t))$ is zero, the transformation system is critically damped (i.e., has repeated eigenvalues) for $\tau=1$ (i.e., $\beta_z=\alpha_z/4$) \citep{ijspeert2013dynamical}.

Note that the gains $\alpha_z, \beta_z$ are chosen to be identical for all $n$ movements.
This choice is not necessary, and different gains can be used for each movement \citep{park2008movement,hoffmann2009biologically,pastor2009learning}.
Nevertheless, this is not only an unnecessary complication of the design, but also hinders the usage of other applications such as movement recognition \citep{ijspeert2013dynamical} or movement combination \citep{nah2025combining}.

Given $\mathbf{F}(s(t))$, $\mathbf{S}_{q}$, and initial conditions $\mathbf{q}_0(t=0),\mathbf{z}_q(t=0)$, the transformation system is forward integrated to generate $\mathbf{q}_0(t)$, $\mathbf{z}_q(t)$.

\subsubsection{For Task-space Position}\label{subsubsubsec:transformation_system_task_position}
A transformation system to generate trajectories for task-space position is defined by \citep{koutras2020novel}:
\begin{equation}\label{eq:DMP_transformation_task_position}
    \begin{aligned}
        \tau \dot{\mathbf{p}}_0(t) &= \mathbf{z}_{p}(t) \\
        \tau \dot{\mathbf{z}}_{p}(t) &= \alpha_z \{ \beta_z ( \mathbf{p}_{g}-\mathbf{p}_{0}(t)) - \mathbf{z}_{p}(t)\} + \mathbf{S}_{p}\mathbf{F}(s(t))
    \end{aligned}    
\end{equation}
In these equations, $\mathbf{p}_{0}(t)\in\mathbb{R}^{3}$ and $\mathbf{z}_{p}(t) \in\mathbb{R}^{3}$ are task-space position and its (time-scaled) velocity, respectively; 
$\mathbf{p}_{g}\in\mathbb{R}^{3}$ is the goal position for discrete movement, and the average position for rhythmic movement; 
$\mathbf{S}_{p}\in\mathbb{R}^{3\times 3}$ is the scaling matrix for the nonlinear forcing term $\mathbf{F}$.

For discrete movement, the scaling matrix $\mathbf{S}_p$ is defined by \citep{koutras2020novel}:
\begin{equation}\label{eq:imit_learning_scaling_position}
    \mathbf{S}_{p} = \mathbf{R}\frac{\|
    \mathbf{p}_{g}-\mathbf{p}_{i}\|}{\| \mathbf{p}_{g}^{(d)}-\mathbf{p}_{i}^{(d)}\|} 
\end{equation}
In this equation, $\mathbf{p}_{i}\in\mathbb{R}^{3}$ is the initial position of task-space position; $\mathbf{R}\in \text{SO}(3)$ is a spatial rotation matrix that orients the trajectory in three-dimensional space; superscript $(d)$ emphasizes the initial and goal positions of the demonstrated trajectory that we aim to imitate.
Further details on choosing $\mathbf{R}$, $\mathbf{p}_{i}^{(d)}$, and $\mathbf{p}_{g}^{(d)}$ are clarified in Appendix \ref{subsubsubsec:imit_learning_task_position}.

For rhythmic movement, $\mathbf{S}_{p}=r\mathbb{I}_3$, where $r\in\mathbb{R}_{>0}$ scales the amplitude of the rhythmic movement. 

Given $\mathbf{F}(s(t))$, $\mathbf{S}_{p}$, and initial conditions $\mathbf{p}_0(t=0),\mathbf{z}_p(t=0)$, the transformation system is forward integrated to generate $\mathbf{p}_0(t)$, $\mathbf{z}_p(t)$.

\subsubsection{For Task-space Orientation, $\text{SO}(3)$}\label{subsubsubsec:transformation_system_task_orientation_SO3}
Consider the spatial frame $\{S\}$ and the virtual frame $\{0\}$ (Appendix \ref{appx:SO3}).
Consider another frame, $\{G\}$; for discrete movement, $\{G\}$ is the desired frame to which $\{0\}$ aims to converge; for rhythmic movement, $\{G\}$ is located at the average of $^{S}\mathbf{R}_0(t)$, where the details for calculation are deferred to Appendix \ref{subsubsubsec:imit_learning_task_orientation}. With these three frames $\{S\}$, $\{0\}$, $\{G\}$, define ${}^{0}\mathbf{R}_{G}(t)\equiv  {}^{S}\mathbf{R}_{0}^{\top}(t) {}^{S}\mathbf{R}_{G}$ and its exponential coordinates ${}^{0}\mathbf{e}_{G}(t)\equiv \text{Log}_{\text{SO}(3)}({}^{0}\mathbf{R}_{G}(t))$ (Appendix \ref{appx:sec_exp_log_maps}).
To avoid clutter, the superscript and subscript notations of the exponential coordinates are omitted, i.e., ${}^{0}\mathbf{e}_{G}(t)\equiv \mathbf{e}(t)$.

Using $\text{SO}(3)$ matrices, a transformation system to generate trajectories for task-space orientation is defined by \citep{koutras2020correct}:
\begin{equation}\label{eq:DMP_transformation_orient_SO3}
    \begin{aligned}
        \tau \dot{\mathbf{e}}(t) &= \mathbf{z}_e(t) \\
        \tau \dot{\mathbf{z}}_e(t) &= -\alpha_z \{ \beta_z \mathbf{e}(t) + \mathbf{z}_e(t)\} + \mathbf{S}_{e}\mathbf{F}(s(t))
    \end{aligned}    
\end{equation}
In these equations, $\mathbf{z}_{e}(t) \in\mathbb{R}^{3}$ is (time-scaled) velocity of the exponential coordinates; $\mathbf{S}_{e}\in\mathbb{R}^{3\times 3}$ is the scaling matrix for the nonlinear forcing term $\mathbf{F}$. 

For discrete movement, the scaling matrix is defined by $\mathbf{S}_{e}=\text{diag}( \text{Log}_{\text{SO}(3)}({}^{S}\mathbf{R}_{i}^{\top} {}^{S}\mathbf{R}_{G}) )$, where ${}^{S}\mathbf{R}_{i}\equiv {}^{S}\mathbf{R}_{0}(t=0)$ is the initial orientation \citep{koutras2020correct}; $\text{diag}:\mathbb{R}^{n}\rightarrow \mathbb{R}^{n\times n}$ is a diagonalization of a vector. 
For rhythmic movement, $\mathbf{S}_{e}=r\mathbb{I}_{3}$.
Further details of choosing $\mathbf{S}_{e}$ are deferred to Appendix \ref{subsubsubsec:imit_learning_task_orientation}.

The presented transformation system for orientation is different from the one provided by \cite{ude2014orientation}.
While the formulation of \cite{ude2014orientation} (which extends the work of \cite{pastor2011online}) remains to be a prominent approach (with the stability proof provided by \cite{bullo1995proportional}), several limitations exist including the absence of spatial and temporal invariance properties for trajectory generation \citep{koutras2020correct}.
The presented transformation system by \cite{koutras2020correct} addresses the limitations of the one from \cite{ude2014orientation}.

Given $\mathbf{F}(s(t))$ and $\mathbf{S}_{e}$, and initial conditions $\mathbf{e}(t=0),\mathbf{z}_e(t=0)$, the transformation system is forward integrated to generate $\mathbf{e}(t)$, $\mathbf{z}_e(t)$. 
Once $\mathbf{e}(t)$ is generated, the trajectory of ${}^{S}\mathbf{R}_0(t)$ is recovered by the Exponential map (Equation \eqref{appx_eq:exp_log_capitalized}):
\begin{equation}\label{eq:exponential_map_SO3}
    {}^{S}\mathbf{R}_0(t) = {}^{S}\mathbf{R}_G \text{Exp}_{\text{SO}(3)}^{\top}( \mathbf{e}(t))
\end{equation}

\subsubsection{For Task-space Orientation, $\mathbb{H}_{1}$}\label{subsubsubsec:transformation_system_task_orientation_H1}
One can also use unit quaternions to represent spatial orientation (Appendix \ref{appx:H1}).
Consider the unit quaternions ${}^{S}\vec{\mathbf{q}}_{0}(t), {}^{S}\vec{\mathbf{q}}_{G}(t) \in \mathbb{H}_{1}$, which can be derived by ${}^{S}\mathbf{R}_0(t)$, ${}^{S}\mathbf{R}_G(t)\in\text{SO}(3)$ (Appendix \ref{appx:subsec_SO3_to_unit_quaternion}).  
The exponential coordinates of the unit quaternions, $\mathbf{e}(t)\equiv 2\text{Log}_{\mathbb{H}}({}^{S}\vec{\mathbf{q}}_{0}^{*}(t) \otimes {}^{S}\vec{\mathbf{q}}_{G} )$ are defined (Appendix \ref{subsec:appx_exp_Logarithmic_quaternions}).
With the exponential coordinates, a transformation system for task-space orientation is defined by \citep{koutras2020correct}:
\begin{equation}\label{eq:DMP_transformation_orient_unit_quat}
    \begin{aligned}
        \tau \dot{\mathbf{e}}(t) &= \mathbf{z}_e(t) \\
        \tau \dot{\mathbf{z}}_e(t) &= -\alpha_z \{ \beta_z \mathbf{e}(t) + \mathbf{z}_e(t)\} + \mathbf{S}_{e}\mathbf{F}(s(t))
    \end{aligned}    
\end{equation}
In these equations, $\mathbf{S}_{e}\in\mathbb{R}^{3\times 3}$ is the scaling matrix for the nonlinear forcing term $\mathbf{F}$.
Without the nonlinear forcing term input, the differential equation is analogous to the one from \cite{wie1989quarternion}.

For discrete movement, the scaling matrix is defined by $\mathbf{S}_{e}=2\text{diag}( \text{Log}_{\mathbb{H}}({}^{S}\vec{\mathbf{q}}_{i}^{*}\otimes {}^{S}\vec{\mathbf{q}}_{G}))$, where ${}^{S}\vec{\mathbf{q}}_{i} \equiv {}^{S}\vec{\mathbf{q}}_{0}(t=0)$. 
For rhythmic movement, $\mathbf{S}_{e}=r\mathbb{I}_{3}$. 
Further details of choosing $\mathbf{S}_{e}$ are deferred to Appendix \ref{subsubsubsec:imit_learning_task_orientation}.

Given $\mathbf{F}(s(t))$ and $\mathbf{S}_{e}$, and initial conditions $\mathbf{e}(t=0),\mathbf{z}_e(t=0)$, the transformation system is forward integrated to generate $\mathbf{e}(t)$, $\mathbf{z}_e(t)$.

Once $\mathbf{e}(t)$ is generated, trajectory ${}^{S}\vec{\mathbf{q}}_0(t)$ is recovered from the exponential coordinates $\mathbf{e}(t)$ by the Exponential map (Appendix \ref{appx_eq:capitalized_Exp_Log_quat}):
\begin{equation}\label{eq:exponential_map_unit_quaternion}
    {}^{S}\vec{\mathbf{q}}_{0}(t) = {}^{S}\vec{\mathbf{q}}_{G} \otimes \text{Exp}_{\mathbb{H}}^{*}\Big( \frac{1}{2}\mathbf{e}(t) \Big)  
\end{equation}

Note that the technical differences compared to the case of $\text{SO}(3)$ lie in the specifics of deriving the scaling matrix $\mathbf{S}_{e}$ and the exponential coordinates $\mathbf{e}(t)$, as well as the method for recovering the trajectory for spatial orientation from the Exponential maps (Appendix \ref{subsec:appx_exp_Logarithmic_quaternions}).

\subsection{Imitation Learning}\label{subsubsec:imitation_learning}
Without the nonlinear forcing term input, the transformation systems yield a response of the stable second-order linear system.
To generate a wider range of movements (e.g., rhythmic movements), the weights of the nonlinear forcing term input can be learned through various methods (Section \ref{subsubsec:review_DMP}). 
One prominent approach is Imitation Learning \citep{schaal1999imitation,ijspeert2013dynamical,saveriano2023dynamic}, where the trajectory provided by demonstration can be learned through basic matrix algebra. 

The $N$ weight arrays $\mathbf{w}_{i}$ of the nonlinear forcing term $\mathbf{F}$ can be collected as a matrix $\mathbf{W}\in\mathbb{R}^{n\times N}$ ($n=3$ for task-space position and for task-space orientation):
\begin{equation}\label{eq:weight_array}
    \mathbf{W} = 
    \begin{bmatrix}
        | & | & & | \\
        \mathbf{w}_1 & \mathbf{w}_2 & \cdots & \mathbf{w}_N \\
        | & | & & | 
    \end{bmatrix}
\end{equation}
With this weight matrix, the nonlinear forcing terms for discrete and rhythmic movements are:
$$
    \mathbf{F}_{d}(s_{d}(t)) = \mathbf{W}\mathbf{a}_{d}(s_d(t)) ~~~~~ \mathbf{F}_{r}(s_{r}(t)) = \mathbf{W}\mathbf{a}_{r}(s_r(t))
$$
where $\mathbf{a}_{d}:\mathbb{R}\rightarrow \mathbb{R}^{N}$ and $\mathbf{a}_{r}:[0,2\pi)\rightarrow \mathbb{R}^{N}$ are defined by:
\begin{align*}
    \mathbf{a}_{d}(s_d(t)) &= 
    \frac{s_{d}(t)}{\sum_{i=1}^{N}\phi_{i}(s_{d}(t))}        
    \begin{bmatrix}
        \phi_{1}(s_{d}(t)) & \cdots & \phi_{N}(s_{d}(t))
    \end{bmatrix}^{\top} \\
    \mathbf{a}_{r}(s_r(t)) &= 
    \frac{1}{\sum_{i=1}^{N}\psi_{i}(s_{r}(t))}        
    \begin{bmatrix}
        \psi_{1}(s_{r}(t)) & \cdots & \psi_{N}(s_{r}(t))
    \end{bmatrix}^{\top}
\end{align*}
Imitation Learning finds the best-fit weights by Locally Weighted Regression:
\begin{equation}
    \mathbf{W} = \mathbf{B}\mathbf{A}^{\top} ( \mathbf{A} \mathbf{A}^{\top})^{-1}
\end{equation}
The matrices $\mathbf{A}$ and $\mathbf{B}$ for joint-space, task-space position and task-space orientation are discussed next. 
To avoid clutter, subscripts $d$ and $r$ for $\mathbf{a}$, representing discrete and rhythmic movements respectively, are omitted.

\subsubsection{For Joint-space Position}\label{subsubsubsec:imit_learning_joint_space}
To imitate joint-space trajectories (Appendix \ref{subsubsubsec:transformation_system_joint_space}), $P$ data points of the demonstrated trajectory, $\mathbf{q}^{(d)}(t_i), \dot{\mathbf{q}}^{(d)}(t_i), \ddot{\mathbf{q}}^{(d)}(t_i) \in\mathbb{R}^{n}$ should be collected for $i\in\{1, 2,\cdots, P\}$. 
With these data points, the following matrices $\mathbf{A}\in \mathbb{R}^{N\times P}$ and $\mathbf{B}\in\mathbb{R}^{n\times P}$ are calculated:
\begin{align}\label{eq:Imitation_Learning_joint_space_AB}
\begin{split}
    \mathbf{A} =&
    \begin{bmatrix}
        \mathbf{a}(s^{(d)}(t_{1})) & \cdots & \mathbf{a}(s^{(d)}(t_{P})) 
    \end{bmatrix} \\
    \mathbf{B} =& 
    \begin{bmatrix}
        \mathbf{b}_{q}(t_{1}) & \mathbf{b}_{q}(t_{2}) & \cdots & \mathbf{b}_{q}(t_{P})
    \end{bmatrix}  \\   
    \mathbf{b}_{q}(t_i) =& (\tau^{(d)})^2  \ddot{\mathbf{q}}^{(d)}(t_i) + \alpha_z \tau^{(d)}  \dot{\mathbf{q}}^{(d)}(t_i)\\
    &+ \alpha_z \beta_z (\mathbf{q}^{(d)}(t_i) -\mathbf{q}_{g}^{(d)})
\end{split}
\end{align}
where $s^{(d)}$ for discrete (Appendix \ref{subsubsubsec:canonical_system_discrete}) and rhythmic movements (Appendix \ref{subsubsubsec:canonical_system_rhythmic}) are:
\begin{align}\label{eq:canonical_system_w_data}
    \begin{split}
        s_d^{(d)} &= \exp\Big( -\frac{\alpha_s}{\tau_d^{(d)}}t \Big)s_d(0) \\
        s_r^{(d)} &=  \frac{t}{\tau_{r}^{(d)}} \mod 2\pi
    \end{split}
\end{align}
For discrete movement, $\tau_{d}^{(d)}$, and $\mathbf{q}_{g}^{(d)}$ are the duration and final point of the demonstrated trajectory, respectively, i.e., $\tau_d^{(d)}=t_P-t_1$ and $\mathbf{q}_{g}^{(d)}=\mathbf{q}^{(d)}(t_P)$; 
for rhythmic movement, $\tau_r^{(d)}$ and $\mathbf{q}_{g}^{(d)}$ are the period divided by $2\pi$ and average of the demonstrated trajectory, respectively, i.e., $\tau_r^{(d)}=(t_P-t_1)/2\pi$ and $\sum_{i=1}^{N}\mathbf{q}^{(d)}(t_i)/N$.

To collect the $P$ sample points of $\mathbf{q}^{(d)}(t_i), \dot{\mathbf{q}}^{(d)}(t_i), \ddot{\mathbf{q}}^{(d)}(t_i)$, one can sample the joint-space trajectories $\mathbf{q}^{(d)}(t)$ and use a finite difference method with smoothing to collect the higher derivative terms. 

Once the weights are learned through Imitation Learning, one can spatially and temporally scale the discrete or rhythmic movement, without losing the qualitative property of the trajectory. 
These are the spatial and temporal invariance properties of DMP, which make DMP preferable over spline methods \citep{ijspeert2013dynamical,saveriano2023dynamic} (Section \ref{subsubsec:review_DMP}). 
For instance, $\tau_d$ and $\tau_r$ modulate the duration and period of discrete and rhythmic movements, respectively. 
By scaling the matrix $\mathbf{S}_q$, the amplitude of the joint-space movement can be modulated. 

\subsubsection{For Task-space Position}\label{subsubsubsec:imit_learning_task_position}
To imitate trajectories of task-space position (Appendix \ref{subsubsubsec:transformation_system_task_position}), $P$ data points of the demonstrated trajectory, $\mathbf{p}^{(d)}(t_i), \dot{\mathbf{p}}^{(d)}(t_i), \ddot{\mathbf{p}}^{(d)}(t_i) \in\mathbb{R}^{3}$ should be collected for $i\in\{1, 2,\cdots, P\}$. 
With these data points, the following matrices $\mathbf{A}\in \mathbb{R}^{N\times P}$ and $\mathbf{B}\in\mathbb{R}^{3\times P}$ are calculated:
\begin{align}\label{eq:Imitation_Learning_task_space_position_AB}
\begin{split}
    \mathbf{A} =&
    \begin{bmatrix}
        \mathbf{a}(s^{(d)}(t_{1})) & \cdots & \mathbf{a}(s^{(d)}(t_{P})) 
    \end{bmatrix} \\
    \mathbf{B} =& 
    \begin{bmatrix}
        \mathbf{b}_{p}(t_{1}) & \mathbf{b}_{p}(t_{2}) & \cdots & \mathbf{b}_{p}(t_{P})
    \end{bmatrix} \\    
    \mathbf{b}_{p}(t_i) =& (\tau^{(d)})^2  \ddot{\mathbf{p}}^{(d)}(t_i) + \alpha_z \tau^{(d)}  \dot{\mathbf{p}}^{(d)}(t_i) \\
    &+ \alpha_z \beta_z (\mathbf{p}^{(d)}(t_i)-\mathbf{p}_{g}^{(d)}) 
\end{split}
\end{align}
where $s^{(d)}$ for discrete (Appendix \ref{subsubsubsec:canonical_system_discrete}) and rhythmic movements (Appendix \ref{subsubsubsec:canonical_system_rhythmic}) are identical to those in Equation \eqref{eq:canonical_system_w_data}. 

For discrete movement, $\tau_{d}^{(d)}$ is the duration of the demonstrated trajectory, i.e., $\tau_d^{(d)}=t_P-t_1$; 
for rhythmic movement, $\tau_r^{(d)}$ is the period of the demonstrated trajectory divided by $2\pi$, i.e., $\tau_r^{(d)}=(t_P-t_1)/2\pi$. 

For discrete movement, $\mathbf{p}_{i}^{(d)}=\mathbf{p}^{(d)}(t_1)$ and $\mathbf{p}_{g}^{(d)}=\mathbf{p}^{(d)}(t_P)$ are chosen to be the initial and final data points of the demonstrated trajectory, respectively (Equation \eqref{eq:imit_learning_scaling_position}).
For rhythmic movement, $\mathbf{p}_{g}^{(d)}$ is chosen to be the mean value of $\mathbf{p}^{(d)}(t_i)$ for $i\in\{1,2,\cdots, P\}$. 

Note that in contrast to the original formulation of DMP \citep{hoffmann2009biologically,ijspeert2013dynamical} where the weights cannot be learned if one of the coordinates starts and ends at the same position (despite a non-zero movement), the formulation adopted by \cite{koutras2020novel} addresses this problem (Equation \eqref{eq:DMP_transformation_task_position}), allowing Imitation Learning to be applied in such cases.

To conduct Imitation Learning, the $P$ sample points of $\mathbf{p}^{(d)}(t_i), \dot{\mathbf{p}}^{(d)}(t_i), \ddot{\mathbf{p}}^{(d)}(t_i)$ for $i\in\{1,2,\cdots, P\}$ must be collected.
For this, one can sample the task-space position trajectories $\mathbf{p}^{(d)}(t)$ and use finite difference methods with smoothing to collect the higher derivative terms. 

\paragraph{For Robotic Manipulator}
If one uses a robotic manipulator to conduct Imitation Learning of task-space position (e.g., motion planning for the end-effector's position), data points $\mathbf{p}^{(d)}(t)$ and their higher derivatives can be derived by collecting the joint trajectories, their higher derivatives up to joint velocity and acceleration, and the differential kinematics of the robotic manipulator. 

In detail, to derive $\mathbf{p}^{(d)}(t)$, joint trajectory collected from demonstration, $\mathbf{q}^{(d)}(t)$ and the Forward Kinematics map of the robot for position, $\mathbf{h}_p$ are required, i.e., $\mathbf{p}^{(d)}(t)=\mathbf{h}_{p}(\mathbf{q}^{(d)}(t))$. 

To derive $\dot{\mathbf{p}}^{(d)}(t)$, along with the joint trajectory $\mathbf{q}^{(d)}(t)$ collected from demonstration, the Jacobian matrix of the robot $\mathbf{J}_{p}(\mathbf{q}^{(d)}(t))$, joint velocity $\dot{\mathbf{q}}^{(d)}(t)$ are required \citep{siciliano2008springer,lachner2024exp}. 
With these data points, $\dot{\mathbf{p}}^{(d)}(t)=\mathbf{J}_{p}(\mathbf{q}^{(d)}(t))\dot{\mathbf{q}}^{(d)}(t)$. The Jacobian matrix $\mathbf{J}_p(\mathbf{q})$ can be derived by partial derivatives of the Forward Kinematics map of the robot, i.e., for $\mathbf{h}_{p}:\mathcal{Q}\rightarrow \mathbb{R}^{3}$ such that $\mathbf{h}_{p}(\mathbf{q})=\mathbf{p}$, $\mathbf{J}_p(\mathbf{q})=\frac{\partial \mathbf{h}_{p}}{\partial \mathbf{q}}(\mathbf{q})$ \citep{siciliano2008springer}. 
One can also derive $\mathbf{J}_p(\mathbf{q})$ using the ``Product-of-Exponentials'' formula \citep{brockett1983robotic,murray1994mathematical,brockett2005robotic,lynch2017modern,lachner2024exp}.

To derive $\ddot{\mathbf{p}}^{(d)}(t)$, the time derivative of the Jacobian matrix $\dot{\mathbf{J}}_{p}(\mathbf{q}^{(d)}(t))$ and joint acceleration are additionally required. 
With these data, the following differential kinematics is used to derive $\ddot{\mathbf{p}}^{(d)}(t)$:
$$
    \ddot{\mathbf{p}}^{(d)}(t) = \dot{\mathbf{J}}_{p}(\mathbf{q}^{(d)}(t))\dot{\mathbf{q}}^{(d)}(t) + \mathbf{J}_{p}(\mathbf{q}^{(d)}(t))\ddot{\mathbf{q}}^{(d)}(t)
$$
The analytical derivation of $\dot{\mathbf{J}}_{p}(\mathbf{q}(t))$ can be used for the computation. 
If the sampling rate of the demonstrated trajectory is sufficiently high, numerical differentiation of the Jacobian matrices $\mathbf{J}_{p}(\mathbf{q}^{(d)}(t))$ with smoothing can also be used.

\subsubsection{For Task-space Orientation}\label{subsubsubsec:imit_learning_task_orientation}
To imitate task-space trajectories for spatial orientation, either using $\text{SO}(3)$ (Appendix \ref{subsubsubsec:transformation_system_task_orientation_SO3}) or $\mathbb{H}_1$ (Appendix \ref{subsubsubsec:transformation_system_task_orientation_H1}), $P$ data points of the demonstrated trajectory, $\mathbf{e}^{(d)}(t_i), \dot{\mathbf{e}}^{(d)}(t_i), \ddot{\mathbf{e}}^{(d)}(t_i) \in\mathbb{R}^{3}$ should be collected for $i\in\{1, 2,\cdots, P\}$. 
With these data points, the following matrices $\mathbf{A}\in \mathbb{R}^{N\times P}$ and $\mathbf{B}\in\mathbb{R}^{3\times P}$ are calculated:
\begin{align}
\begin{split}
    \mathbf{A} =&
    \begin{bmatrix}
        \mathbf{a}(s^{(d)}(t_{1})) & \cdots & \mathbf{a}(s^{(d)}(t_{P})) 
    \end{bmatrix} \\       
    \mathbf{B} =& 
    (\mathbf{S}_{e}^{(d)})^{-1}
    \begin{bmatrix}
        \mathbf{b}_{e}(t_{1}) & \mathbf{b}_{e}(t_{2}) & \cdots & \mathbf{b}_{e}(t_{P})
    \end{bmatrix} \\
    \mathbf{b}_{e}(t_i) =& (\tau^{(d)})^2\ddot{\mathbf{e}}^{(d)}(t_i) + \alpha_z \tau^{(d)}  \dot{\mathbf{e}}^{(d)}(t_i) \\
    &+ \alpha_z \beta_z \mathbf{e}^{(d)}(t_i)
\end{split}
\end{align}
where $s^{(d)}$ for discrete (Appendix \ref{subsubsubsec:canonical_system_discrete}) and rhythmic movements (Appendix \ref{subsubsubsec:canonical_system_rhythmic}) are identical to those in Equation \eqref{eq:canonical_system_w_data}. 

For discrete movement, $\tau_{d}^{(d)}$ is the duration of the demonstrated trajectory, i.e., $\tau_d^{(d)}=t_P-t_1$; 
for rhythmic movement, $\tau_r^{(d)}$ is the period of the demonstrated trajectory divided by $2\pi$, i.e., $\tau_r^{(d)}=(t_P-t_1)/2\pi$. 

To derive the exponential coordinates $\mathbf{e}^{(d)}(t_i)$, the trajectories for spatial orientation are collected either using the representation of $\text{SO}(3)$ (Appendix \ref{subsubsubsec:transformation_system_task_orientation_SO3}) or $\mathbb{H}_1$ (Appendix \ref{subsubsubsec:transformation_system_task_orientation_H1}). 
The former is denoted by ${}^{S}\mathbf{R}_{0}^{(d)}(t)\in \text{SO}(3)$ and the later is denoted by ${}^{S}\vec{\mathbf{q}}_0^{(d)}(t)\in\mathbb{H}_{1}$.
To avoid clutter, the left superscript and right subscript notations are suppressed unless clarification is required, i.e., ${}^{S}\mathbf{R}_{0}^{(d)}(t)\equiv \mathbf{R}^{(d)}(t)$ and ${}^{S}\vec{\mathbf{q}}_0^{(d)}(t) \equiv \vec{\mathbf{q}}^{(d)}(t)$. 

Given the $P$ data points of $\mathbf{R}^{(d)}(t_i)$ or $\vec{\mathbf{q}}^{(d)}(t_i)$ for $i\in\{1,2,\cdots, P\}$, the corresponding exponential coordinates $\mathbf{e}^{(d)}(t_i)\in\mathbb{R}^3$ can be derived by:
\begin{align}
    \begin{split}
        \mathbf{e}^{(d)}(t_i) &= \text{Log}_{\text{SO}(3)}( \{\mathbf{R}^{(d)}(t_i)\}^{\top} \mathbf{R}_G^{(d)} ) ~~~ \text{or} \\ \mathbf{e}^{(d)}(t_i) &= 2\text{Log}_{\mathbb{H}}( \{\vec{\mathbf{q}}^{(d)}(t_i)\}^{*} \otimes \vec{\mathbf{q}}_G^{(d)} )
    \end{split}
\end{align}
For discrete movement, the last data point is used as the goal orientation of the demonstration, i.e., $\mathbf{R}^{(d)}(t_P)\equiv \mathbf{R}_g^{(d)}$ and $\vec{\mathbf{q}}^{(d)}(t_P)\equiv \vec{\mathbf{q}}_g^{(d)}$. 

For rhythmic movement, $\mathbf{R}_g^{(d)}$ and $\vec{\mathbf{q}}_{g}^{(d)}$ are defined by the Exponential map of the mean of the exponential coordinates:
\begin{align}
    \begin{split}
        \mathbf{R}_g^{(d)} &= \text{Exp}_{\text{SO}(3)}\bigg( \frac{1}{P}\sum_{i=1}^{P}\text{Log}_{\text{SO}(3)}( \mathbf{R}^{(d)}(t_i) ) \bigg) ~~~ \text{or} \\
        \vec{\mathbf{q}}_g^{(d)} &= \text{Exp}_{\mathbb{H}}\bigg( \frac{1}{P}\sum_{i=1}^{P}\text{Log}_{\mathbb{H}}( \vec{\mathbf{q}}^{(d)}(t_i) ) \bigg) 
    \end{split}
\end{align}
Once $P$ data points for $\mathbf{e}^{(d)}(t_i)$ are derived, a finite difference method with smoothing can be used to derive the higher derivative terms $\dot{\mathbf{e}}^{(d)}(t_i)$, $\ddot{\mathbf{e}}^{(d)}(t_i)$. 
If the data points $\dot{\mathbf{R}}^{(d)}(t_i)$ or $\dot{\vec{\mathbf{q}}}^{(d)}(t_i)$ are available, one can also use analytical solutions to derive $\dot{\mathbf{e}}^{(d)}(t_i)$ (Appendices \ref{appx_sec:time_derivative_of_logarithm_SO3} and \ref{appx_sec:time_derivative_of_logarithm_H1}).

For discrete movement, the scaling matrix from demonstration $\mathbf{S}_{e}^{(d)}\in\mathbb{R}^{3\times 3}$ is defined by:
\begin{align}
\begin{split}
    \mathbf{S}_{e}^{(d)} &=\text{diag}(\text{Log}_{\text{SO}(3)}( \{\mathbf{R}_{i}^{(d)}\}^{\top} \mathbf{R}_{g}^{(d)} ))) ~~~ \text{or} \\
\mathbf{S}_{e}^{(d)}&=2\text{diag}(\text{Log}_{\mathbb{H}}( \{\vec{\mathbf{q}}_i^{(d)}\}^* \otimes \vec{\mathbf{q}}_{g}^{(d)} )))
\end{split}
\end{align}
where $\mathbf{R}_{i}^{(d)} \equiv \mathbf{R}^{(d)}(t_1)$ and $\vec{\mathbf{q}}_{i}^{(d)} \equiv \vec{\mathbf{q}}^{(d)}(t_1)$.

\paragraph{For Robotic Manipulator}
If one uses a robotic manipulator to conduct Imitation Learning of task-space orientation (e.g., motion planning for the end-effector's orientation), data points $\mathbf{e}^{(d)}(t)$ and their higher derivatives can be derived by collecting the joint trajectories, their higher derivatives up to joint velocity and acceleration, and the differential kinematics of the robotic manipulator. 

In detail, consider frames $\{S\}$ and $\{B\}$. The former is attached at the base of the robot; the latter is attached at the end-effector of the robot, although the frame can be attached anywhere on the robot.
To derive $\mathbf{e}^{(d)}(t)$, the joint trajectory collected from demonstration, $\mathbf{q}^{(d)}(t)$ and the Forward Kinematics map of the robot for orientation, $\mathbf{h}_r$ are required. 
The spatial orientation of the end-effector is derived by ${}^{S}\mathbf{R}_{B}^{(d)}(t)=\mathbf{h}_{r}(\mathbf{q}^{(d)}(t))$. 
If one uses unit quaternions, the derived ${}^{S}\mathbf{R}_{B}^{(d)}(t)$ is converted to unit quaternion ${}^{S}\vec{\mathbf{q}}_{B}^{(d)}(t)$ (Appendix \ref{appx:subsec_SO3_to_unit_quaternion}). 
With either ${}^{S}\mathbf{R}_{B}^{(d)}(t)$ or ${}^{S}\vec{\mathbf{q}}_{B}^{(d)}(t)$, an appropriate Logarithmic map is used to derive the corresponding exponential coordinates $\mathbf{e}^{(d)}(t)$.

To derive $\dot{\mathbf{e}}^{(d)}(t)$, terms ${}^{S}\dot{\mathbf{R}}_{B}^{(d)}(t_i)$ (Appendix \ref{appx_sec:time_derivative_of_logarithm_SO3}) or ${}^{S}\dot{\vec{\mathbf{q}}}_{B}^{(d)}(t_i)$ (Appendix \ref{sec:quat_kin_equation}) are required. 
These terms can be derived from the angular velocity of frame $\{B\}$ expressed in frame $\{S\}$, ${}^{S}\bm{\omega}^{(d)}(t)$:
\begin{align}
    \begin{split}
        {}^{S}\dot{\mathbf{R}}_{B}^{(d)}(t_i) &= [ {}^{S}\bm{\omega}^{(d)}(t_i)] {}^{S}\mathbf{R}_{B}^{(d)}(t_i) ~~~ \text{or} \\
        {}^{S}\dot{\vec{\mathbf{q}}}_{B}^{(d)}(t_i) &= \frac{1}{2} {}^{S}\vec{\bm{\omega}}^{(d)}(t_i) \otimes {}^{S}\vec{\mathbf{q}}_{B}^{(d)}(t_i)
    \end{split}
\end{align}
To derive ${}^{S}\bm{\omega}^{(d)}(t)$, the Spatial Jacobian matrix \citep{lynch2017modern} of the robot for angular velocity ${}^{S}\mathbf{J}_{r}(\mathbf{q}^{(d)}(t))$ and joint velocities $\dot{\mathbf{q}}^{(d)}(t)$ are required. 
With these data, the angular velocity can be calculated by ${}^{S}\bm{\omega}^{(d)}(t)={}^{S}\mathbf{J}_{r}(\mathbf{q}^{(d)}(t))\dot{\mathbf{q}}^{(d)}(t)$.

To derive $\ddot{\mathbf{e}}^{(d)}(t)$, an analytical equation can be used. Nevertheless, it is sufficient to conduct numerical differentiation of $\dot{\mathbf{e}}^{(d)}(t)$ with smoothing for the derivation.
 
\renewcommand{\theequation}{D.\arabic{equation}}
\setcounter{equation}{0}  

\section{Alternative Formulation of a Module}\label{appx:alternative_formulation}
As shown in Section \ref{subsec:module_for_task_space_orientation_SO3}, the module for task-space orientation using spatial rotation matrices involves the partial derivative with respect to $\mathbf{q}$ (Equation \eqref{eq:module_for_SO3_complex}), which can be computationally expensive. 
Hence, an equivalent controller using unit quaternions is introduced, eliminating the need for partial derivatives (Section \ref{subsec:module_for_task_space_orientation_H1}).

Alternatively, a module that directly utilizes spatial rotation matrices without the need for partial derivatives can be employed, denoted as $\mathbf{Z}_{r}'({}^{S}\mathbf{R}_{B}, {}^{S}\mathbf{R}_{0})$ \citep{hermus2021exploiting}:
\begin{align}\label{eq:SO3_new_type_controller}
    \begin{split}
        \mathbf{Z}_{r}'({}^{S}\mathbf{R}_{B}, {}^{S}\mathbf{R}_{0}) =& {}^{B}\mathbf{J}_r^{\top}(\mathbf{q}) \{ \mathbf{K}_{r}' \text{Log}_{\text{SO}(3)}( {}^{S}\mathbf{R}_{B}^{\top} {}^{S}\mathbf{R}_{0} ) \\
        &- \mathbf{B}_{r} {}^{B}\bm{\omega} \}    
    \end{split}
\end{align}
In this equation, $\mathbf{K}_{r}'\in\mathbb{R}^{3\times 3}$ is the rotational stiffness matrix. 
As shown in \citep{kim2011concurrent}, given a kinematically redundant robotic manipulator, if $\mathbf{B}_{q}$, $\mathbf{K}_{p}$, $\mathbf{B}_{p}$, $\mathbf{K}_{r}'$, $\mathbf{B}_{r}$ are chosen to be symmetric positive definite matrices, $\mathbf{p}(t)\rightarrow \mathbf{p}_{g}$, ${}^{S}\mathbf{R}_{B}(t)\rightarrow {}^{S}\mathbf{R}_{G}$, and $\dot{\mathbf{q}}\rightarrow \mathbf{0}$ \citep{kim2011concurrent}.

\section{An Overview of Dynamical Systems Approach}\label{subsubsec:review_DS}
Dynamical Systems (DS)-based approaches use autonomous (nonlinear) dynamical systems, $\dot{\mathbf{x}}=\mathbf{f}(\mathbf{x})$ as fundamental building blocks to design a robot controller. 
Hence, the goal is to learn a stable vector field $\mathbf{f}$ (potentially from data) and how to combine  learned vector fields to generate a desired robot behavior \citep{billard2022learning}.

For stability analysis of vector fields, Lyapunov theory has been used \citep{slotine1991applied,khalil2002nonlinear}.
For discrete movements, stability of autonomous dynamical systems and their combinations has been demonstrated using the Lyapunov equation with identity matrix \citep{khansari2011learning} and quadratic matrices \citep{figueroa2018physically}. Recently, the stability analysis has been generalized to nonlinear geometric spaces \citep{fichera2024learning}.
For rhythmic movements, dynamical systems with stable limit cycles (e.g., the Andronov-Hopf oscillator \citep{khalil2002nonlinear}) and their smooth deformation are employed \citep{khoramshahi2018human}.
For the combination of discrete and rhythmic movements, bifurcation has been used, where the attractor dynamics of an autonomous dynamical system switch between stable limit cycles and stable fixed points through changing a small set of parameters \citep{khadivar2021learning}.
Rhythmic movement can also be generated by sequencing (or smoothly activating) point-to-point discrete movements \citep{medina2017learning}.

By modulating a vector field, rapid adaptation of robot behavior to unknown environments can be achieved \citep{billard2022learning}. One example is real-time obstacle avoidance, where the vector field is real-time modulated to push  the robot away from (possibly non-stationary) obstacles \citep{khansari2012dynamical,huber2019avoidance,huber2022fast,huber2022avoiding,li2024representing}. 
Usually, a dynamic modulation matrix \citep{huber2019avoidance} is multiplied to locally or globally deform the vector field for obstacle avoidance.

Methods to learn the vector field from human demonstrations have been proposed. 
Often referred to as Learning from Demonstration (LfD) \citep{ravichandar2020recent,billard2022learning}, the key idea is to parameterize the vector field using a set of variables, collect a training dataset consisting of sample trajectories, and learn the parameters from this dataset.
For both parameterization and learning, probabilistic mixture models such as Gaussian Mixture Models (GMM) are often used \citep{khansari2011learning,calinon2020mixture}. 
To learn stable vector fields defined over general geometric spaces, including the Special Orthogonal Group for spatial rotations \citep{murray1994mathematical} (Appendix \ref{appx:SO3}), the underlying geometry in which the vector field resides is often explicitly accounted for during learning \citep{calinon2020gaussians,jaquier2021geometry,fichera2023implicit,duan2024structured,jaquier2024unraveling}.

One of the key advantages of using autonomous dynamical systems is that the state evolution depends solely on the current state of the system \citep{billard2022learning}. 
As a result, the robot converges to a desired asymptotic behavior, such as reaching a goal location or exhibiting a target rhythmic behavior, from any initial conditions. 
This time-independence property of DS-based approaches contributes to robustness and favorable stability properties, even in the presence of external disturbances or environmental uncertainty. 
However, autonomous dynamical systems cannot encode self-intersecting trajectories, although approaches to work around this limitation have been proposed \citep{khansari2011learning}. 
Moreover, the autonomous dynamical systems mostly represent movements. 
Thus, additional methods are necessary to map these learned movements into robot commands.
For instance, if the dynamical system represents task-space movements, solving Inverse Kinematics is required (Chapter 12 of \cite{billard2022learning}).

\bibliographystyle{SageH}
\bibliography{literature}

\end{document}